\documentclass[10pt,twocolumn,letterpaper]{article}

\usepackage{cvpr}      

\usepackage[dvipsnames]{xcolor}

\usepackage{graphicx}
\usepackage{amsmath}
\usepackage{amssymb}
\usepackage{booktabs}
\usepackage{bm}
\usepackage{mathtools}

\usepackage[export]{adjustbox}

\newcommand{\scene}[1]{\textit{#1}}

\newcommand{\nerf}{{NeRF}}

\definecolor{purple}{rgb}{0.5,0.0,0.97}
\definecolor{cyan}{rgb}{0.0,0.6,0.8}

\newcommand{\alen}{0.16\linewidth}

\newcommand{\flen}{0.16\linewidth}

\newcommand{\dlen}{0.137\linewidth}
\newcommand{\elen}{0.33\linewidth}

\definecolor{cvprblue}{rgb}{0.21,0.49,0.74}
\usepackage[pagebackref,breaklinks,colorlinks,citecolor=cvprblue]{hyperref}

\def\paperID{****} 
\def\confName{CVPR}
\def\confYear{2024}

\title{Learning Novel View Synthesis from Heterogeneous Low-light Captures}

\author{Quan Zheng$^1$, Hao Sun$^2$, Huiyao Xu$^3$, Fanjiang Xu$^1$\\[5pt]
$^1$~Institute of Software, Chinese Academy of Sciences, $^2$~UCAS, $^3$~Zhejiang University\\
}

\begin{document}
\maketitle
\begin{abstract}
	Neural radiance field has achieved fundamental success in novel view synthesis from input views with the same brightness level captured under fixed normal lighting. Unfortunately, synthesizing novel views remains to be a challenge for input views with heterogeneous brightness level captured under low-light condition. The condition is pretty common in the real world. It causes low-contrast images where details are concealed in the darkness and camera sensor noise significantly degrades the image quality. To tackle this problem, we propose to learn to decompose illumination, reflectance, and noise from input views according to that reflectance remains invariant across heterogeneous views. To cope with heterogeneous brightness and noise levels across multi-views, we learn an illumination embedding and optimize a noise map individually for each view. To allow intuitive editing of the illumination, we design an illumination adjustment module to enable either brightening or darkening of the illumination component. Comprehensive experiments demonstrate that this approach enables effective intrinsic decomposition for low-light multi-view noisy images and achieves superior visual quality and numerical performance for synthesizing novel views compared to state-of-the-art methods.
\end{abstract}    
\section{Introduction} \label{sec:intro}

Neural radiance field (NeRF)~\cite{mildenhall2020nerf} has recently become a new paradigm for novel view synthesis from a corpus of input images of a scene. NeRF assumes that all input images are captured under sufficient lighting and each image has the same brightness level. Thus, it has difficulty in learning from input images captured under a low-light condition. The difficult case, however, is pretty common in the real world. Imagine the scenario to capture a low-light scene from multiple views with a smartphone.

\begin{figure}[t]
    \centering
	\includegraphics[width=0.123\textwidth]{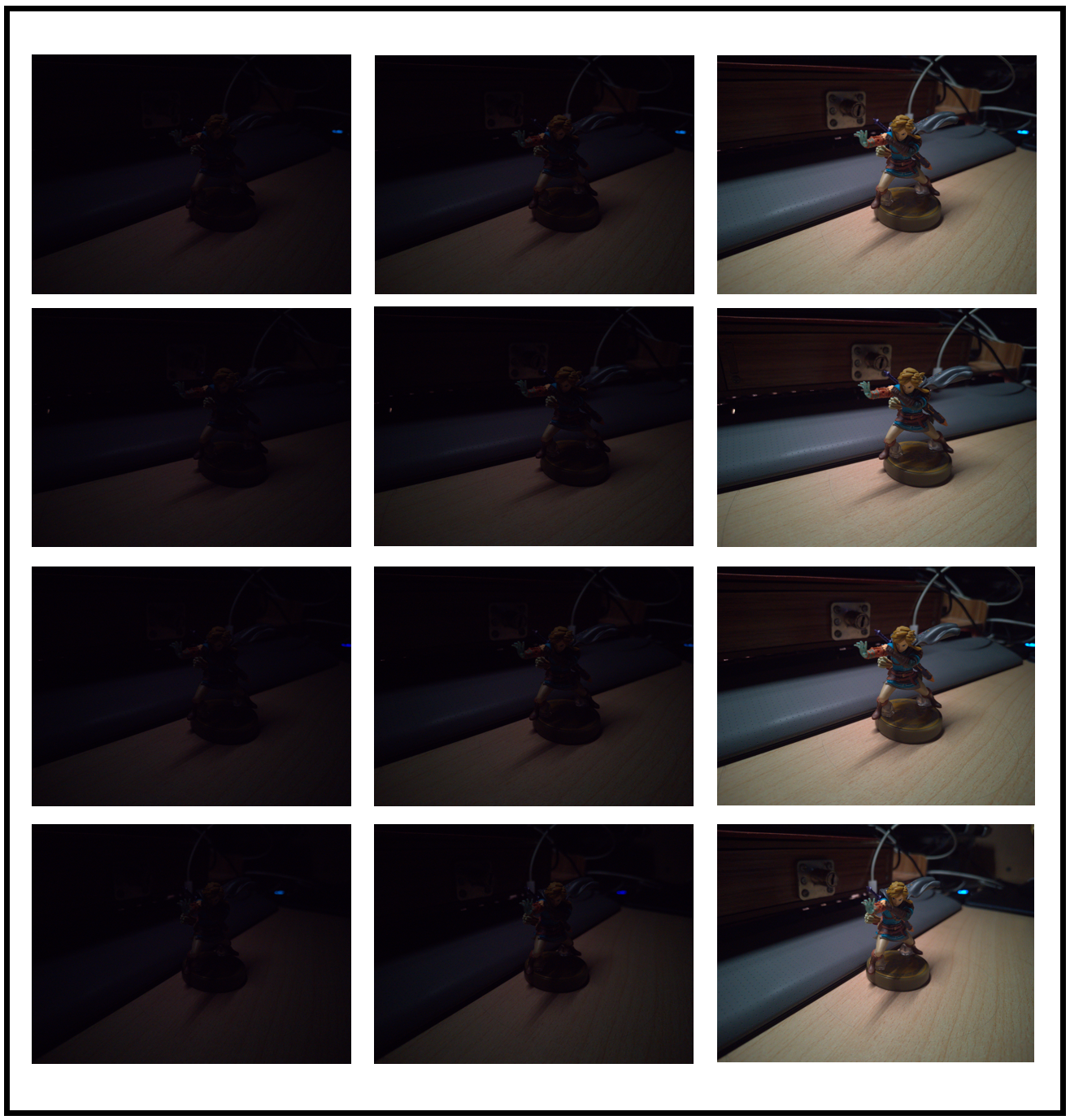}
	\includegraphics[width=0.169\textwidth]{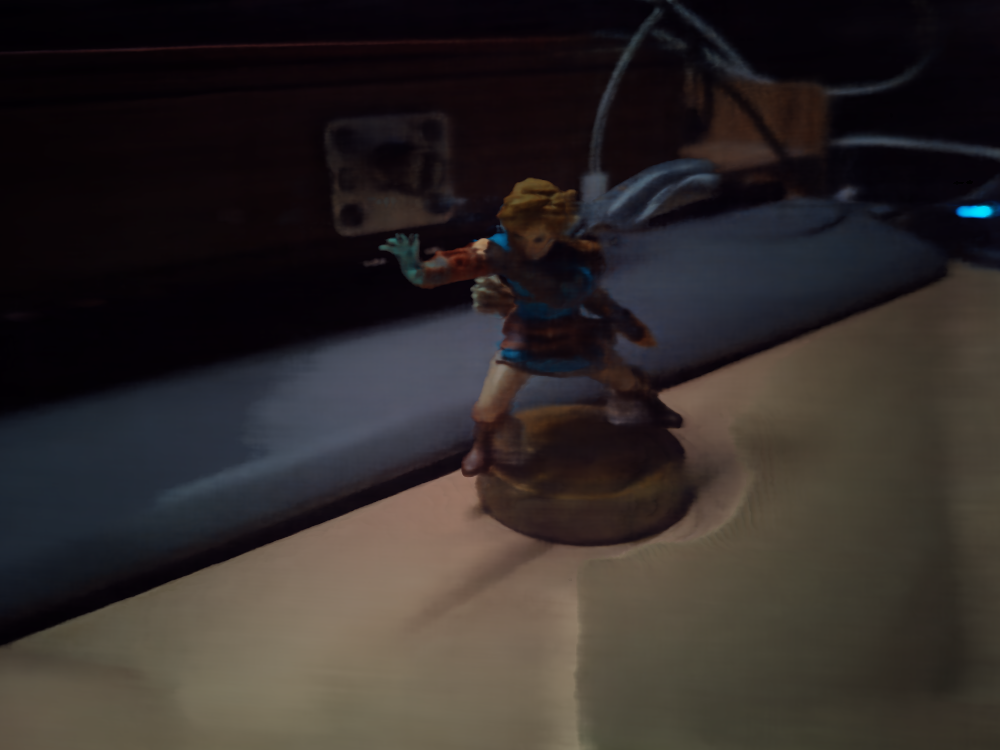}
	\includegraphics[width=0.169\textwidth]{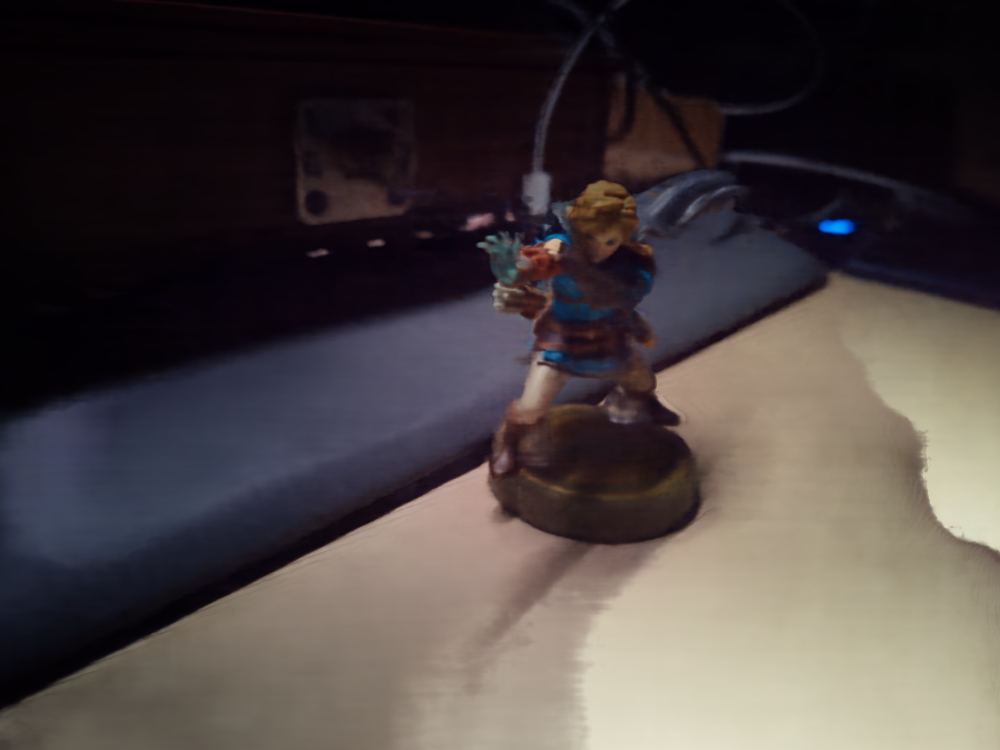}
	\caption{Images rendered from NeRF~\cite{mildenhall2020nerf} trained on the input views with heterogeneous brightness (\textit{left}) present unevenly illuminated artifacts  (\textit{middle} and \textit{right}).}
	\label{fig:inconsist}
\end{figure}

In the above low-light scenario, multi-view images are expected to be captured in a relatively short exposure time to simplify the operation and reduce the total time cost of the capturing. These images pose two challenges for NeRF training.
First, the images are featured with varying low brightness which violates the assumption of the vanilla NeRF. 
Second, the images are plagued with significant noise which causes distraction for NeRF training (\cref{fig:inconsist}).

Aleth-NeRF~\cite{cui2023aleth} proposes to learn an albedo and a concealing field for a low-light image, but this method requires that all input images have the same brightness levels. NeRFactor~\cite{zhang2021nerfactor} factorizes a scene into lighting, normals, albedo, and materials and assumes that multi-view images share the same brightness. For images with varying brightness, NeRF-W~\cite{martin2021nerf} proposes to encode the varying image appearance with view-wise appearance embedding. ExtremeNeRF~\cite{lee2023extremenerf} proposes to decompose a normal-light image into albedo and shading. All of them do not consider the noise issue which is non-negligible for real-world low-light images.

Inspired by the property that the intrinsic reflectance of a scene remains illumination-invariant across multiple views, we propose to decompose the input views into reflectance, illumination, and noise in a self-supervised manner, according to the generalized Retinex theory~\cite{land1977retinex}. The decomposition allows to edit the illumination component and eliminate the impact of noise. Yet, the decomposition is an ill-posed problem due to the ambiguity to explain the image with the three decomposed components. For instance, a dark pixel may be caused by low reflectance, low illumination, or even a noise value. To mitigate the ambiguity and form a plausible decomposition, we incorporate into the decomposition several priors that the reflectance is multi-view consistent, the reflectance value ranges from 0 to 1, and the illumination is locally smooth. Specifically, we design constraints on the reflectance, illumination, and noise components based on the priors.

In addition, we introduce an illumination embedding to encode the view-wise illumination. Considering the possible dynamic range of the illumination, we first learn high dynamic range (HDR) illumination, followed by a learnable tone mapping module to simulate the effect of converting HDR illumination to LDR values. Meanwhile, to cope with cross-view varying noise levels, we propose to learn an individual noise map for each view.

To allow an intuitive enhancement operation for synthesizing a novel view, we design an illumination adjustment module to learn the illumination adjustment operation. The module takes as input the decomposed illumination and an adjustment ratio, which serves as an intuitive interface to edit the illumination, without altering the intrinsic reflectance component.

To summarize, this paper has the following contributions:
\begin{itemize}
	\item We propose an unsupervised scheme to decompose real-world low-light captures into reflectance, illumination, and noise, and the decomposition enables the novel view synthesis from low-light noisy images with varying brightness.
	\item We propose to learn illumination embedding and individual noise map to cope with view-wise heterogeneous brightness and noise levels.
	\item We further design an illumination adjustment module to allow intuitive editing of the illumination of novel views.
\end{itemize}

\section{Related work}
\label{sec:related}

Our work explores novel view synthesis and brightness enhancement from low-light inputs. In the following, we focus on the majority of research works that are close to our work, and refer readers to~\cite{tewari2022advances} for a broad survey on neural rendering and~\cite{li2021low} on low-light image enhancement.

\noindent\textbf{Novel View Synthesis.}~Neural radiance field (NeRF)~\cite{mildenhall2020nerf, barron2022mipnerf360, mueller2022instant, yu2022plenoxels} leverages coordinate-based neural networks to learn a continuous representation and enables to synthesize novel views. However, NeRF assumes input views with homogeneous brightness and bakes the illumination of a scene into its radiance field, thus it does not allow editing the illumination. To provide the editability, a bunch of works have expanded NeRF to handle varying illumination and transient occluders~\cite{martin2021nerf,chen2022hallucinated,tancik2022block}. In terms of varying exposures, Huang et al.~\cite{huang2022hdr} and~\cite{jun2022hdr} learned a high dynamic range radiance field and a tone mapper for the conversion of dynamic ranges, but they are not designed for low-light captures with the sensor noise. To deal with the noise issue, Pearl et al.~\cite{pearl2022noiseaware} propose NeRF-based burst denoising to obtain clean photographs. 
RawNeRF~\cite{mildenhall2022nerf} proposed to learn a denoised high dynamic range radiance field from noisy RAW images. However, the multi-view RAW format images incur relatively high storage cost on mobile devices, thus reducing the practicality. In contrast, we propose to build a noise-aware neural radiance field from the commonly used sRGB mutli-view low-light images with varying brightness.

\noindent\textbf{Intrinsic Decomposition and Factorization.} Instead of learning the radiance, recent works ~\cite{verbin2022ref, boss2022samurai, Ye2023IntrinsicNeRF, liu2023nero} proposed to learn decomposed factors like albedo, roughness, and normals from multi-view images based on the assumed rendering models. Similarly, other works~\cite{srinivasan2021nerv, zhang2021physg, boss2021nerd, zhang2021nerfactor} leverage the inverse rendering framework to recover material and geometric properties~\cite{munkberg2022extracting, Jin2023TensoIR}. 
However, these methods generally assume the scene is captured under sufficient lighting, different from the tough low-light condition like ours. For low-light captures, camera sensor noise is 
non-negligible, but these methods do not have mechanisms to process such noise. A recent work~\cite{wang2023lighting} learns neural representation from low-light images via intrinsic decomposition, but its decomposition model includes only the reflectance and illumination. Also, it is not designed for input images with varying illumination.
In contrast, our approach learns to decompose the image into the reflectance, illumination and noise, which enables denoising and provides an intuitive control for the illumination. 

\noindent\textbf{2D Low-light Enhancement.}~Low-light Image Enhancement (LLIE) strives to improve the perception of images captured in poorly or unevenly lit environments. Early work are based on histogram equalization~\cite{pizer1990contrast} and the Retinex model-based iterative optimization~\cite{jobson1997properties,jobson1997multiscale}.
Guo et al.~\cite{guo2016lime} introduced structural prior along with the Retinex decomposition to estimate the illumination. 
Lore et al.~\cite{lore2017llnet} started a new genre of LLIE methods with the autoencoder architecture. Afterwards, recent years have witnessed the rapid development of deep learning-based LLIE methods based on diverse learning schemes, including supervised learning~\cite{chen2018learning,lv2018mbllen,wang2019underexposed,zhang2019kindling,zhu2020eemefn}, semi-supervised learning~\cite{yang2021band}, unsupervised learning~\cite{jiang2021enlightengan}, and reinforcement learning \cite{yu2018deepexposure}. Zero-DCE~\cite{guo2020zero} and its follow up~\cite{li2021learning} proposed to learn pixel-wise curve to adjust the dynamic ranges. 
Also, several works~\cite{ren2020lr3m,wei2018deep,yang2021sparse,zhao2021retinexdip} incorporate the Retinex decomposition into the design of neural network architectures. 
Recently, unrolling optimization with neural networks and weight sharing~\cite{liu2021retinex,ma2022toward,wu2022uretinex} are also intensively explored to accelerate the speed of LLIE. Fu et al.~\cite{fu2023learning} decompose the Retinex components from a pair of monocular low-light images, but the noise is not considered. Different from the above 2D LLIE methods, we focus on learning 3D disentangled neural representations from multi-view low-light images to allow synthesis and enhancement of novel views. 

\begin{figure*}[t]
	\centering
	\includegraphics[width=1.0\linewidth]{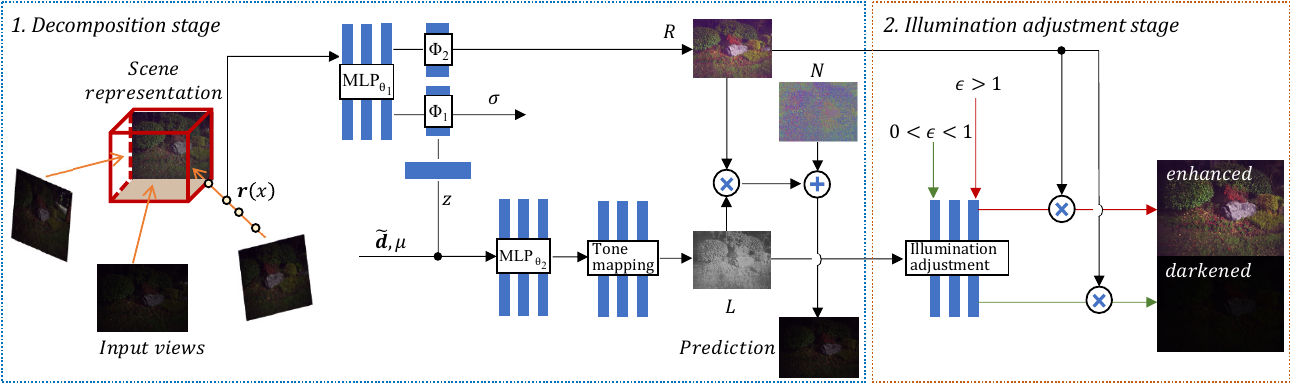}
	\caption{An overview of the proposed framework. In the first stage, our method learns disentangled reflectance, illumination, and noise components from low-light input images with varying brightness based on the Retinex theory. In the second stage, our method can robustly enhance or darken the illumination component. 	The adjusted image is synthesized by the product of the decomposed reflectance and the adjusted illumination.}
	\label{fig:fullframe}
\end{figure*}
\section{Preliminaries}\label{sec:prelim}

\textbf{Neural Radiance Fields.}~\nerf~learns a continuous volumetric representation of a scene through a series of posed images. Its 5D inputs, consisting of a spatial location and a viewing direction, are mapped to volume density and color through a multilayer perceptron (MLP), and the image is calculated by volume rendering. \nerf~emits a ray $\bm{r}(t)=\bm{o} +t\cdot\bm{d}$ from the camera projection center $\bm{o}$ with the distance $t$ along the direction $\bm{d}$. Sampling on the ray is performed between the near plane $t_n$ and the far plane $t_f$.
Feeding 5D inputs directly to \nerf~causes low quality at representing high-frequency content, so \textit{positional encoding} on the location and direction coordinates is employed.

The incident radiance of ray $\bm{r}$ can be computed with
\begin{equation} \label{eq:radiance}
	\hat{C}\left({\bm{r}}\right) = \int_{t_{n}}^{t_{f}}T\left(\bm{r}(t)\right)\sigma\left(\bm{r}(t)\right)c\left(\bm{r}(t), -\bm{d}\right) dt,
\end{equation}
where $T$ is the transmittance, $\sigma$ denotes the volume density, and $c$ is the direction-dependent color. The integral can be evaluated with numerical quadrature. It firstly casts rays and draws point samples along each ray to get density and color. The approximate radiance of each ray then can be computed by
$\hat{C}\left(\bm{r}\right)=\sum_{k=1}^{K}T\left(t_k\right)\alpha(\sigma(t_k)\delta_k)c(t_k)$, with transmittance $\ T\left(t_k\right)=\exp\left(-\sum_{j=1}^{k-1}{\sigma(t_{j})\delta_{j}}\right)$. $\hat{C}\left(\bm{r}\right)$ is the final predicted color of the pixel, 
where $\alpha(x)=1-\exp(-x)$, and $\delta_k=t_{k+1}-t_k$ is the distance between adjacent point samples.

\noindent\textbf{Robust Retinex decomposition.}~Robust Retinex model~\cite{li2018structure} assumes that an image can be decomposed into reflectance, illumination, and noise components. Let $S$ denote the source image, then the decomposition can be expressed as:
\begin{equation}\label{eqn:retinex}
	S=R\odot L + N ,
\end{equation}
where $R$ represents the reflectance, $L$ stands for the illumination, $N$ denotes the noise, and $\odot$ denotes the element-wise multiplication. The reflectance component describes the intrinsic color of a scene and is invariant under different lighting conditions.

\section{Method} \label{sec:method}

Targeting at real-world images captured in a low-light environment with varying brightness, we aim to learn a neural representation from the images and enable to synthesize and enhance novel views. According to the robust Retinex model~\cite{li2018structure}, we propose to learn a neural scene representation consisting of decomposed reflectance, illumination, and noise. In addition, we design an illumination adjustment module to allow intuitive editing of the brightness for novel views. 

\cref{fig:fullframe} shows an overview of our method, which consists of two stages. The first stage is the unsupervised decomposition process. In \cref{sec:decomp}, we introduce the modules to estimate the reflectance, illumination, and noise components. Then, we present the illumination adjustment module in \cref{sec:adj} and the training losses in \cref{sec:loss}, followed by the implementation details in \cref{sec:imple}.

\subsection{Unsupervised Intrinsic Decomposition} \label{sec:decomp}

\noindent\textbf{Density and reflectance estimation.}~
NeRF forms a volumetric scene representation, in which volume density only depends a spatial location. Similar to the density, reflectance is only related to the spatial location, based on the property that the reflectance at a location stays invariant across different views. While separate MLPs can be trained to learn the mapping from spatial locations to the density and the reflectance, we propose to reduce neural network parameters by weight sharing. Specifically, we leverage a single $MLP_{\theta_1}$ to learn the shared feature $w$, and send $w$ to two shallow neural networks $\phi_1$, $\phi_2$ for predicting the density $\sigma$ and the reflectance $R$, respectively (\cref{fig:fullframe}). Since the density is non-negative, Softplus ($f_{p}$) activation function is used. Meanwhile, Sigmoid ($f_{g}$) is employed for the reflectance to adapt to the value range prior from 0 to 1. In summary, the density and the reflectance of a spatial location can be formulated as:
\begin{equation}\label{}
\begin{gathered}
	w={MLP}_{\theta_{1}}\left(\bm{r}\left(x\right)\right),\\
	\sigma\left(\bm{r}\left(x\right)\right)=f_{p}\left(\phi_1\left(w\right)\right),\,
	R\left(\bm{r}\left(x\right)\right)=f_{g}\left(\phi_2\left(w\right)\right)
\end{gathered}
\end{equation}
where $\bm{r}(x)$ stands for a point on the ray $\bm{r}$, $\sigma(\cdot)$ is the density, $R(\cdot)$ refers to the reflectance.

\par
\noindent\textbf{Illumination estimation.}~Different from the multi-view invariant reflectance, the illumination may vary across multiple views. Therefore, we learn an illumination embedding $\mu$ for each input view to encode the varying illumination by leveraging \textit{Generative Latent Optimization}~\cite{bojanowski2018optimizing}. The embedding vector $\mu$ has the length $\eta$ ($\eta$ is set to $48$).

To minimize the coupling between the illumination and the reflectance, we leverage the feature from the hidden layer of $\phi_1$ in the density branch as one input to $MLP_{\theta_{2}}$. 
Then, $\mu$ and the position-encoded direction $\Tilde{\bm{d}}$ are concatenated with the feature $z$ to form the input for $MLP_{\theta_{2}}$. Since the Retinex theory assumes all color channels share the same illumination, $MLP_{\theta_{2}}$ predicts a one-channel illumination component $L_{h}$ in high dynamic range
\begin{equation}\label{}
	L_{h}\left(\bm{r}\left(x\right)\right)={MLP}_{\theta_2}\left(z,\Tilde{\bm{d}},\mu\right).
\end{equation}
Note that the feature $z$ of the position $\bm{r}\left(x\right)$ also make $MLP_{\theta_{2}}$ aware of the 3D location information. 
Then we use a neural network-based tone mapping module $\mathcal{T}$ to convert the high dynamic range illumination to low dynamic range illumination $L_{l}\left(\bm{r}\left(x\right)\right)=\mathcal{T}\left(L_{h}\left(\bm{r}\left(x\right)\right)\right)$.
The training of the decomposition stage only produces illumination embeddings for training images. The illumination embeddings of test images are initially unavailable. For test images, we first optimize its illumination embeddings such that the reconstructed images match the input.

\noindent\textbf{Noise estimation.}~Vanilla NeRF assumes that the input images are noise-free and share the same brightness level. However, images captured under a low-light condition suffer from the camera sensor noise. The noise generally varies across different views. RawNeRF~\cite{mildenhall2022nerf} learns a radiance field in the linear RGB space with multi-view RAW images to remove noise,  
whereas our method targets at the commonly used sRGB images. Note that the quantization operation of sRGB will alter the distribution of noise, such that the multi-view setting of NeRF cannot eliminate the impact of noise thoroughly. 

To this end, we propose to guide the neural networks to learn clean reflectance and illumination components, and optimize a 2D noise map $N$ individually for each view
\begin{equation}\label{eq:noiserange}
    N = \lambda_N\cdot \mathrm{Tanh}\left(\Pi\right),
\end{equation}
where $\Pi$ is a trainable image with the same resolution as the corresponding view, $\mathrm{Tanh}(\cdot)$ is the tangent hyperbolic function, $\lambda_N$ is a hyperparameter to constrain the magnitude of noise. Since the noise is camera sensor related, we do not learn a noise value for every 3D point sample.

\noindent\textbf{Rendering.}~Given the illumination $L$ and reflectance $R$ for a point sample $\bm{r}(t_k)$, we compute its color $c$ by taking a product as $c(\bm{r}\left(t_k\right)) = L\left(\bm{r}\left(t_k\right)\right)\odot R\left(\bm{r}\left(t_k\right)\right)$. Finally, the pixel color of the ray $\bm{r}$ can be calculated by
\begin{equation}
    \widehat{C}\left(\bm{r}\right)=N\left(\bm{r}\right)+\sum_{k=1}^{K}T\left(t_k\right)\alpha\left(\sigma\left(t_k\right)\delta_k\right)c\left(t_k\right),
\end{equation}
where $N\left(r\right)$ is the corresponding noise for the ray $\bm{r}$ and $t_k$ indicates the parametric distance.

\subsection{Illumination Adjustment}\label{sec:adj}

With the decomposed illumination, a trivial adjustment method is to apply linear transformation $\kappa\cdot L_{l}$ or gamma transformation $\kappa\cdot\left(L_{l}\right)^\gamma$ to the illumination, where $\kappa$ and $\gamma$ are parameters. 
However, these transformations require manual parameter tuning and tends to introduce color distortion artifacts~\cite{li2021low}.

To this end, we propose an illumination adjustment stage to transform the input illumination into the adjusted illumination (\cref{fig:fullframe}). The adjusted image can be composited from the product of the adjusted illumination and the reflectance. Because the brightness of images does not have a quantitative characterization, we introduce a variable $\epsilon$ to describe the ratio between the output brightness and the input brightness. 
We leverage images of the same input view with both higher and lower brightness than the input view as the supervisory signals for the stage $2$. Images of the high brightness level use two times longer exposure time than the input view, meanwhile images in the low brightness level use only half the exposure time of the input view. Based on the supervisory data, we train the stage 2 to learn the illumination adjustment operations by using the $\epsilon$ as a conditional input.

\subsection{Training Losses}\label{sec:loss}
To enable the robust Retinex decomposition for stage 1, we design the loss $\mathcal{L}_{s1}$ which can be expressed as:
\begin{equation}\label{eq:stage1loss}
    \mathcal{L}_{s1}=\mathcal{L}_{recon}+\mathcal{L}_{R}+\mathcal{L}_{L}+\mathcal{L}_{N},
\end{equation}
where $\mathcal{L}_{recon}$, $\mathcal{L}_{R}$, $\mathcal{L}_{L}$ , $\mathcal{L}_{N}$ denotes the constraint on the reconstructed image, the illumination, the reflectance, and the noise, respectively. These loss terms are detailed as below.

\noindent\textbf{Reconstruction loss.}~The decomposed reflectance $R$, illumination $L$, and noise $N$ should be able to reconstruct the input view $S$. Therefore, the reconstruction loss is 
\begin{equation}
\mathcal{L}_{recon} = \lambda_c\cdot\Vert S-\left(R\cdot L+N\right)\Vert_{1}.
\end{equation}

\noindent\textbf{Illumination loss.}~We first use an illumination consistency term $\Vert L-L_{0}\Vert_1$ to guide the estimated illumination $L$ to be similar to the initial illumination estimation $L_{0}$. In the Retinex theory, $L_{0}$ is computed from the maximum value of R, G, B channels as $L_{0}\left(p\right)=\max_{c\in \{R,G,B\}}S\left(p\right)$, where $p$ is a pixel of the input view. Also, to encourage the illumination map be smooth in textural regions and preserve structural boundary, we apply a weighted total variation term $\Vert w_{h}\cdot\left(\nabla_h L\right)\Vert_1 + \Vert w_{v}\cdot\left(\nabla_v L\right)\Vert_1$, where $\nabla_{h,v}$ refers to the horizontal and vertical gradients. The weights $w_{h}$, $w_{v}$ are defined as $w_{h} = 1/\left(\nabla_{h}L_{0}\right)$, $w_{v} = 1/\left(\nabla_{v}L_{0}\right)$, respectively. Here, the weights ensure that structural boundaries are preserved. Therefore, the illumination loss is
\begin{equation}
    \resizebox{0.875\hsize}{!}{$\mathcal{L}_{L}=\lambda_{i}\cdot\Vert L-L_{0}\Vert_{1} + \lambda_{g}\cdot\left(\Vert w_{h}\cdot\left(\nabla_h L\right)\Vert_1 + \Vert w_{v}\cdot\left(\nabla_v L\right)\Vert_1\right)$}.
\end{equation}

\noindent\textbf{Reflectance loss.}~Based on the estimated illumination, we can compute the reflectance via a pixel-wise division as $S/L$. Therefore, we can form the constraint to guide the learning of the reflectance as
\begin{equation}
    \mathcal{L}_{R} = \lambda_{r}\cdot\Vert R- S / \mathrm{sg}\left(L\right) \Vert_1,
\end{equation}
where $\mathrm{sg}(\cdot)$ is the stop gradient operation, which helps to stabilize the training process.

\noindent\textbf{Noise regularization loss.}~Based on the observation that the noise level in the dark regions is perceptually higher, we introduce the constraint $\Vert S\cdot N\Vert_{F}$, where $S$ provides the information of dark regions and $\Vert\cdot\Vert_{F}$ is the Frobenius norm. In addition, the noise distribution of small patches on a noise map should be the same. Thus, we introduce a standard deviation term ${V}\left[{M}\left(\zeta_{\{K\}}\right)\right]$ to encourage that noise patches share a uniform mean value, where ${V}(\cdot)$ and ${M}(\cdot)$ denote the operators to compute standard deviation and mean, respectively, $\zeta_{\{K\}}$ is a set of $K$ noise patches. The noise regularization loss is formulated as
\begin{equation}
    \mathcal{L}_{N} = \lambda_{n}\cdot\Vert S\cdot N\Vert_{F} + \lambda_{s}\cdot{V}\left[{M}\left(\zeta_{\{K\}}\right)\right].
\end{equation}

The above $\lambda_c$, $\lambda_i$, $\lambda_g$, $\lambda_r$, $\lambda_n$, $\lambda_s$ are balancing weights. We find the values of the balancing weights with the grid search method.

To train the illumination adjustment module of stage $2$, we use the $\mathcal{L}^2$ reconstruction loss:
\begin{equation}\label{}
	\mathcal{L}_{s2}=\Vert {C_d}\left(\bm{r}\right)-\widehat{C_d}\left(\bm{r}\right)\Vert^2_2 + \Vert {C_e}\left(\bm{r}\right)-\widehat{C_e}\left(\bm{r}\right)\Vert^2_2.
\end{equation}
Here, $C_d$ and $C_e$ are the reference darkened and enhanced pixel color, respectively, and $\widehat{C_d}$ and $\widehat{C_e}$ are the predicted colors.

\subsection{Implementation Details}\label{sec:imple}

\textbf{Architecture.}~The highest positional encoding frequency is $2^{15}$ for locations and $2^4$ for directions. In the decomposition stage, $MLP_{\theta_1}$ takes in the position encoded location and has eight fully-connected hidden layers with $256$ neurons in each layer. $MLP_{\theta_2}$ receives the incoming features and processes them using a fully-connected hidden layer with $128$ neurons. At the illumination adjustment stage, the adjustment neural network has a 128-dimensional fully-connected hidden layer. All hidden layers are configured with ReLU activations.

\noindent\textbf{Training details.}~We implement our approach with PyTorch. We first train the decomposition stage to minimize the loss (\cref{eq:stage1loss}) by utilizing the images from the middle brightness level. Afterwards, we keep the trained weights of the decomposition stage. For the illumination adjustment stage, we leverage the images from the high brightness level and the low brightness level as supervisory data. To learn the enhancement operation, we construct training pairs that have a two-times relation between the exposure time (\cref{sec:adj}) and set the ratio $\epsilon$ to 2. Similarly, we construct training pairs that have a one-half relation between the exposure time to learn the darkening operation and set the darkening ratio $\epsilon$ to $1/2$.

Both stages are trained by the Adam optimizer with its default hyperparameters. the initial learning rate $5\times10^{-4}$ is exponentially decayed to $5\times10^{-5}$ with a scheduler. $\lambda_{N}$ of \cref{eq:noiserange} is set to 0.2. The values of balancing weights $\lambda_c$, $\lambda_i$, $\lambda_g$, $\lambda_r$, $\lambda_n$, $\lambda_s$ for loss terms are presented in the supplementary. 
The decomposition stage is trained for $45,000$ iterations. The mini-batch size of rays is $10,000$. The rays are collected from $100$ sampled $10\times 10$ small patches from input views. The illumination adjustment stage is trained for $6,000$ iterations with a mini-batch size $30,000$. The training of stage 1 takes about 20 hours on a Nvidia A100 GPU, and stage 2 takes 2 hours. We refer readers to the supplementary for other details.

\begin{figure*}[t]
    \centering
    \hspace{-3.0mm}\rotatebox{90}{\makebox[1.6cm][c]{\small{{\scene{Shrub}}}}}\hfill
    \includegraphics[width=\alen]{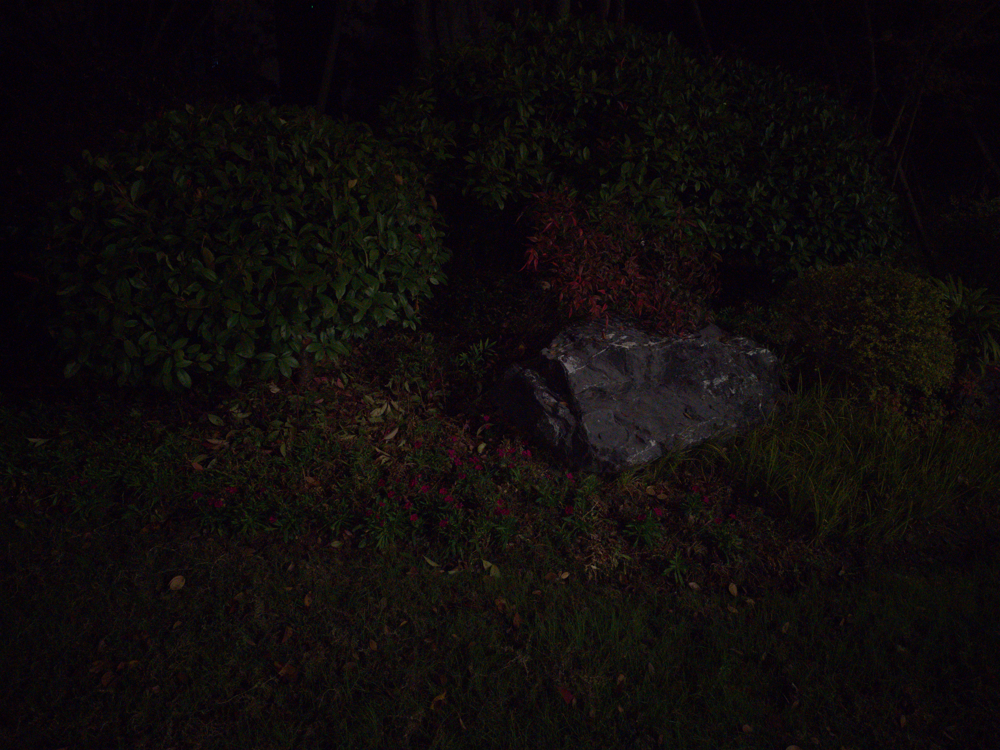}\hfill 
    \includegraphics[width=\alen]{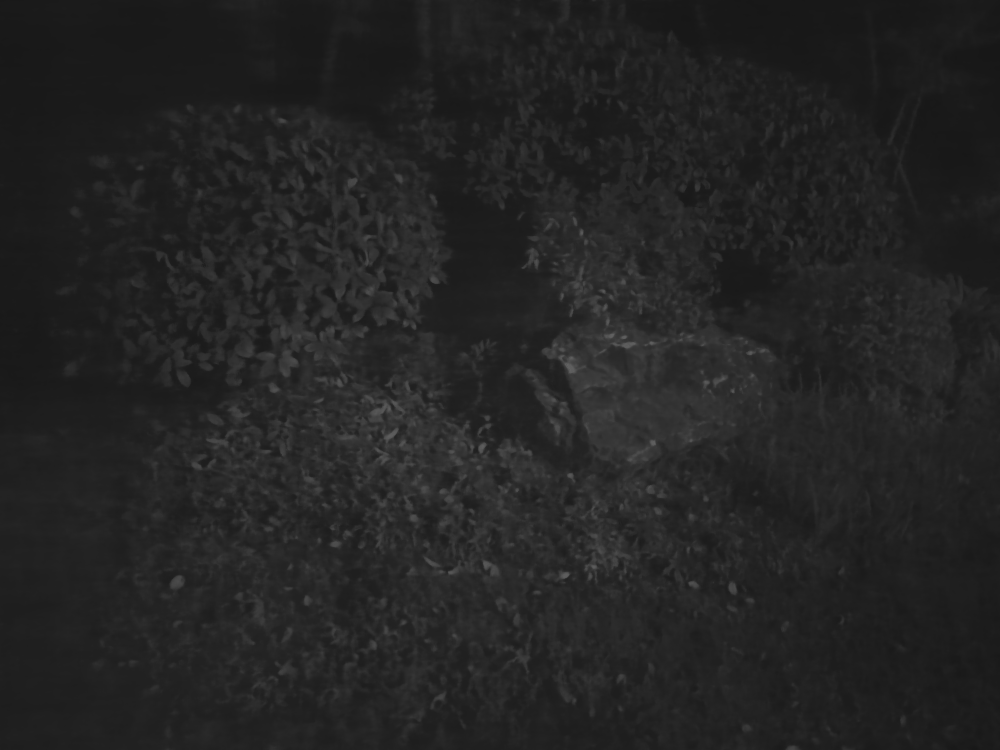}\hfill 
    \includegraphics[width=\alen]{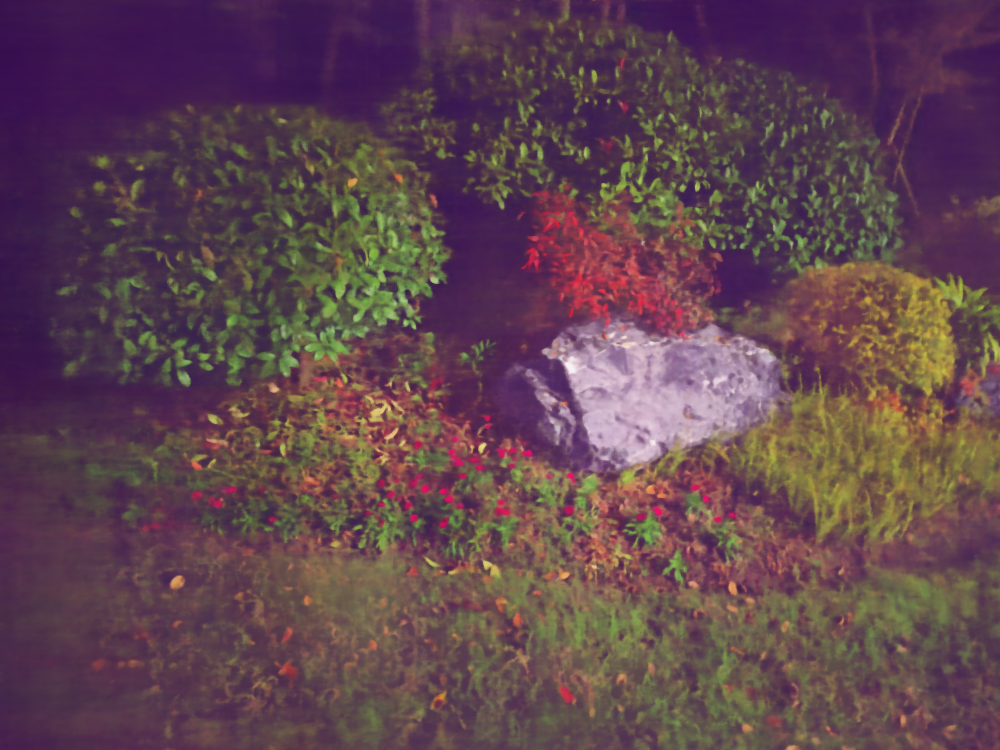}\hfill 
    \includegraphics[width=\alen]{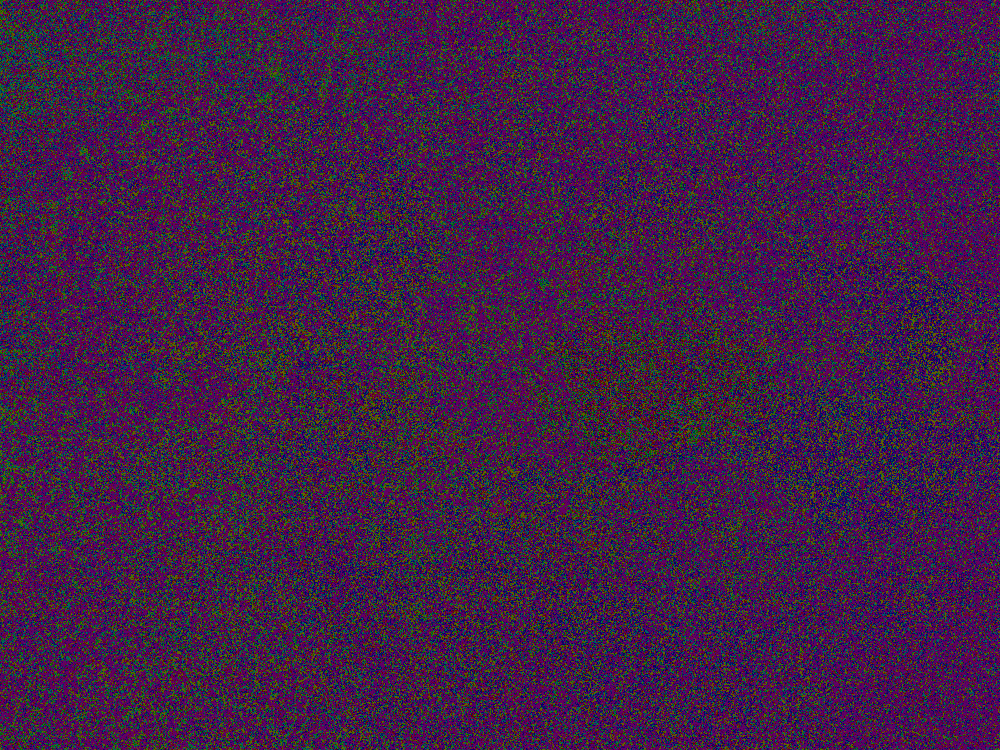}\hfill 
    \includegraphics[width=\alen]{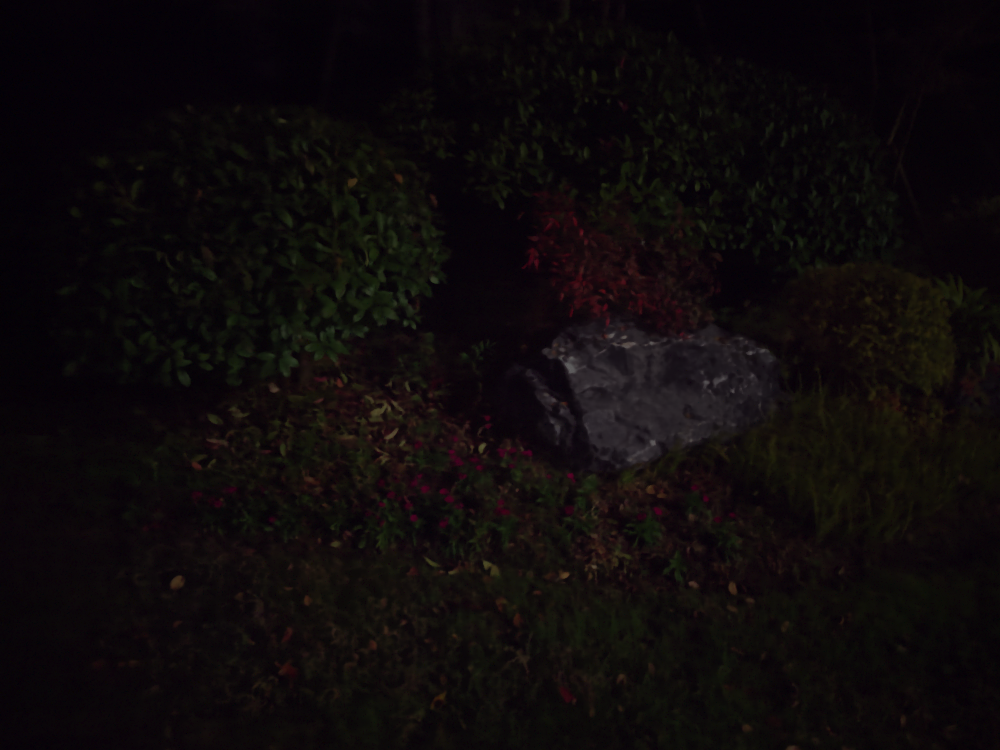}\hfill 
    \includegraphics[width=\alen]{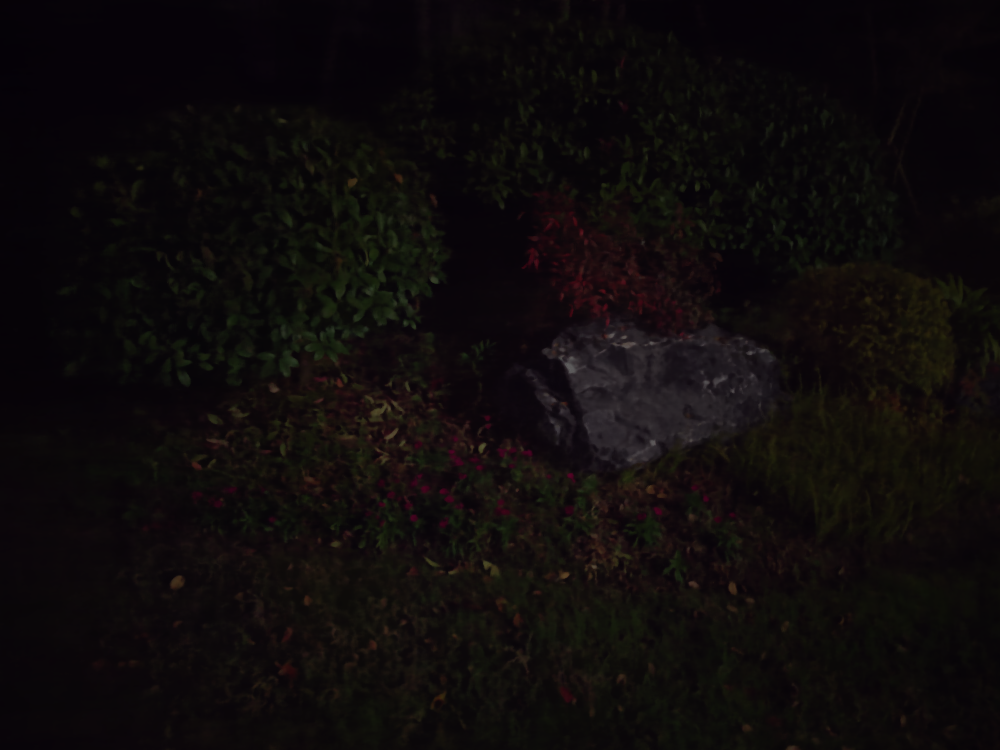}\\
    \hspace{-3.0mm}\rotatebox{90}{\makebox[1.6cm][c]{\small{{\scene{Link}}}}}\hfill
    \includegraphics[width=\alen]{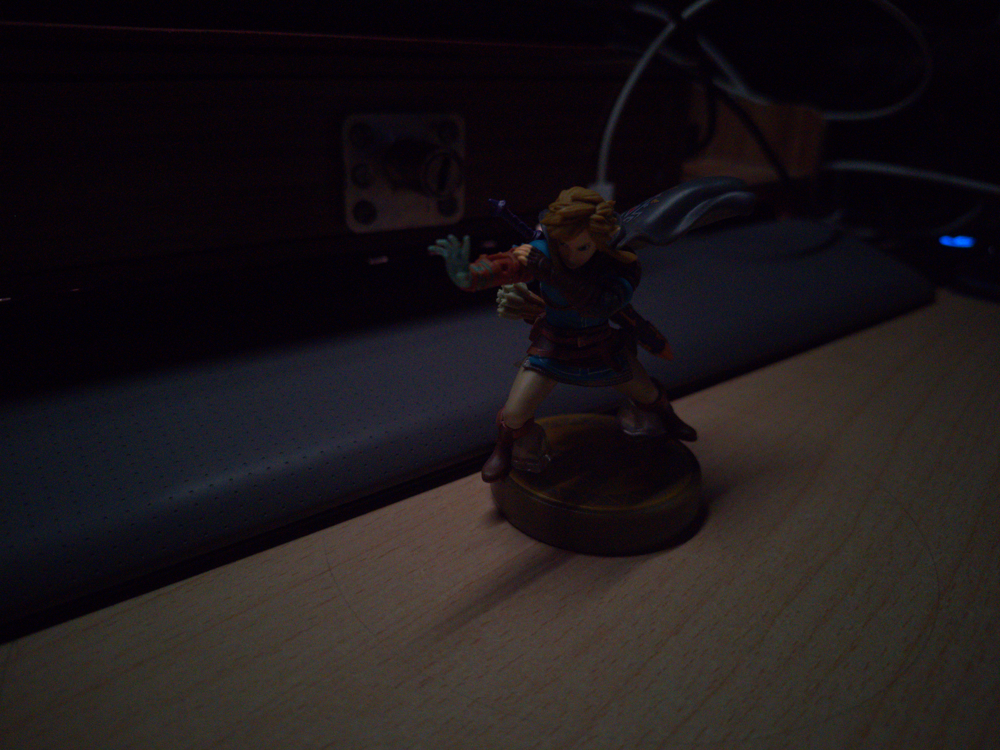}\hfill 
    \includegraphics[width=\alen]{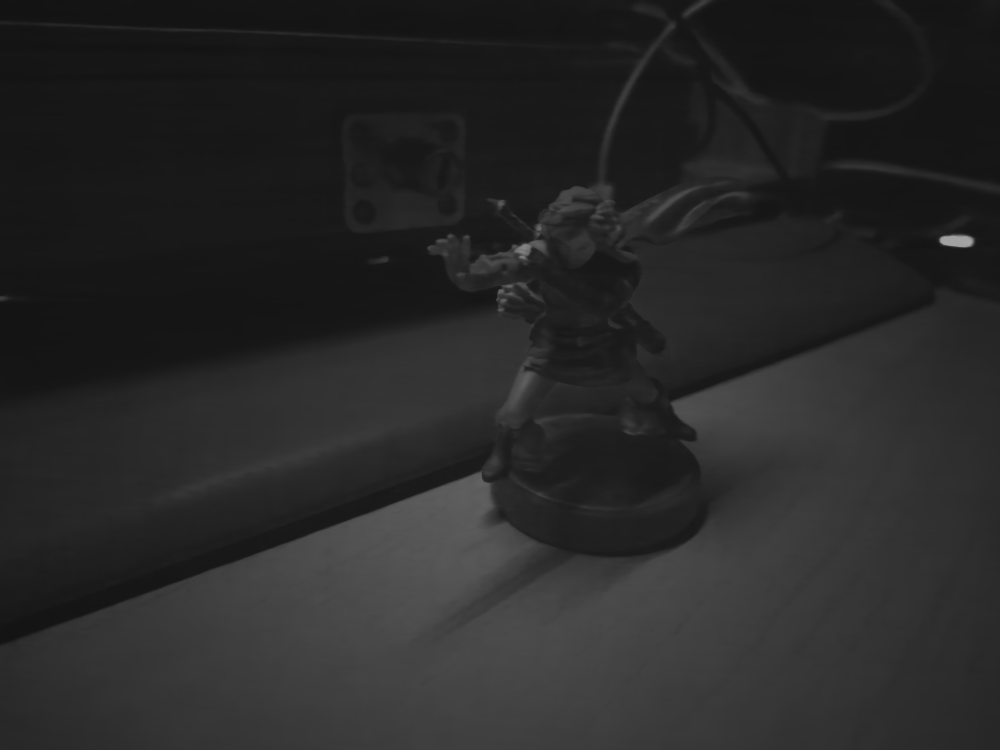}\hfill 
    \includegraphics[width=\alen]{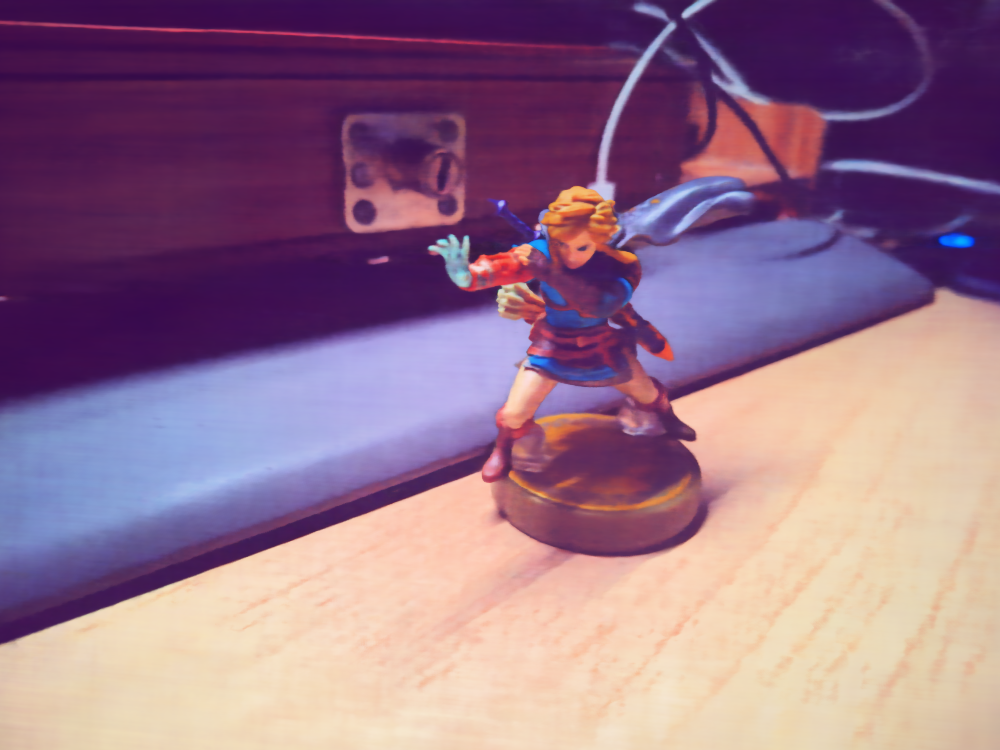}\hfill 
    \includegraphics[width=\alen]{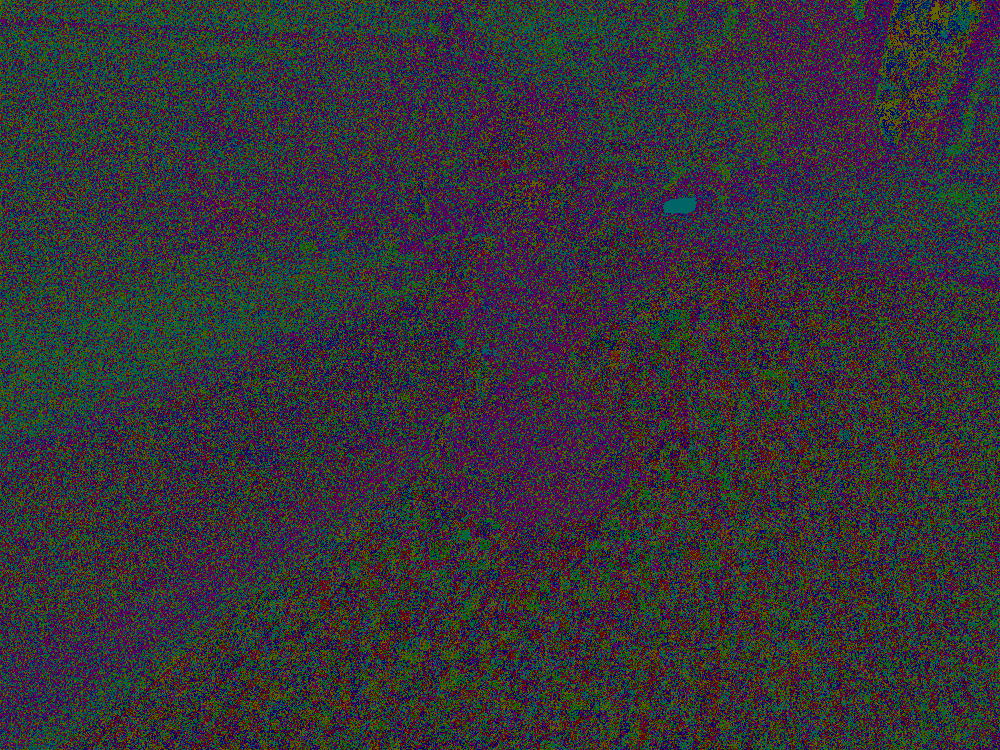}\hfill 
    \includegraphics[width=\alen]{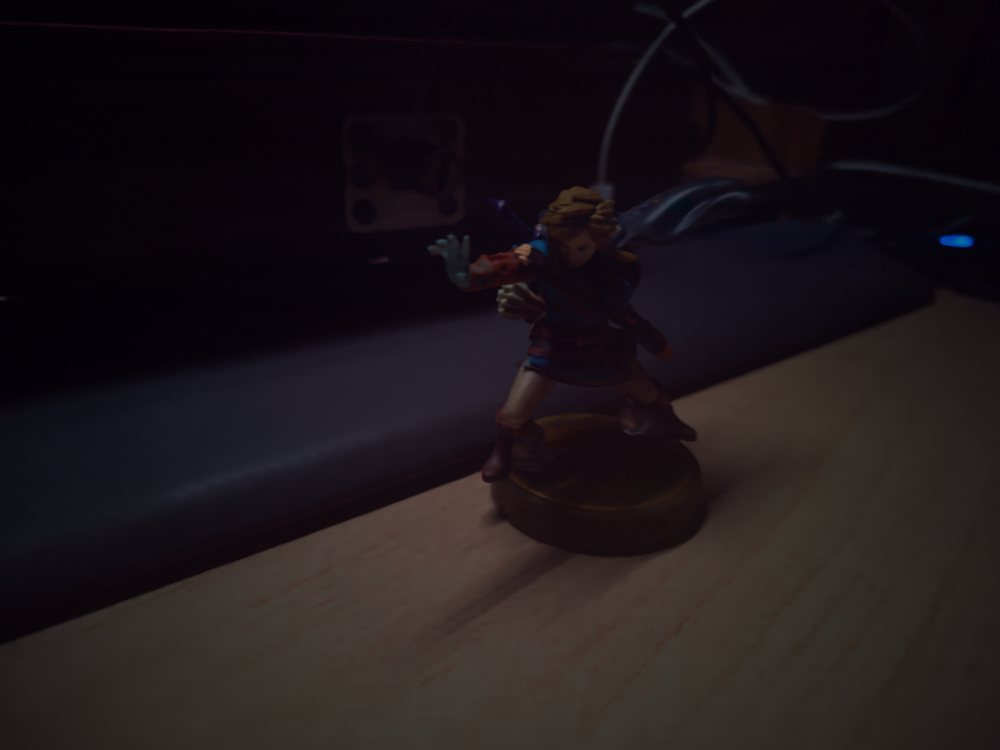}\hfill 
    \includegraphics[width=\alen]{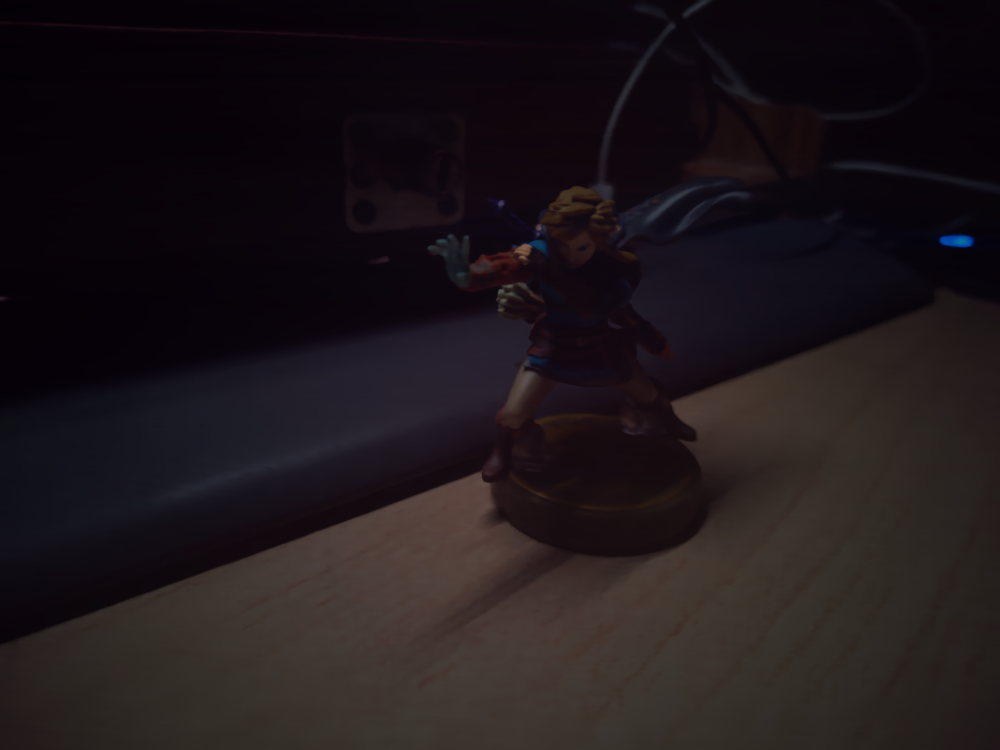}\\
    \makebox[\flen][c]{\small{Test view}}\hfill
    \makebox[\flen][c]{\small{Illumination}}\hfill
    \makebox[\flen][c]{\small{Reflectance}}\hfill
    \makebox[\flen][c]{\small{Noise}}\hfill
    \makebox[\flen][c]{\small{Recon w. noise}}\hfill
    \makebox[\flen][c]{\small{Recon w.o. noise}}\hfill\\
    \caption{The decomposition results of our approach on the test views of the \scene{Shrub} and \scene{Link} scenes. The reconstructions with and without the noise component are also presented. The noise maps are normalized to $[0, 0.4]$ for the visualization.}
    \label{fig:decomp_eval}
\end{figure*}
\section{Experiments} \label{sec:exp}

In the experiments, we capture five datasets of real-world scenes. Four of them are indoor scenes and the other one is an outdoor scene. In this section, we present results of four forward-facing scenes. The other $360^{\circ}$ scene is in the supplementary material. All scenes are captured with a smartphone under low-light conditions. Each dataset has three different scales with the brightness ranging from low to high. Each scale has $30$ views and captures two brightness levels for a view. The image resolution is $1000\times 750$ and camera poses are estimated from the captures in the high scale using COLMAP~\cite{Johannes2016colmap}. For the stage $1$, we use $25$ views of the middle scale for training and the remaining five views for evaluation. To train the stage $2$, we utilize the corresponding $25$ views from the low and the high scales as the supervision. Also, the remaining views are used for test.

We first evaluate the decomposition stage of our approach on the test images of two scenes. Then, we compare our approach on the novel view synthesis and enhancement task against the state-of-the-art NeRF-W~\cite{martin2021nerf} method. Note that Aleth-NeRF~\cite{cui2023aleth} does not deal with camera sensor noise and cannot process low-light images with heterogeneous brightness. Therefore, it is not included in the comparison.

\subsection{Evaluation of the Decomposition}

Since ground-truth decomposition is unavailable, we present the qualitative decomposition results of a test view from the {\scene{Shrub}} scene and the {\scene{Link}} scene in \cref{fig:decomp_eval}. Our approach achieves a plausible decomposition including illumination, reflectance, and noise. The decomposed results are consistent with the assumption of the Retinex theory, that the illumination component is smooth in textual details and the reflectance component is free of noise since the noise has been separated out. Please refer to the supplementary for the decomposition results on other scenes.

\subsection{Novel View Synthesis and Enhancement}

\begin{figure*}[t]
    \centering
    \hspace{-3.0mm}\rotatebox{90}{\makebox[1.6cm][c]{\small{{\scene{Potter}}}}}\hfill
    \includegraphics[width=\flen]{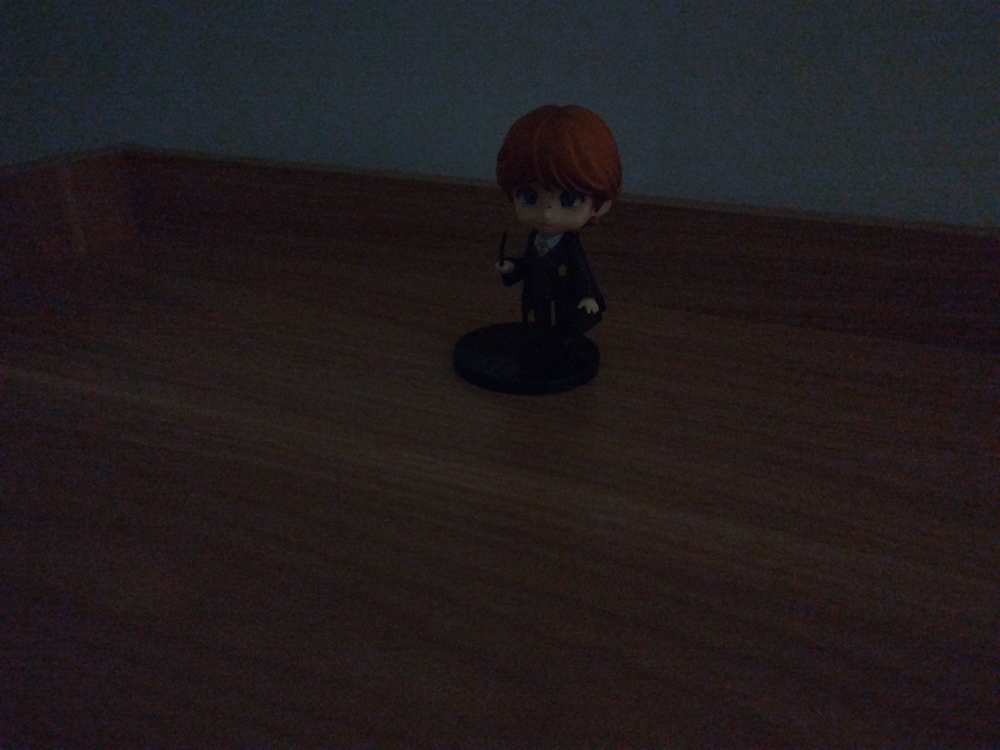}\hfill 
    \includegraphics[width=\flen]{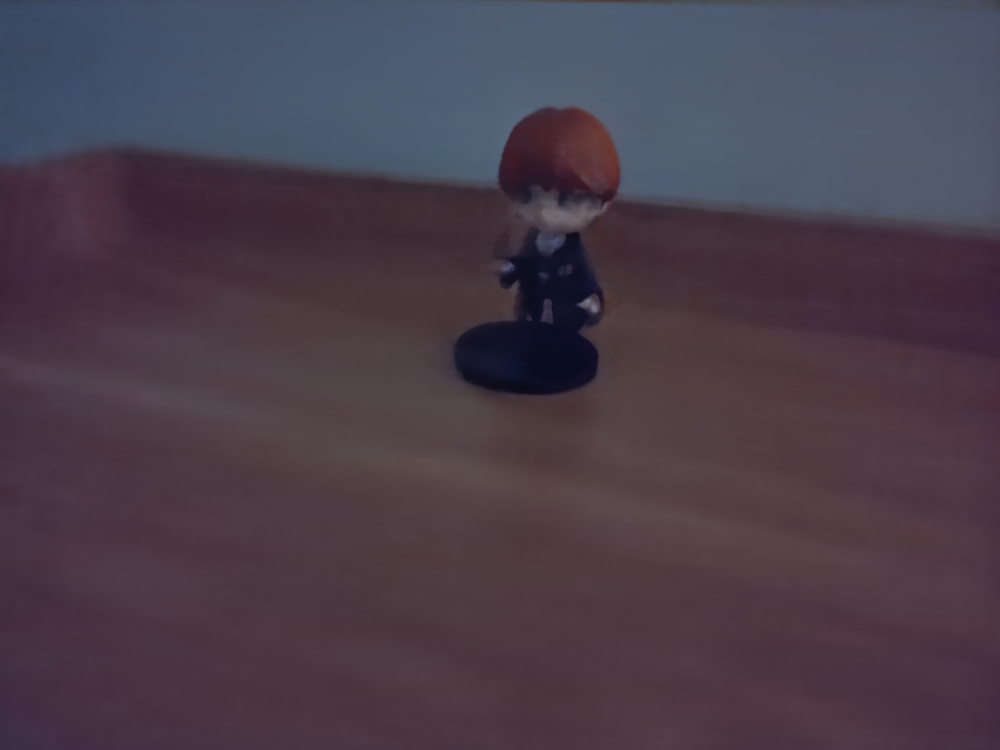}\hfill 
    \includegraphics[width=\flen]{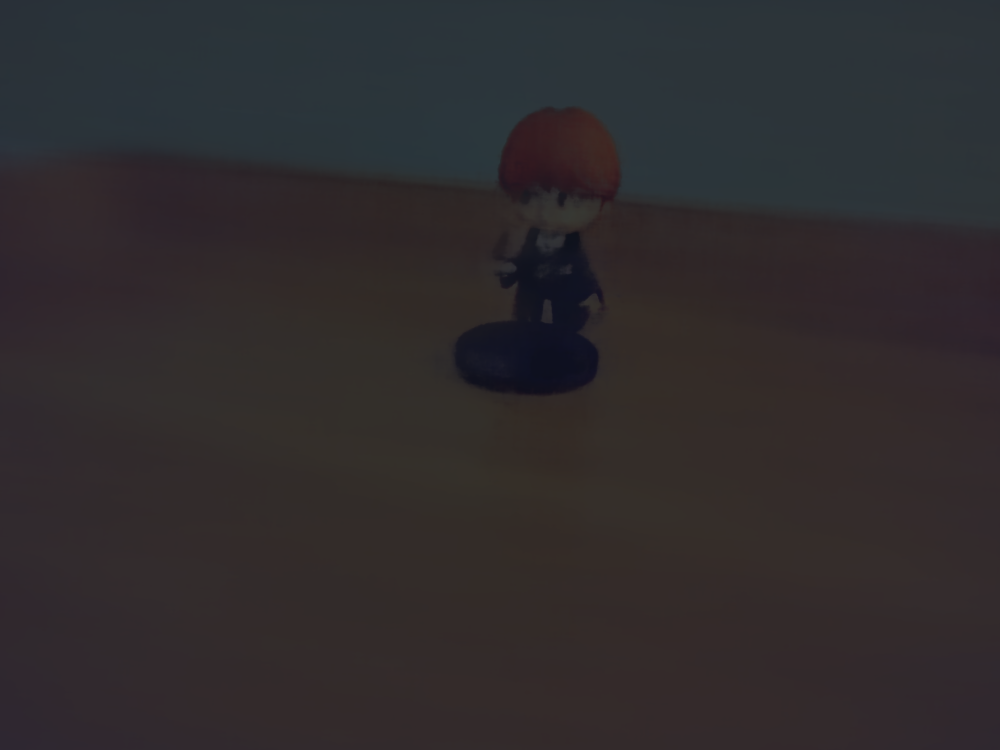}\hfill
    \includegraphics[width=\flen]{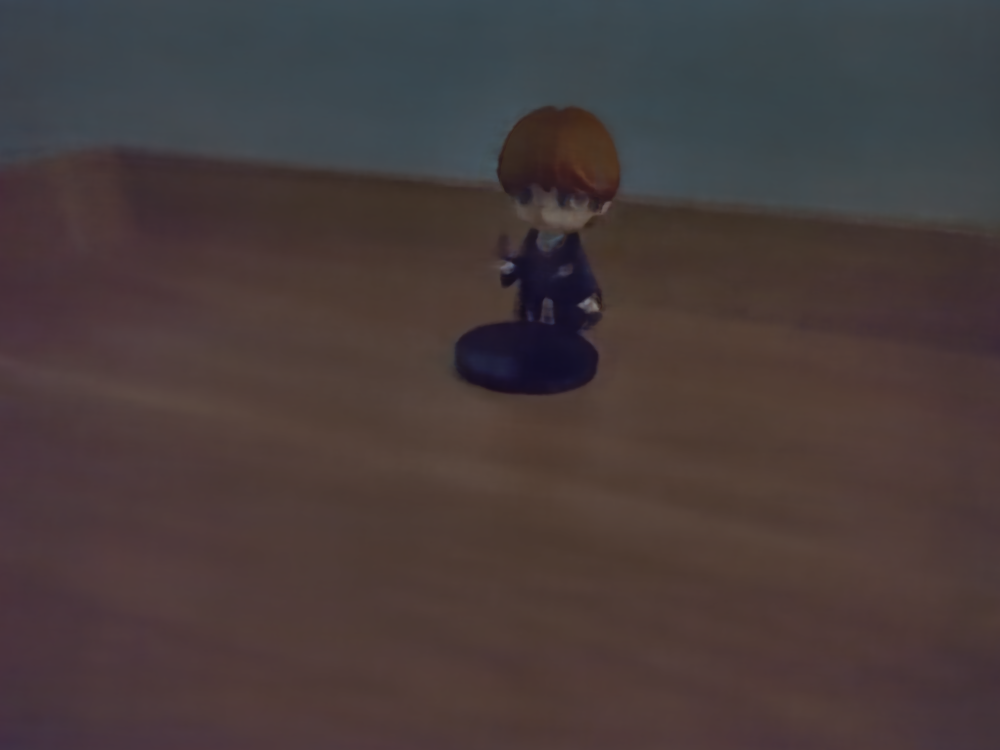}\hfill 
    \includegraphics[width=\flen]{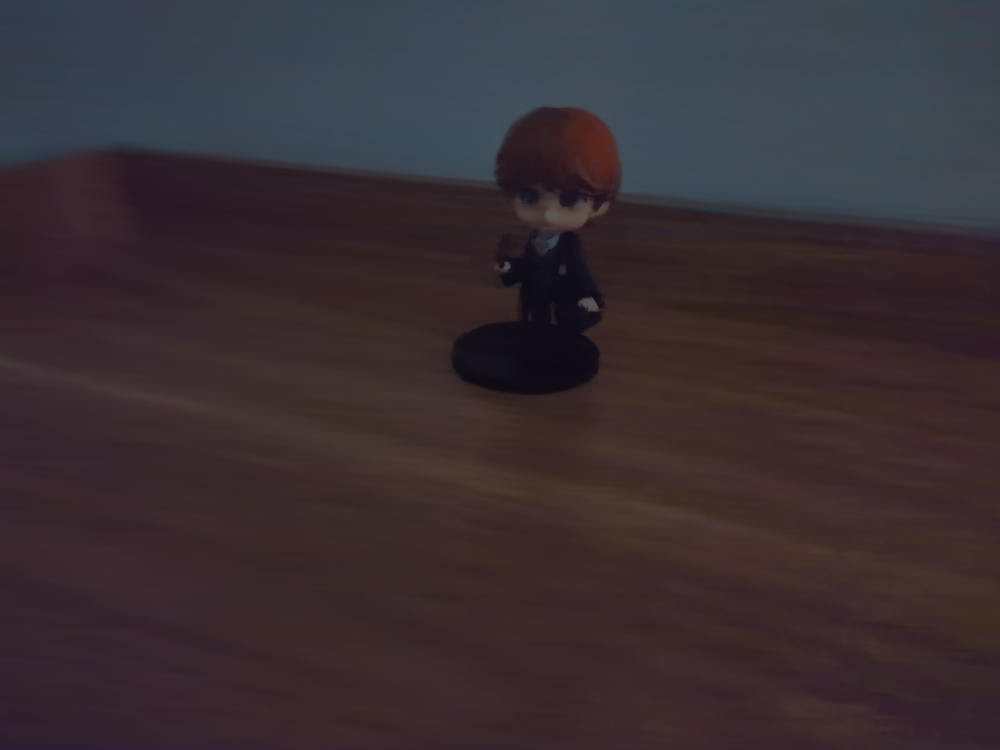}\hfill 
    \includegraphics[width=\flen]{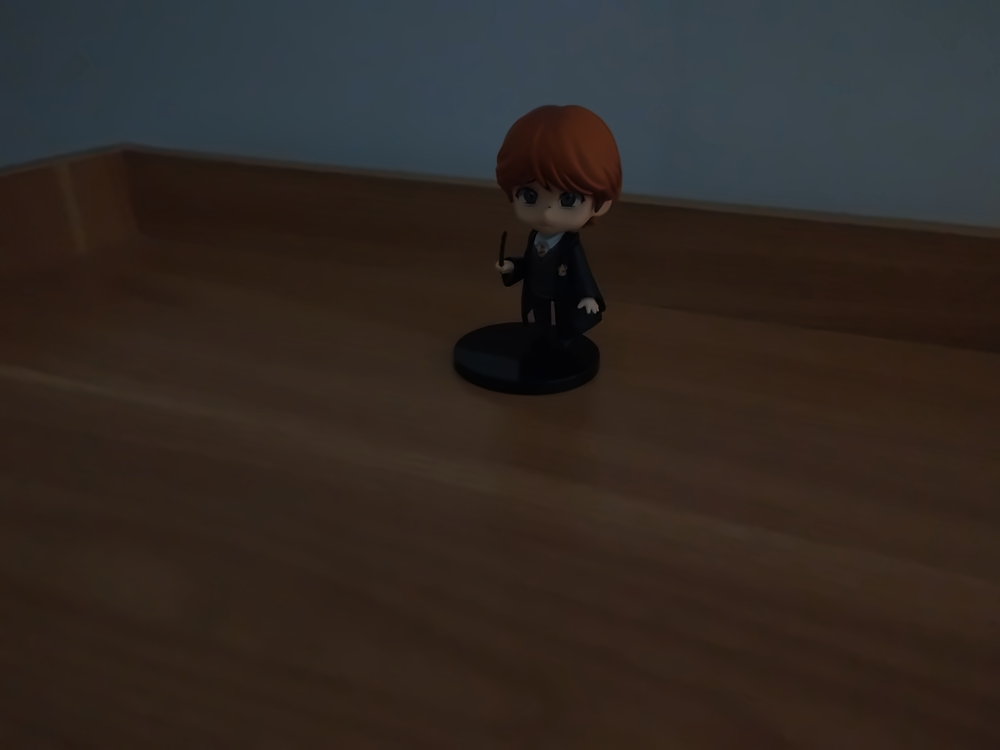}\\
    \hspace{-3.0mm}\rotatebox{90}{\makebox[1.6cm][c]{\small{{\scene{Plant}}}}}\hfill
    \includegraphics[width=\flen]{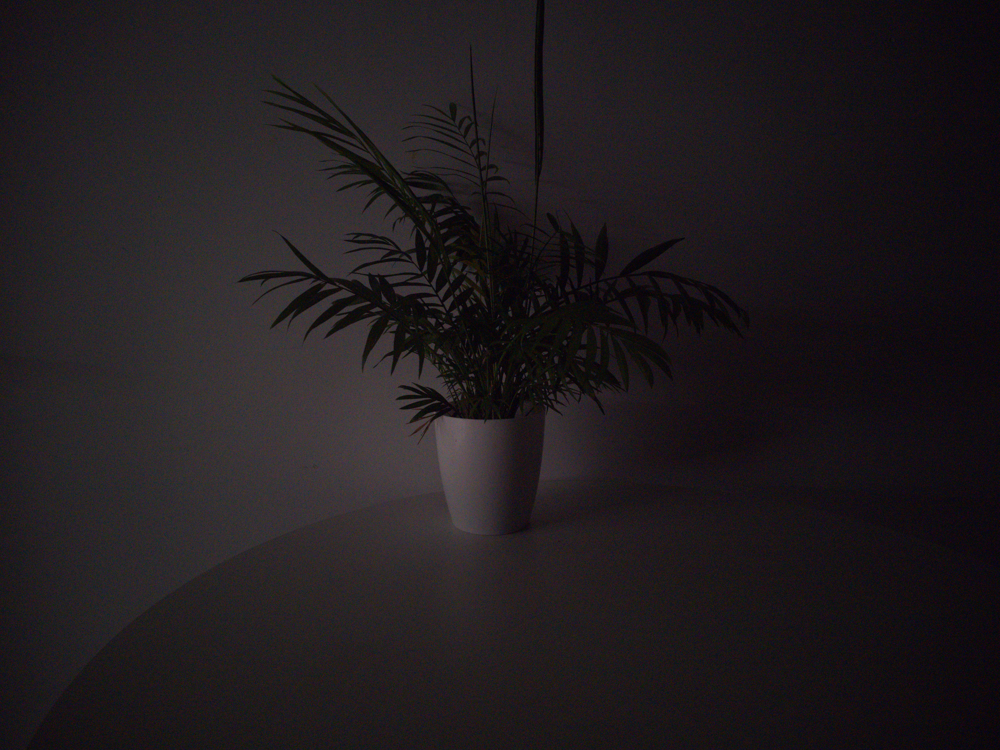}\hfill 
    \includegraphics[width=\flen]{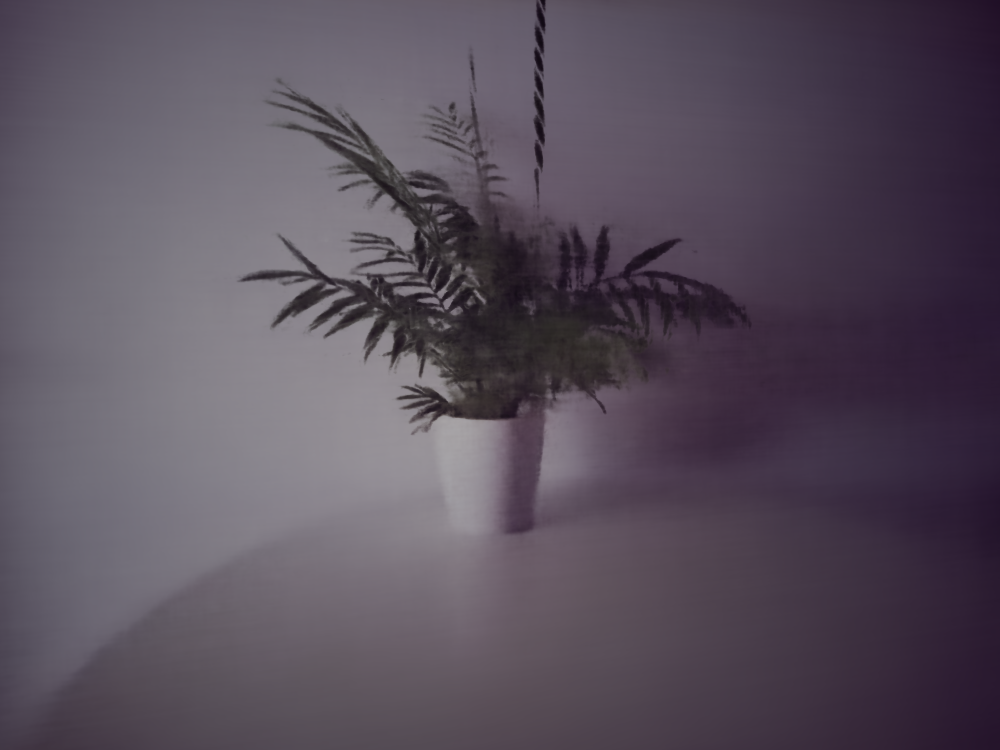}\hfill 
    \includegraphics[width=\flen]{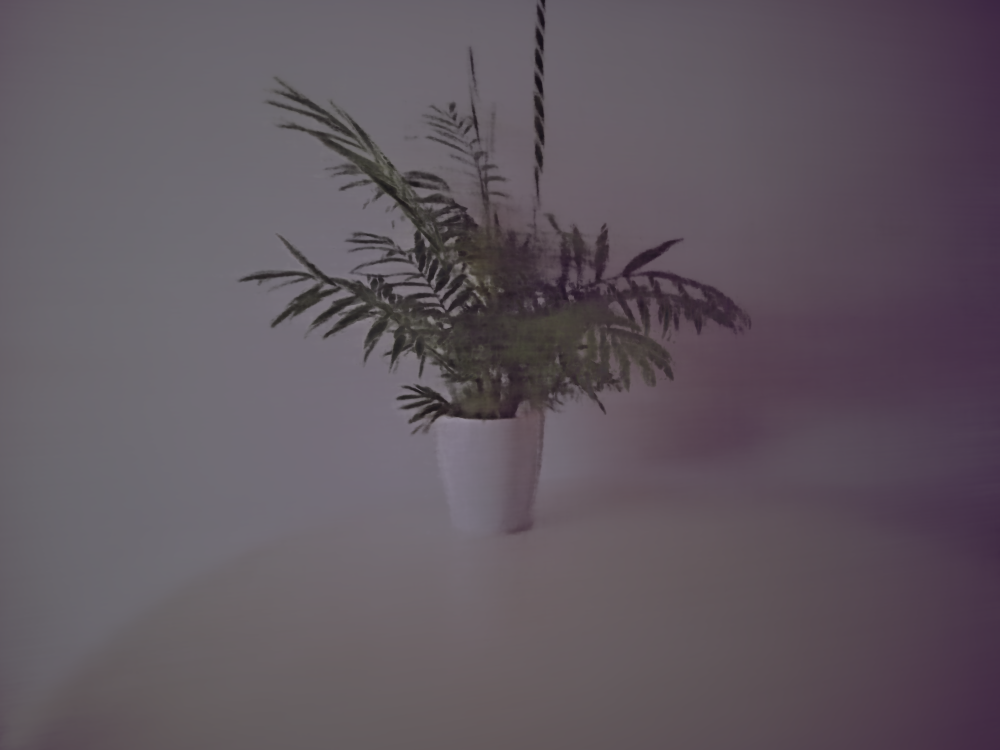}\hfill
    \includegraphics[width=\flen]{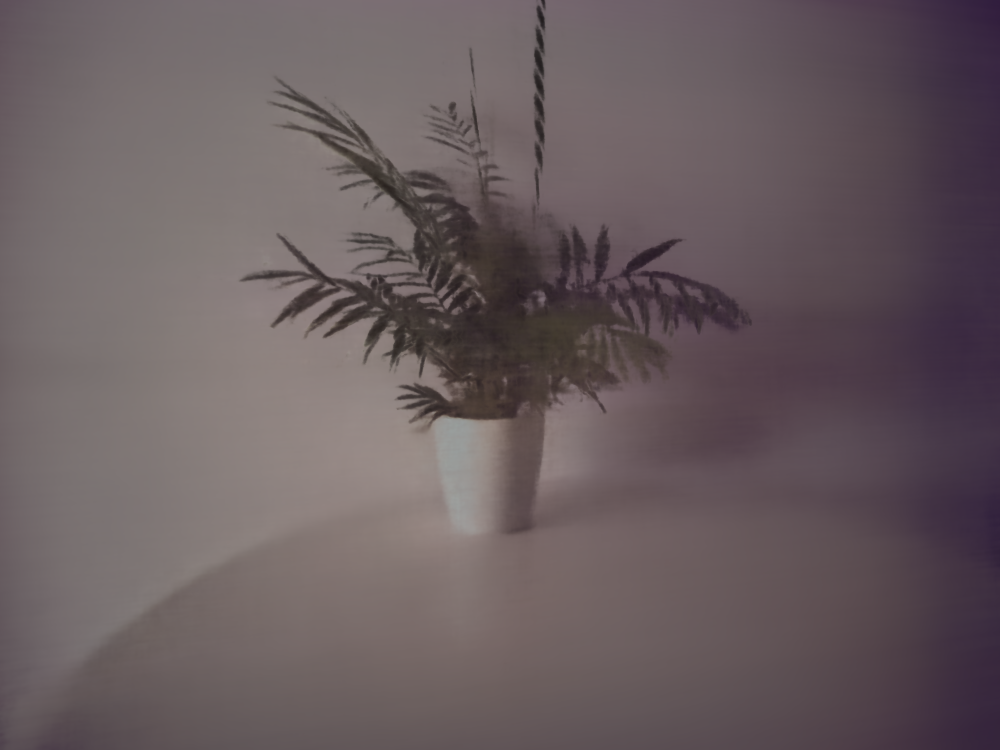}\hfill 
    \includegraphics[width=\flen]{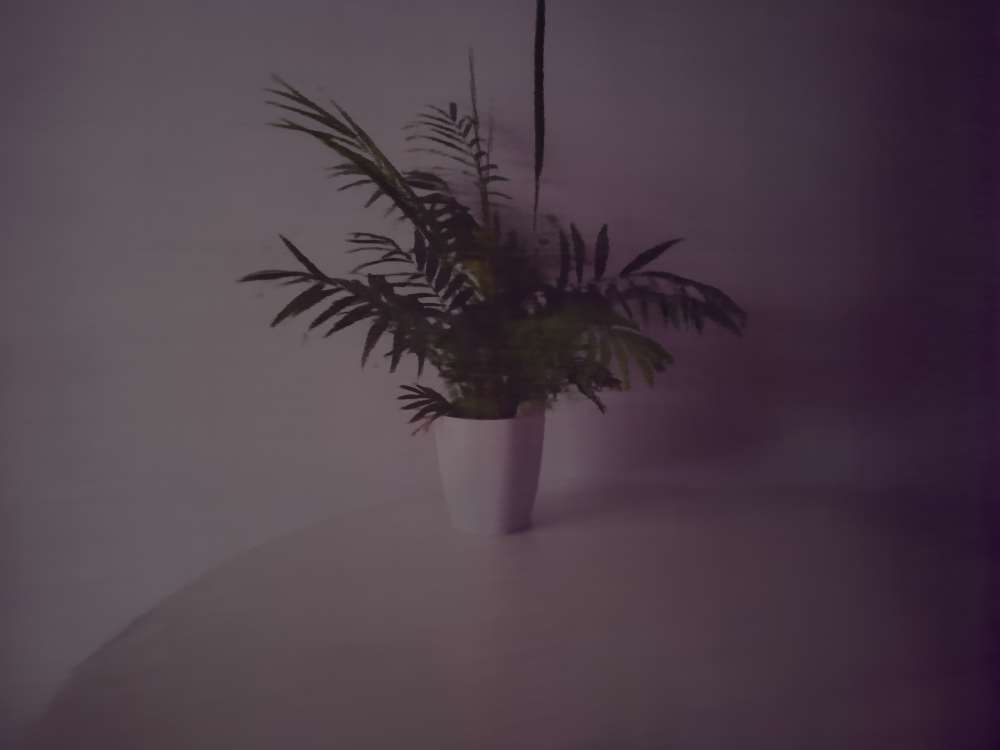}\hfill 
    \includegraphics[width=\flen]{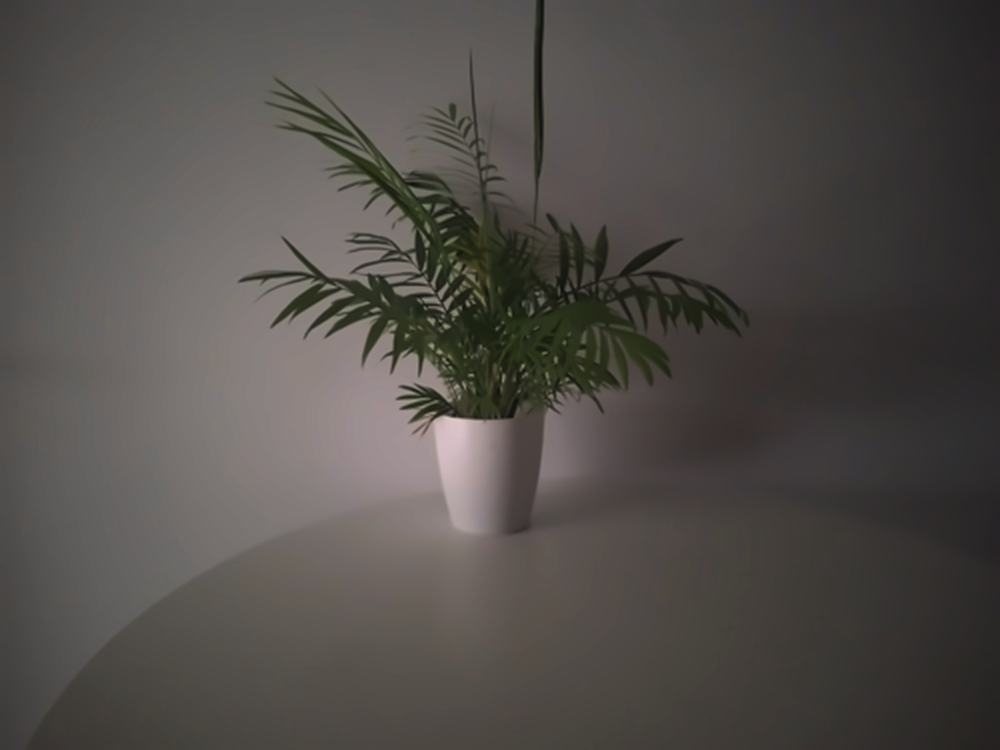}\\
    \hspace{-3.0mm}\rotatebox{90}{\makebox[1.6cm][c]{\small{{\scene{Shrub}}}}}\hfill
    \includegraphics[width=\flen]{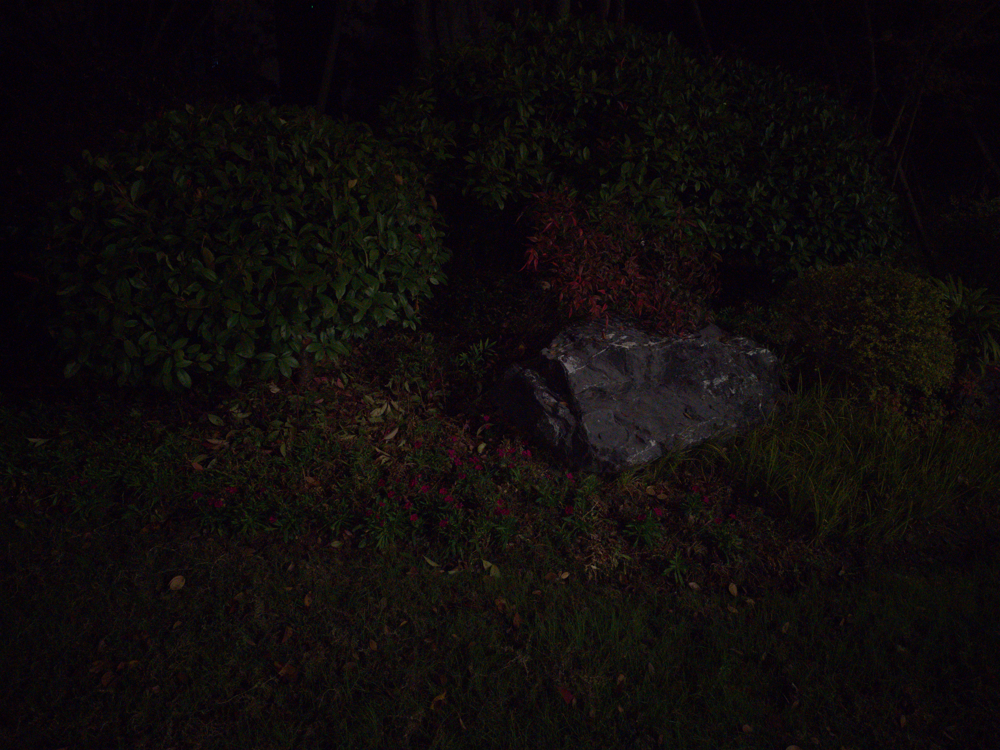}\hfill 
    \includegraphics[width=\flen]{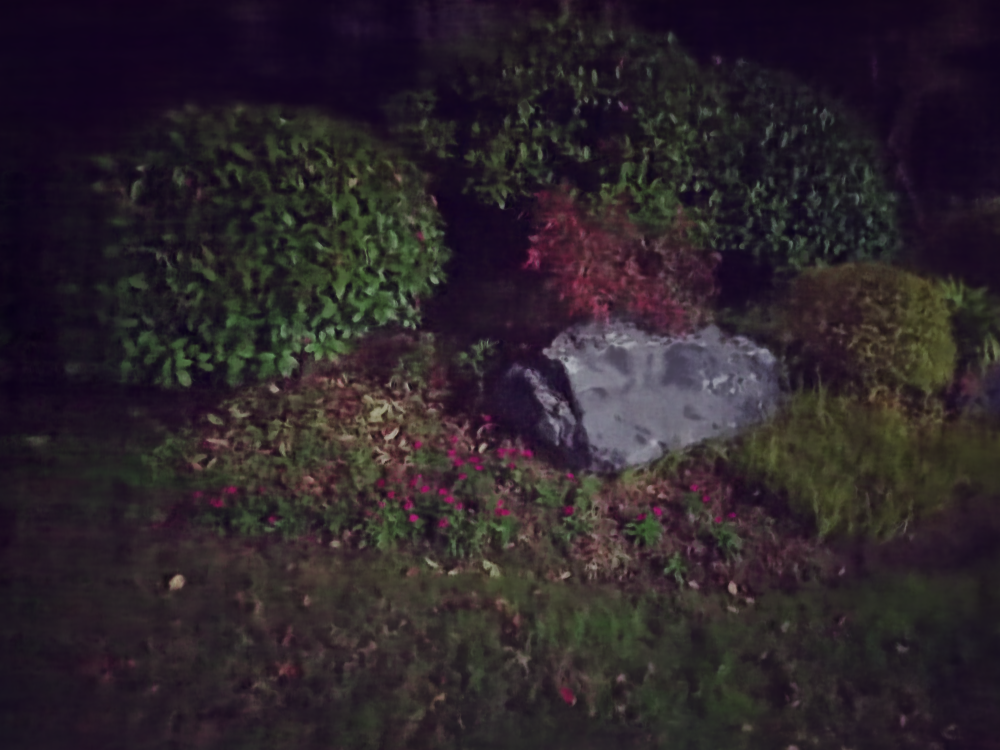}\hfill 
    \includegraphics[width=\flen]{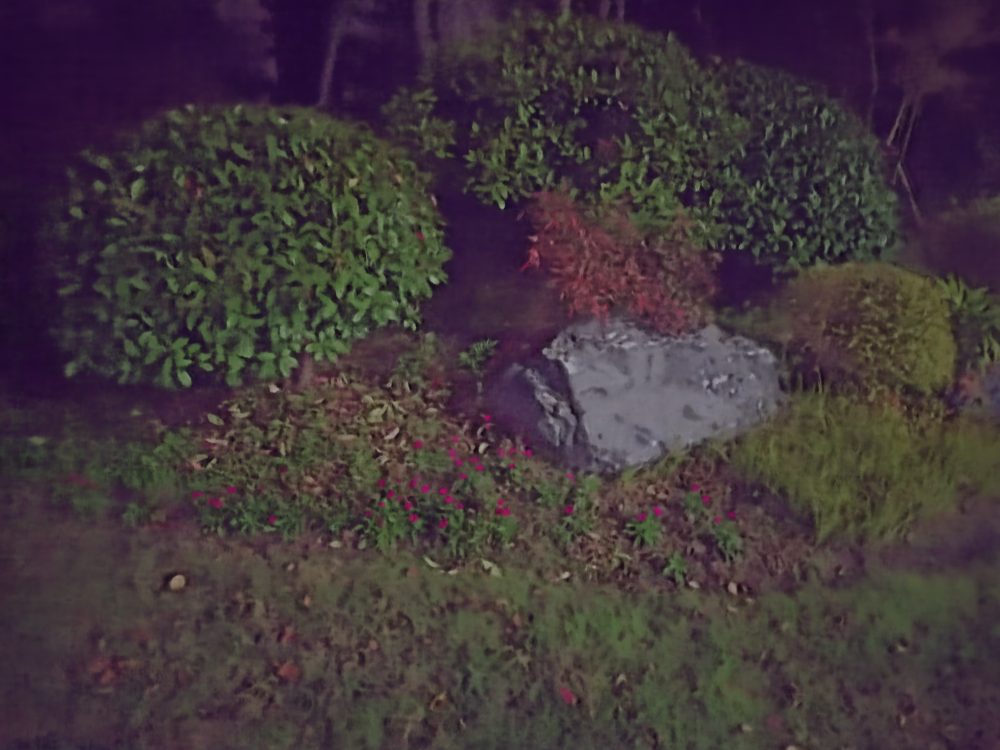}\hfill
    \includegraphics[width=\flen]{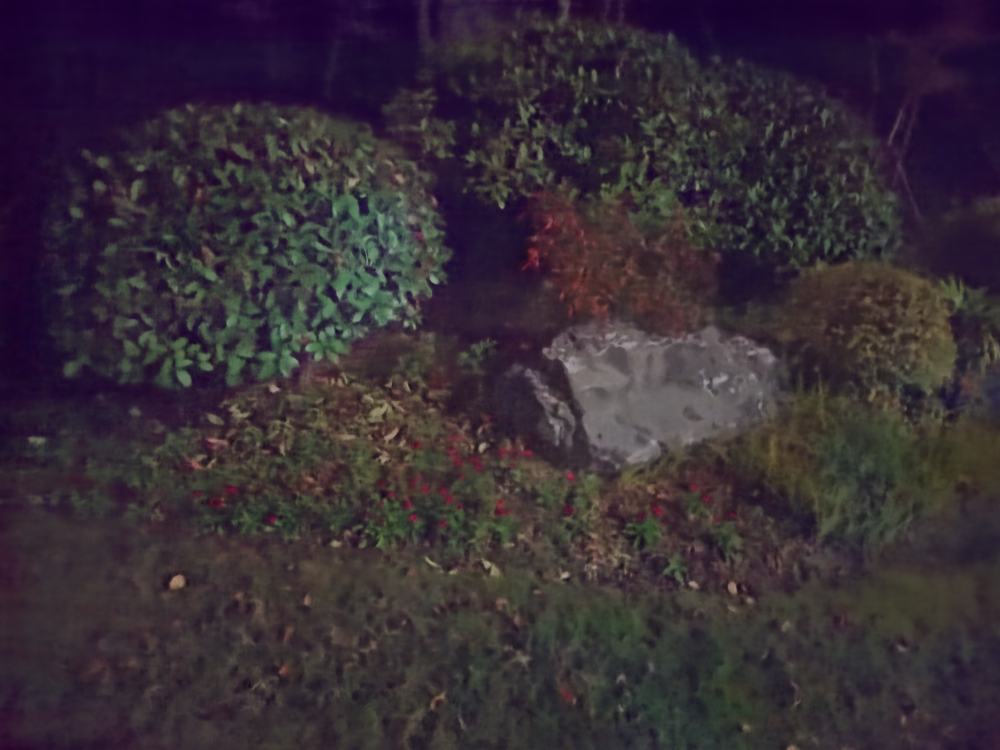}\hfill 
    \includegraphics[width=\flen]{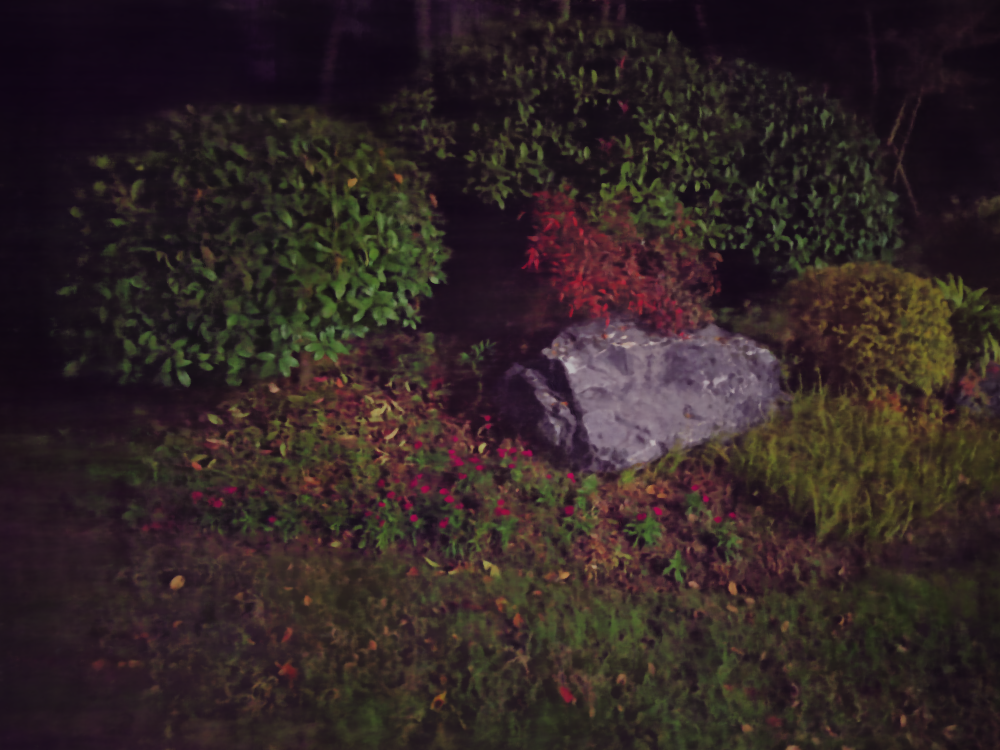}\hfill 
    \includegraphics[width=\flen]{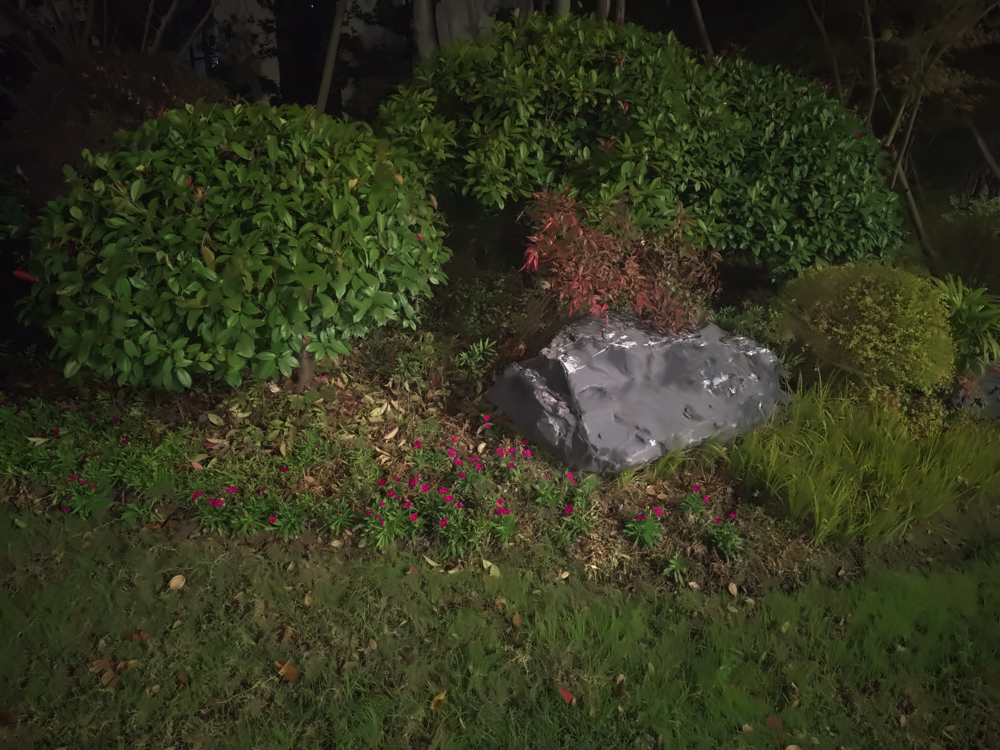}\\
    \hspace{-3.0mm}\rotatebox{90}{\makebox[1.6cm][c]{\small{{\scene{Link}}}}}\hfill
    \includegraphics[width=\flen]{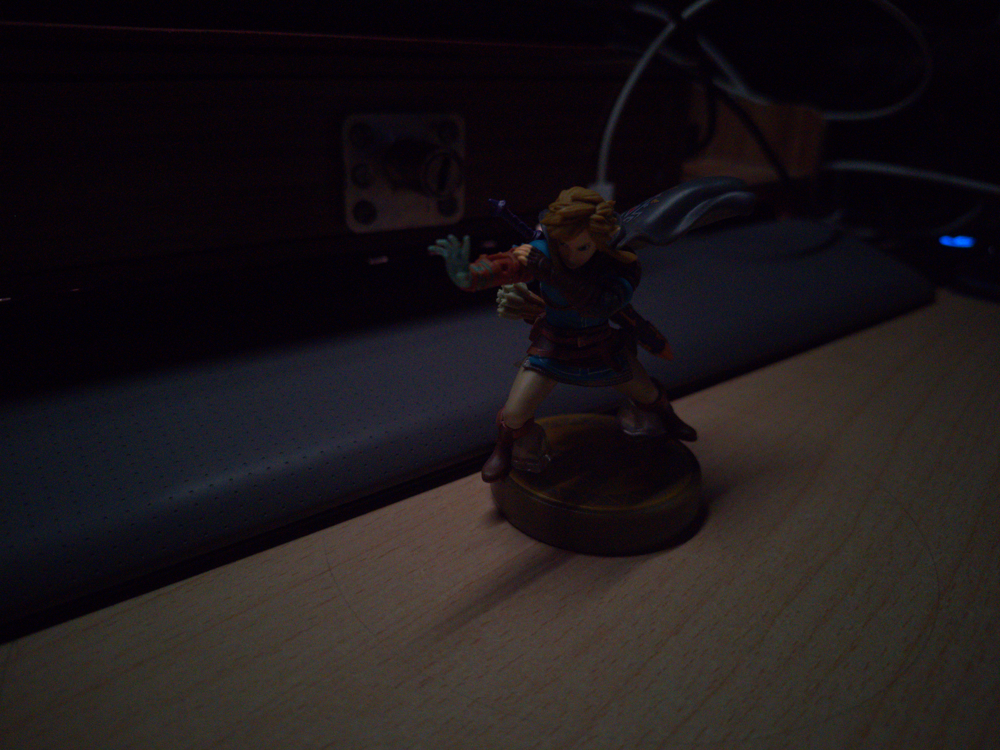}\hfill 
    \includegraphics[width=\flen]{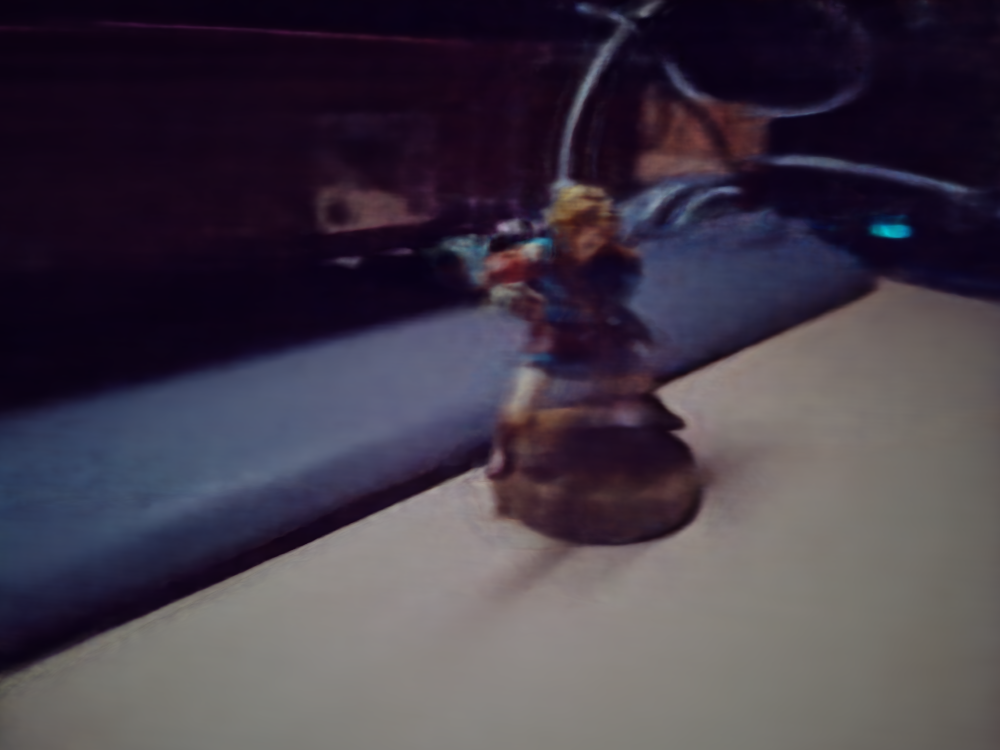}\hfill 
    \includegraphics[width=\flen]{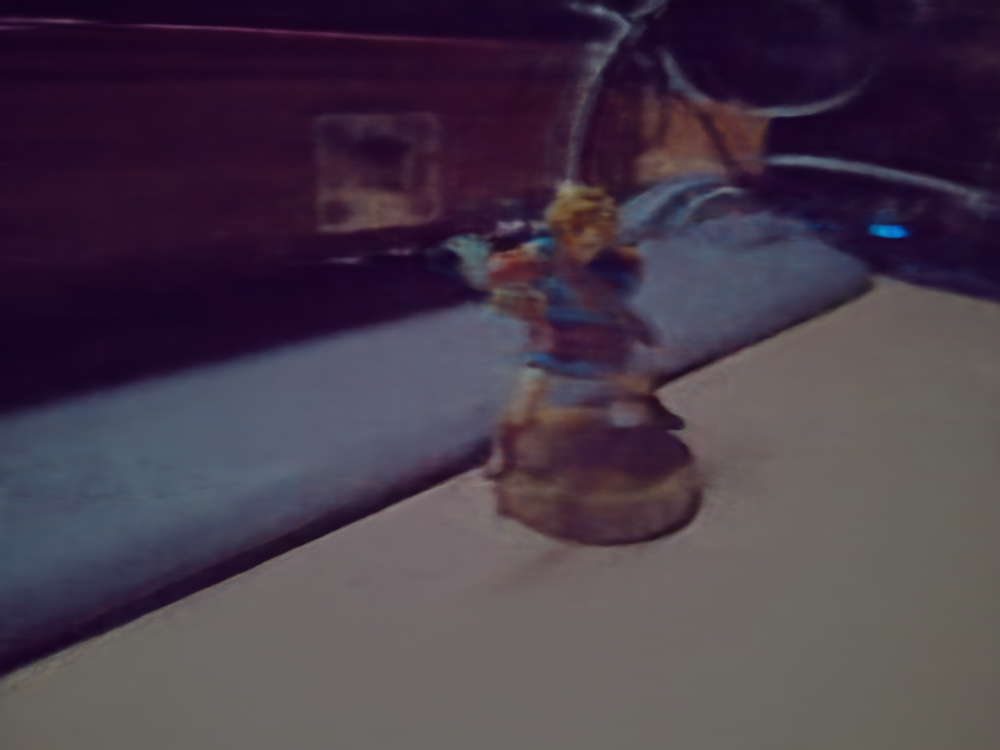}\hfill
    \includegraphics[width=\flen]{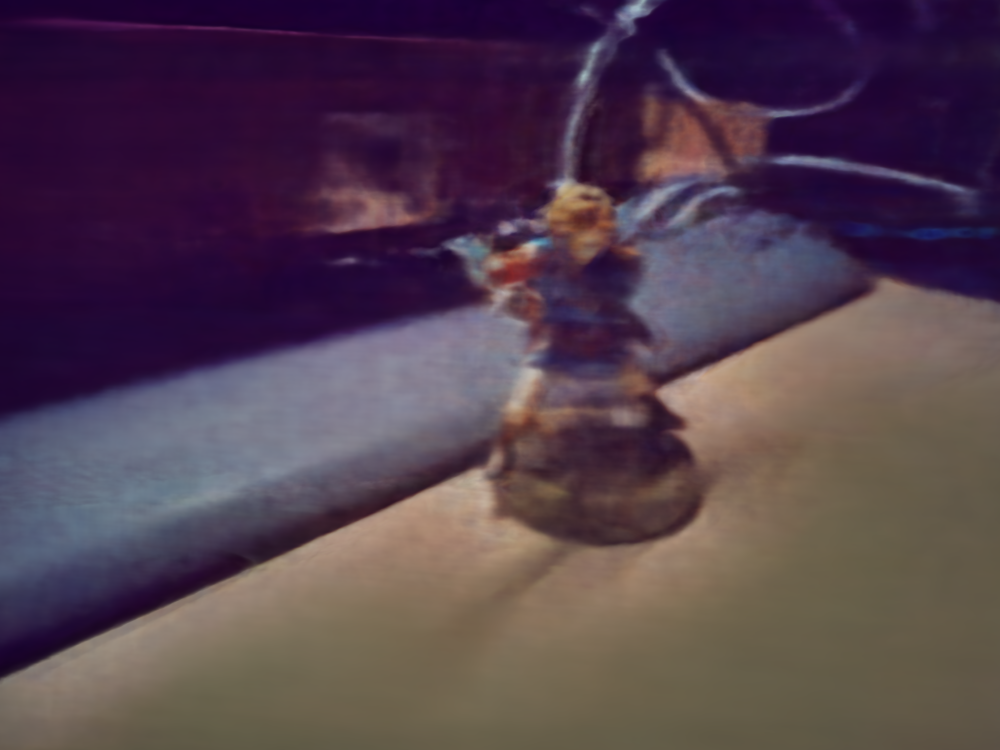}\hfill 
    \includegraphics[width=\flen]{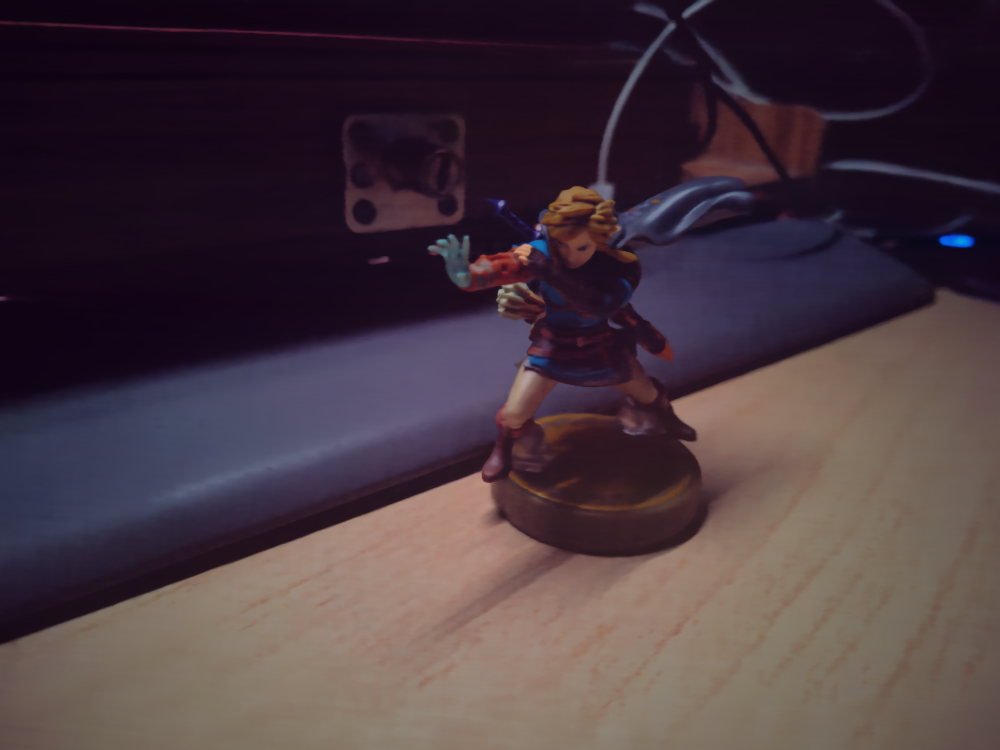}\hfill 
    \includegraphics[width=\flen]{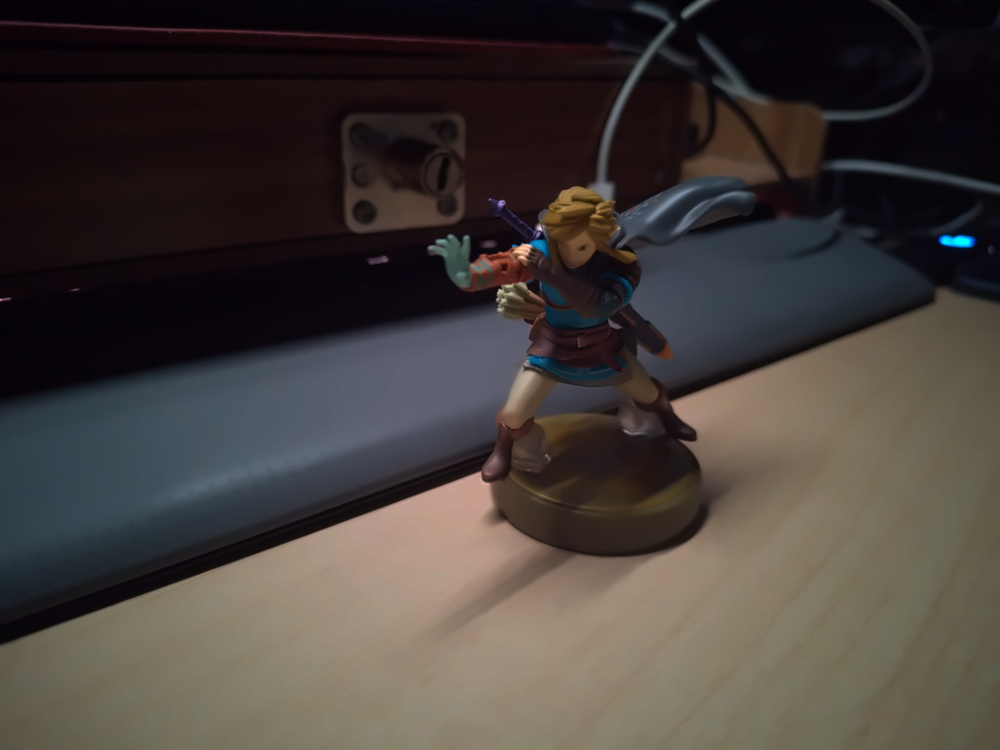}\\
    \makebox[\flen][c]{\small{Test input}}\hfill
    \makebox[\flen][c]{\small{SCI+NeRF-W}}\hfill
    \makebox[\flen][c]{\small{DCE+NeRF-W}}\hfill
    \makebox[\flen][c]{\small{EnGAN+NeRF-W}}\hfill
    \makebox[\flen][c]{\small{\textbf{Ours}}}\hfill
    \makebox[\flen][c]{\small{Reference}}\hfill\\
    \caption{\label{fig:compare_enhance}View synthesis and enhancement comparisons between our approach and the SCI+NeRF-W, DCE+NeRF-W, and EnGAN+NeRF-W methods on the same test views of four scenes. The reference images are from the high scale and they are pre-processed by a non-local mean denoiser.}
\end{figure*}

In the experiment, we compare the novel view synthesis and enhancement results between our approach and the NeRF-W~\cite{martin2021nerf} method on five scenes. Note that NeRF-W has no designs to process the camera sensor noise of low-light images and cannot synthesize novel views with an enhanced brightness level beyond the training views. Thus, for the NeRF-W, we incorporate the non-local mean method of OpenCV as the pre-denoiser and leverage three state-of-the-art 2D low-light enhancement methods SCI~\cite{ma2022toward}, DCE~\cite{guo2020zero}, and EnGAN~\cite{jiang2021enlightengan} for pre-enhancing the training images. 
Our approach decomposes a test view into three components and enhances the illumination component in stage 2. The qualitative comparisons on the enhanced test views of four scenes are presented in \cref{fig:compare_enhance}. \cref{tab:metric1} presents the corresponding quantitative metrics in terms of PSNR, SSIM, and LPIPS. Our approach produces enhanced novel views with both higher visual quality and better numerical metrics than the compared methods. Note that the results of the NeRF-W suffer from severe contour artifacts and loss of details, whereas our approach successfully reconstructs more geometric and textural details. 
\begin{table*}[th]
    \centering
		\small
		\caption{\label{tab:metric1} Comparisons of the average quantitative metrics on the enhanced test views of four scenes. The best metrics are in bold.}
		\setlength{\tabcolsep}{4pt}
		\renewcommand{\arraystretch}{0.9}
		\begin{tabular}{l ccc ccc ccc ccc}
			\toprule
			{}&  \multicolumn{3}{c}{Potter} & \multicolumn{3}{c}{Plant}&\multicolumn{3}{c}{Shrub}&\multicolumn{3}{c}{Link}\\
			\cmidrule(lr){2-4}\cmidrule(lr){5-7}\cmidrule(lr){8-10}\cmidrule(lr){11-13}
			Method & PSNR$\uparrow$ & SSIM$\uparrow$ &LPIPS$\downarrow$ & PSNR$\uparrow$  & SSIM$\uparrow$ &LPIPS$\downarrow$&PSNR$\uparrow$  & SSIM$\uparrow$ &LPIPS$\downarrow$&PSNR$\uparrow$  & SSIM$\uparrow$ &LPIPS$\downarrow$\\
			\midrule
			{SCI+NeRF-W}& 20.539&  0.868& 0.188& 22.299& 0.902& 0.227& 22.121&  0.641& 0.568& 20.659& 0.831&0.382\\
			{DCE+NeRF-W}& 25.063&  0.938& 0.164& 18.978& 0.848& 0.221& 18.808&  0.570& 0.513& 19.091& 0.774&0.427\\
			{EnGAN+NeRF-W}& 21.568&  0.883& 0.166& 18.354& 0.841& 0.208& 19.931& 0.591 &  0.566& 17.243& 0.741&0.457\\
			{\bf{Ours}}& \bf{27.393}&  \bf{0.952}& \bf{0.152}& \bf{22.946}& \bf{0.917}& \bf{0.216}& \bf{23.436}&  \bf{0.701}& \bf{0.443}& \bf{26.442}& \bf{0.901}&\bf{0.184}\\
			\bottomrule
		\end{tabular}
\end{table*}

\begin{table}[ht]
    \centering
    \small
    \caption{Average quantitative results on the test images of the \scene{Potter} scene for the ablations of loss terms. The best metrics (including PSNR, SSIM, and LPIPS) are shown in bold.}\label{tab:loss_ablate}
    \setlength{\tabcolsep}{3pt}
    \renewcommand{\arraystretch}{0.9}
    \begin{tabular}{cccc}
        \toprule
        Ablation & PSNR$\uparrow$ & SSIM$\uparrow$ & LPIPS$\downarrow$\\
        \midrule
        \small{w.o. $\Vert L-L_{0}\Vert_{1}$} &    35.769&   0.973&  0.087\\
        \small{w.o. $\Vert w\cdot\left(\nabla L\right)\Vert_1$} & 35.890   & 0.967  &0.070  \\
        \small{w.o. $\Vert R- S / L\Vert_{1}$} &36.099  & 0.971 & 0.071\\ 
        \small{w.o. $\Vert S\cdot N\Vert_{F}$} & 35.627& 0.967&0.071 \\
        \small{w.o. $V\left[M\left(\zeta_{\{K\}}\right)\right]$} &35.624 &0.963  &0.074 \\
        \midrule
        \bf{Ours} & \bf{36.864}& \bf{0.975} &\bf{0.065} \\
        \bottomrule
    \end{tabular}
\end{table}

\subsection{Generalization of Illumination Adjustment}

\begin{figure*}[th]
    \centering
    \hspace{-3.0mm}\rotatebox{90}{\makebox[1.5cm][c]{\small{{\scene{Link}}}}}\hfill
    \includegraphics[width=\dlen]{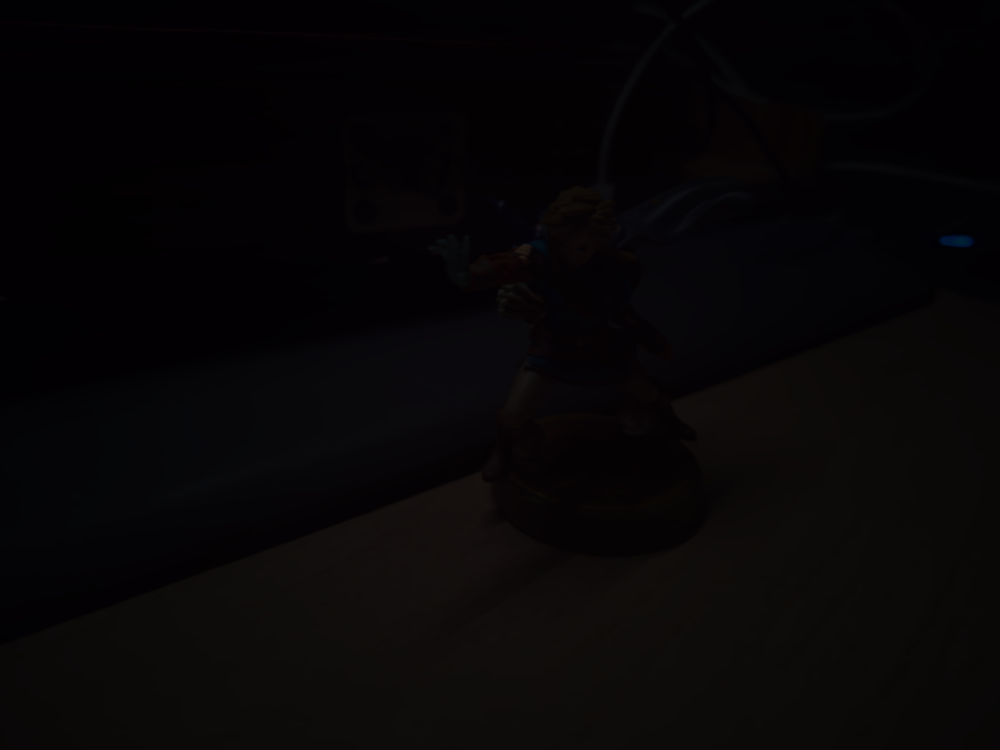}\hfill 
    \includegraphics[width=\dlen]{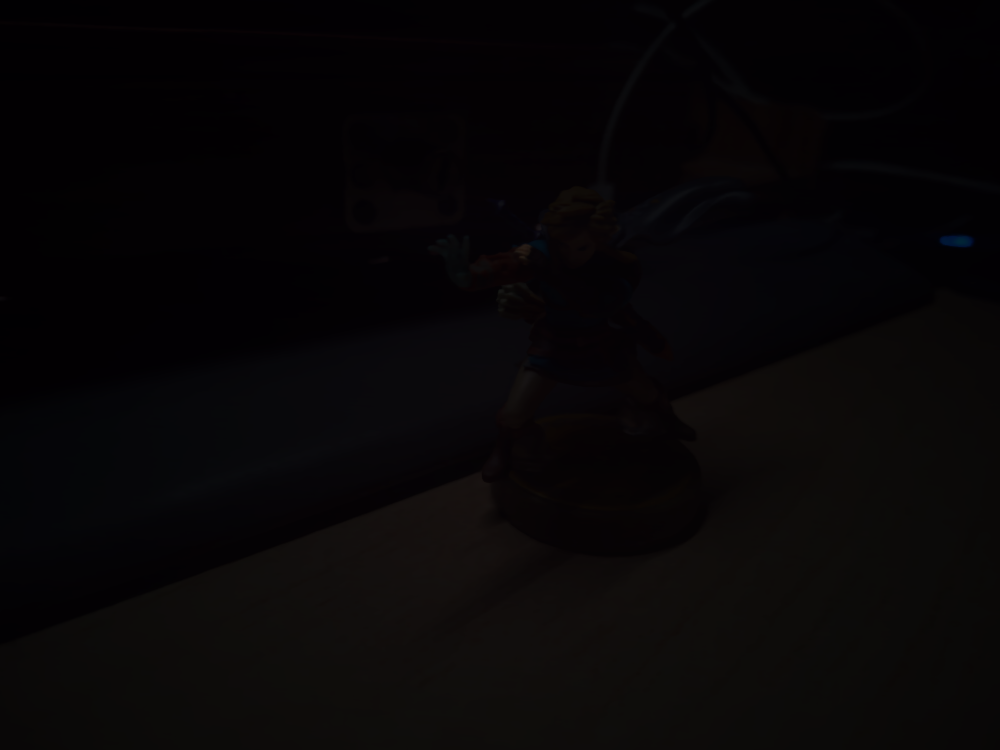}\hfill 
    \includegraphics[width=\dlen]{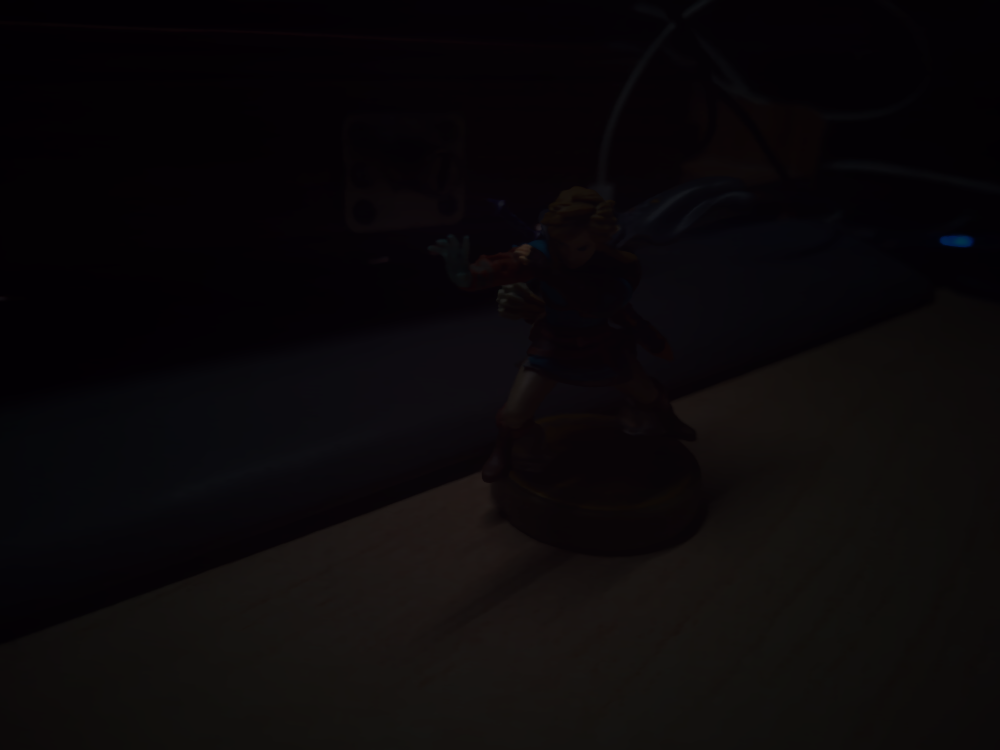}\hfill
    \includegraphics[width=\dlen]{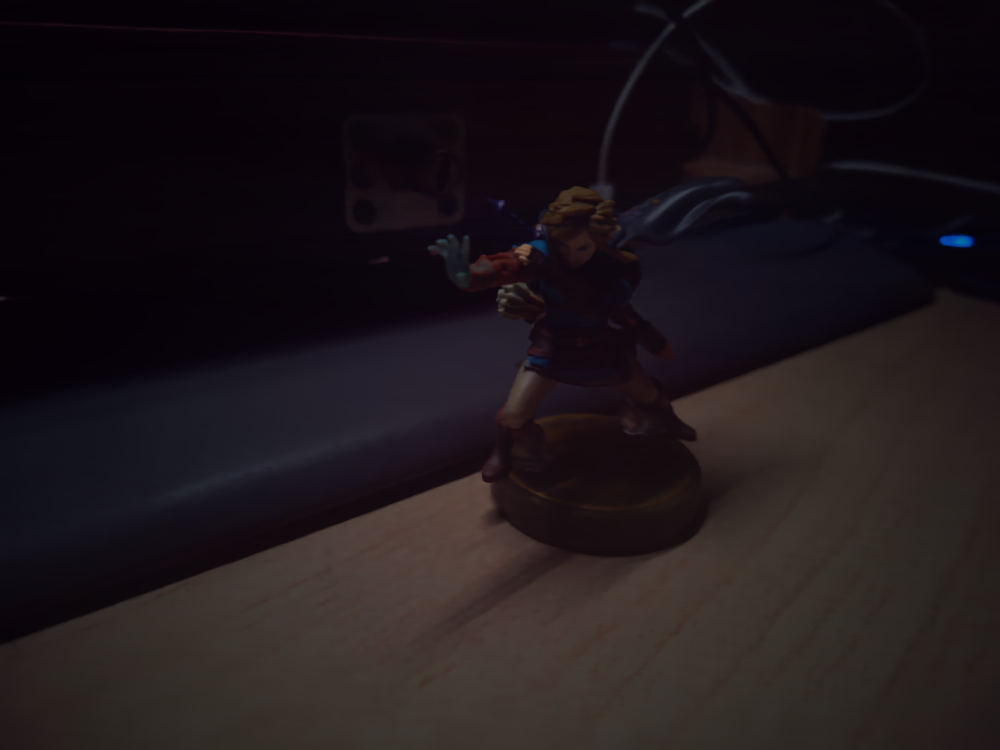}\hfill
    \includegraphics[width=\dlen]{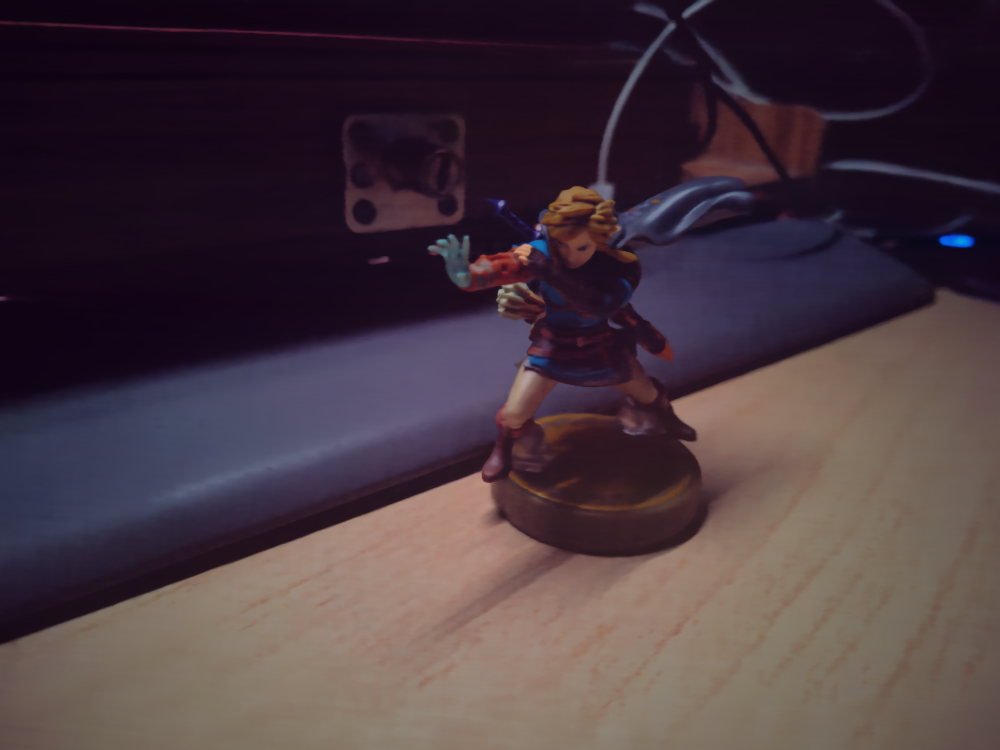}\hfill 
    \includegraphics[width=\dlen]{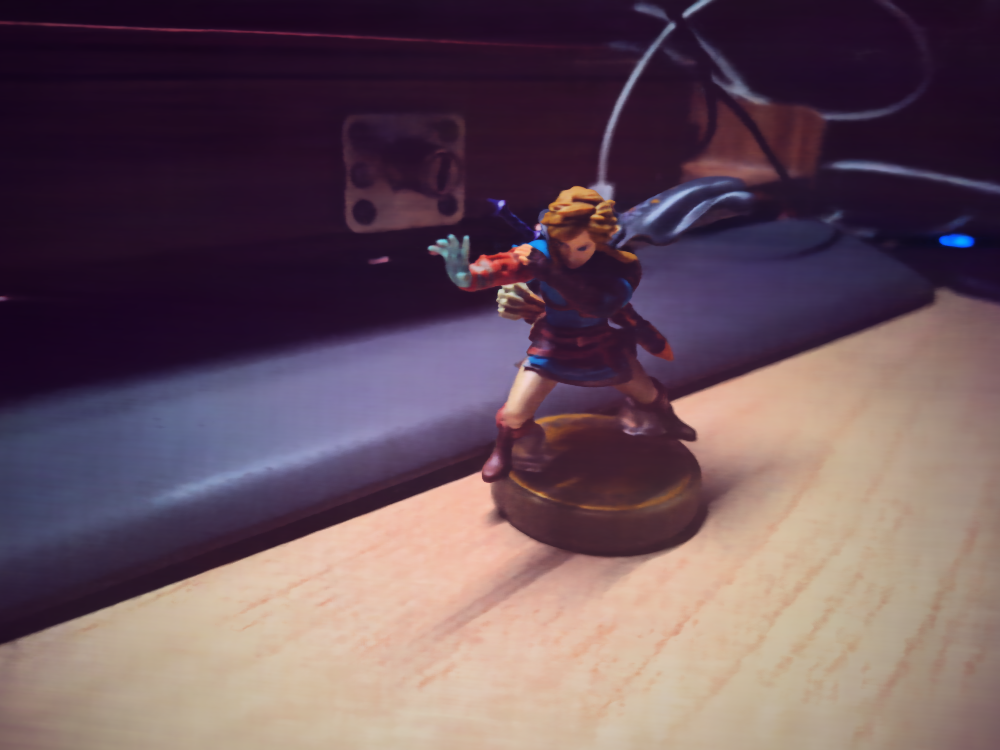}\hfill 
    \includegraphics[width=\dlen]{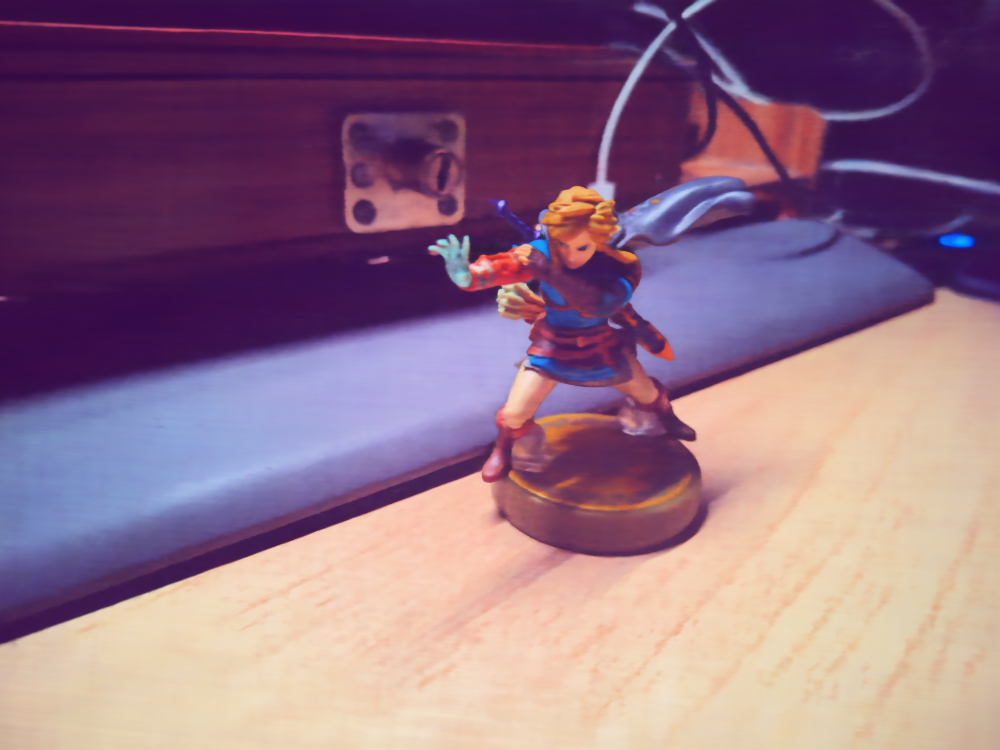}\\
    \hspace{-3.5mm}\rotatebox{90}{\makebox[1.5cm][c]{\small{{\scene{Link (L)}}}}}\hfill
    \includegraphics[width=\dlen]{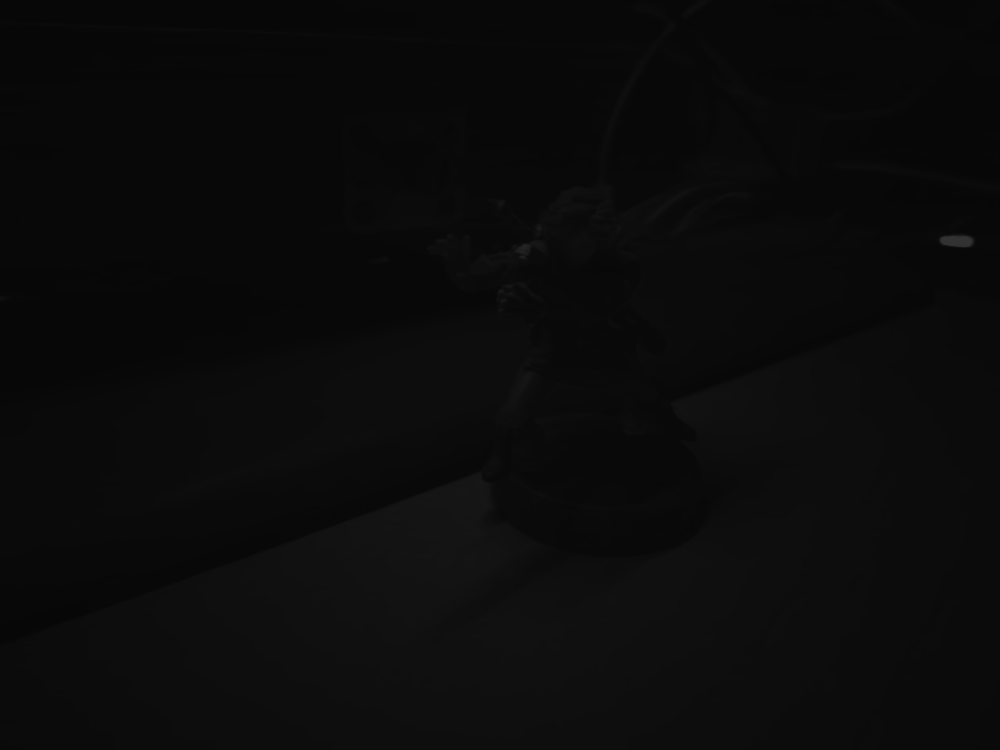}\hfill 
    \includegraphics[width=\dlen]{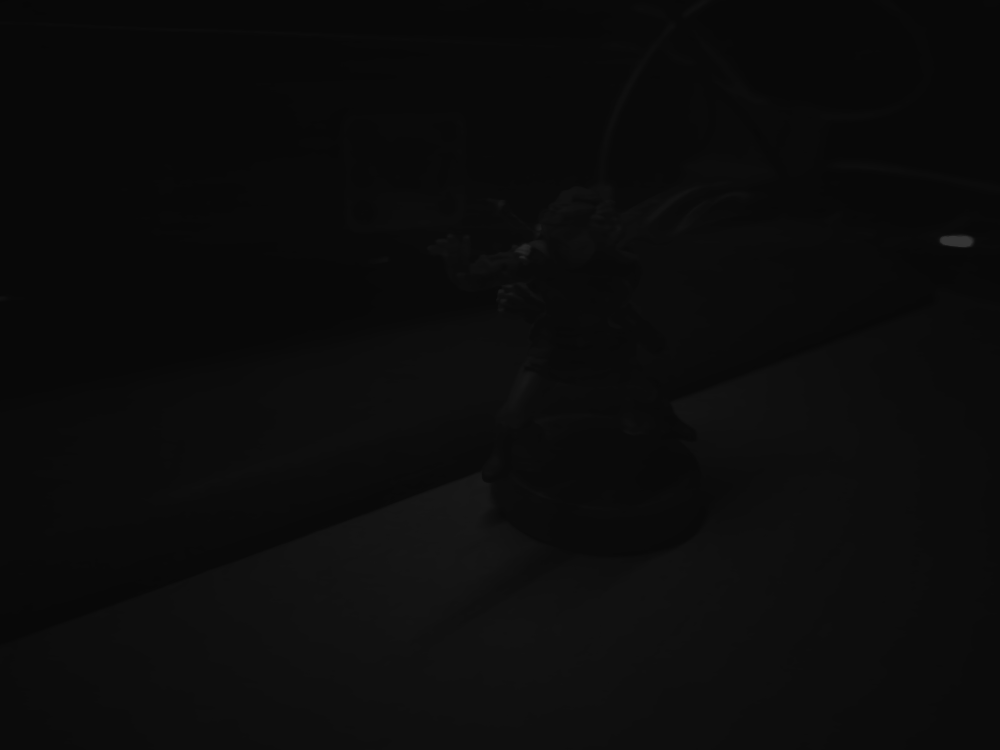}\hfill
    \includegraphics[width=\dlen]{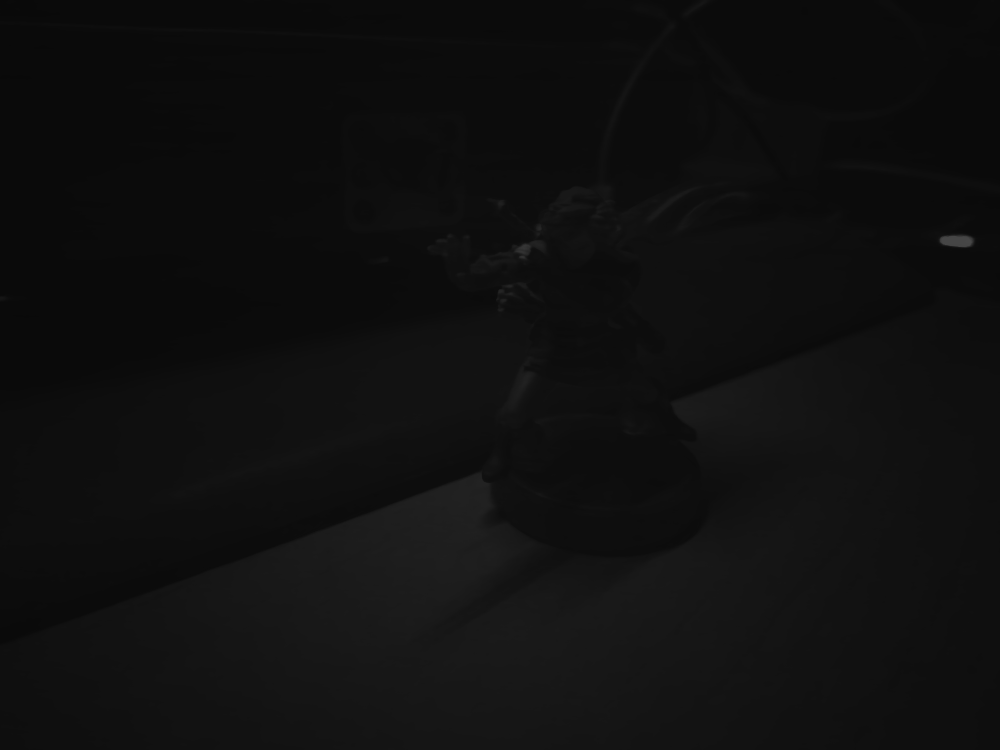}\hfill
    \includegraphics[width=\dlen]{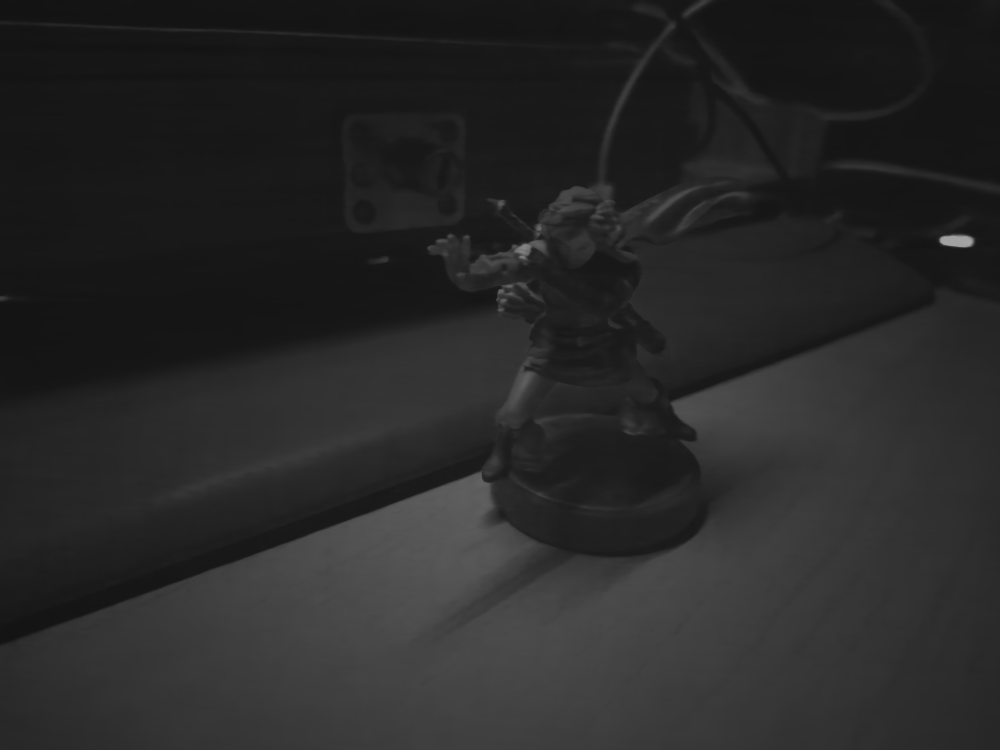}\hfill
    \includegraphics[width=\dlen]{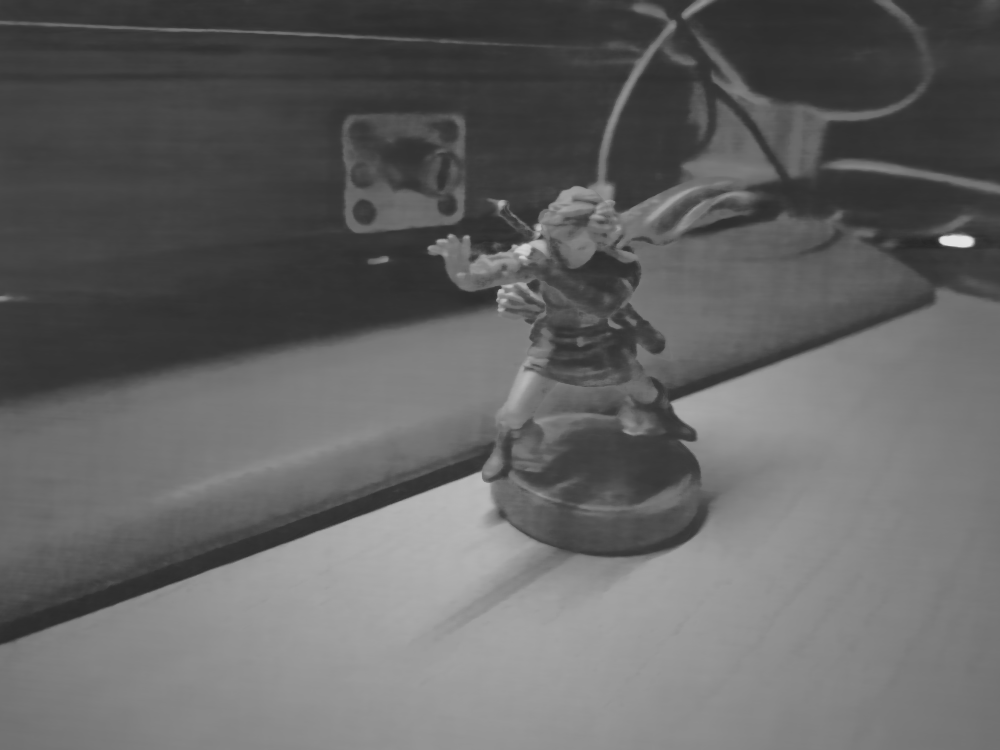}\hfill 
    \includegraphics[width=\dlen]{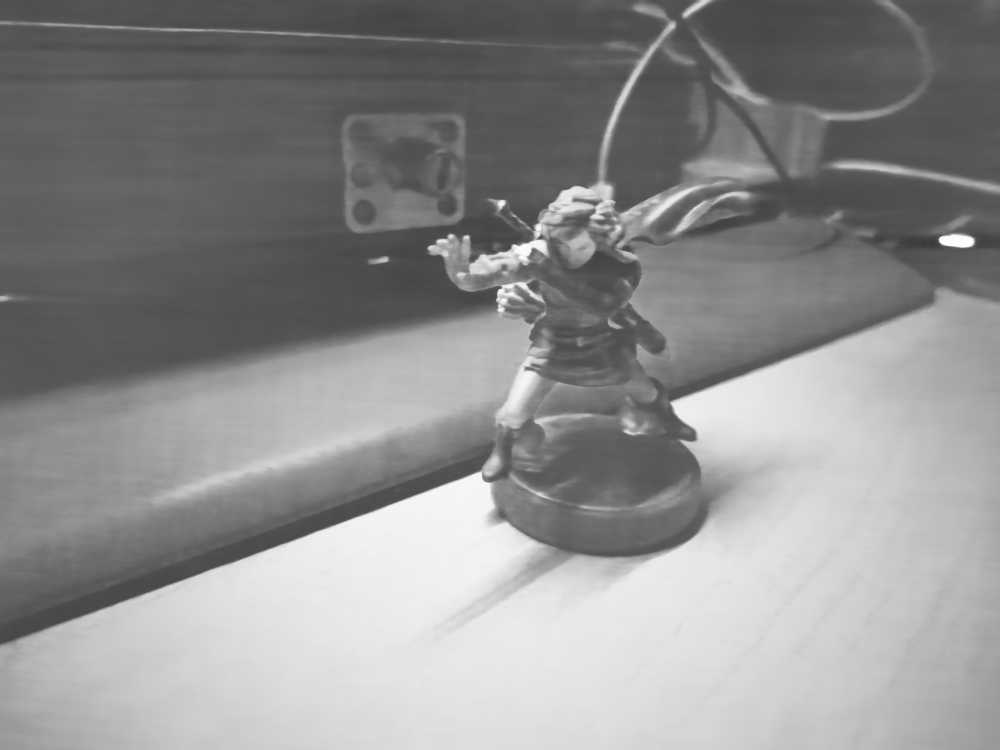}\hfill 
    \includegraphics[width=\dlen]{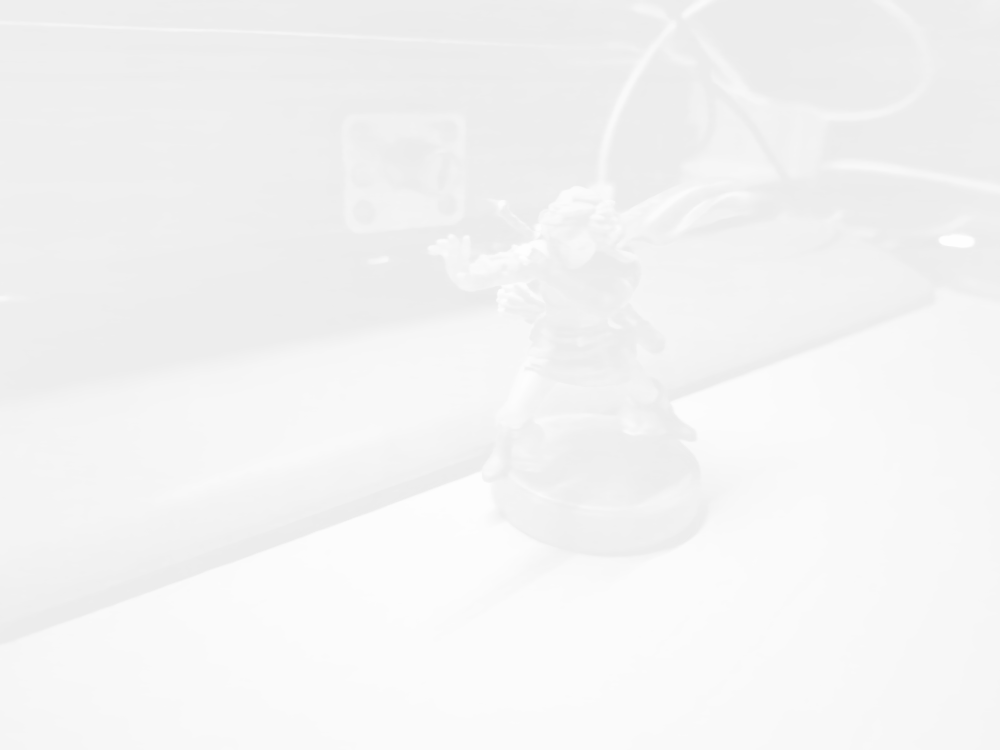}\\
    \hspace{-3.0mm}\rotatebox{90}{\makebox[1.5cm][c]{\small{{\scene{Plant}}}}}\hfill
    \includegraphics[width=\dlen]{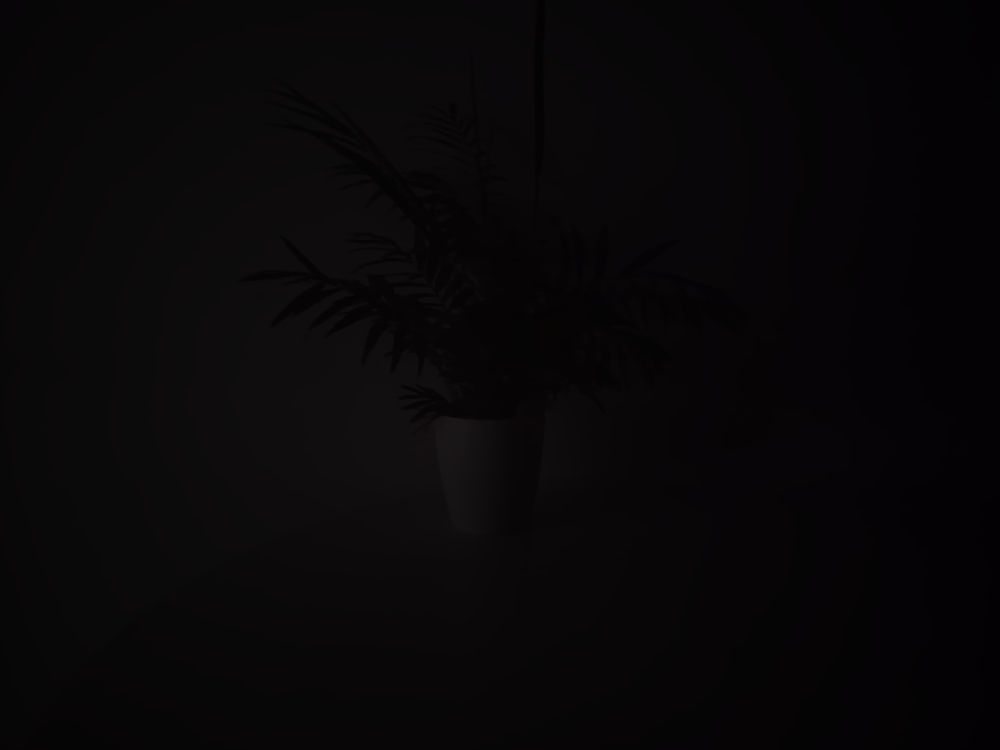}\hfill 
    \includegraphics[width=\dlen]{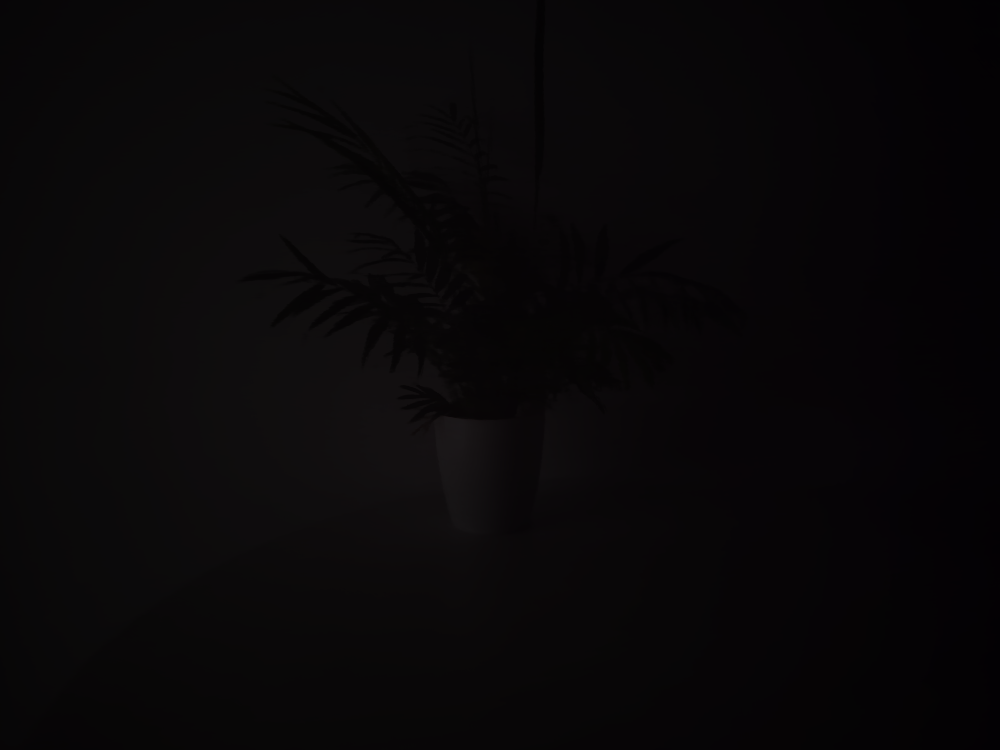}\hfill 
    \includegraphics[width=\dlen]{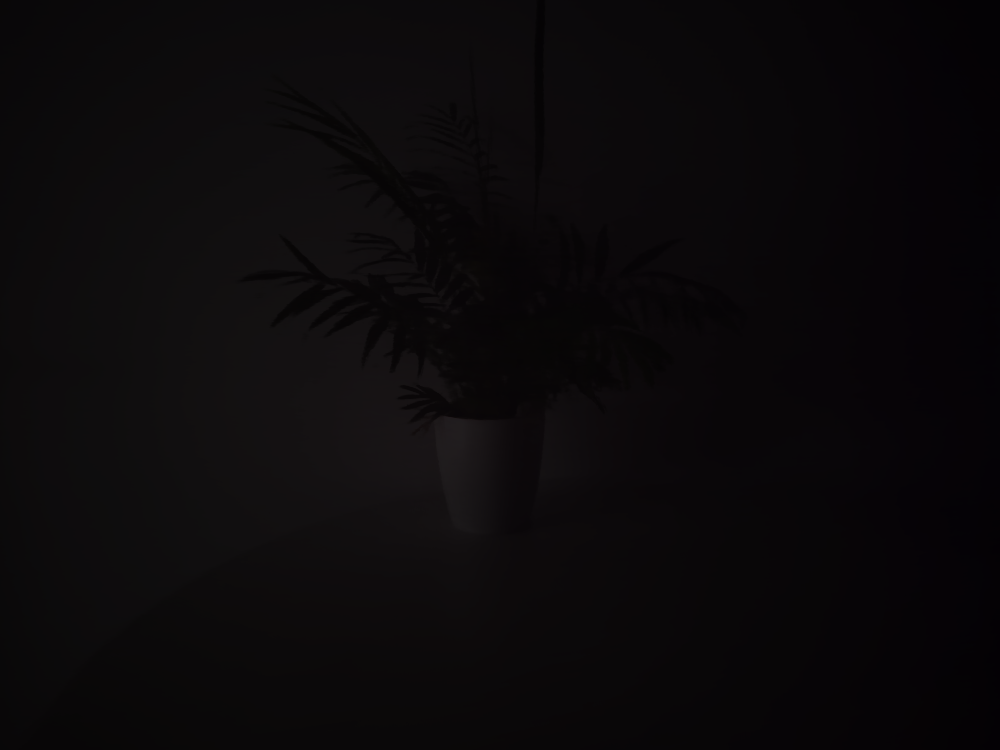}\hfill
    \includegraphics[width=\dlen]{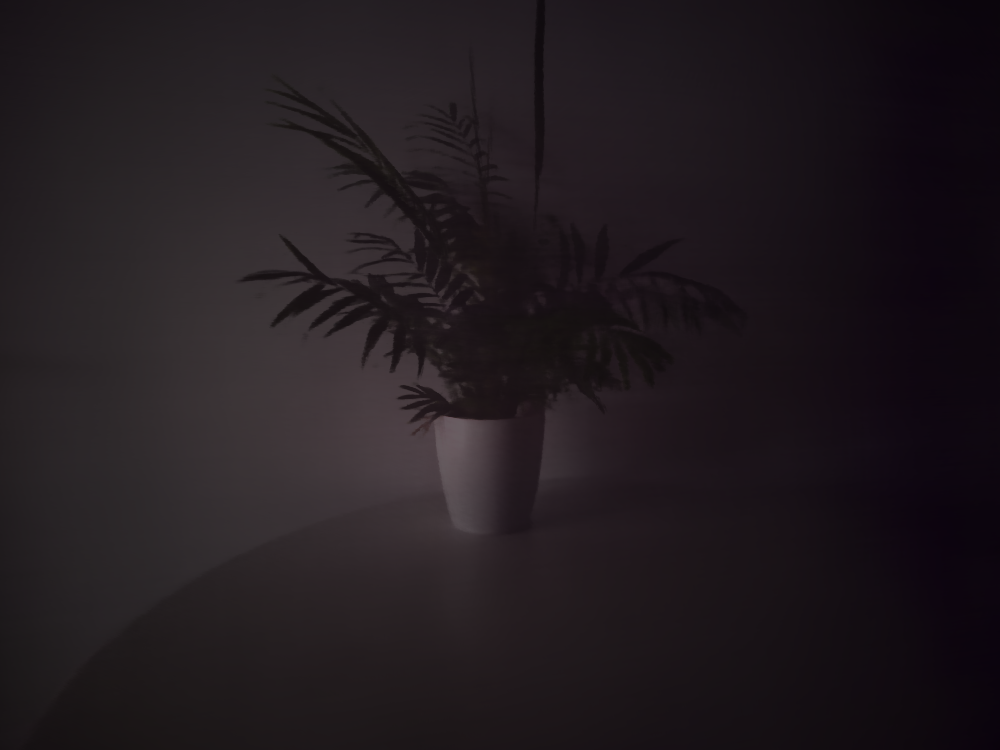}\hfill
    \includegraphics[width=\dlen]{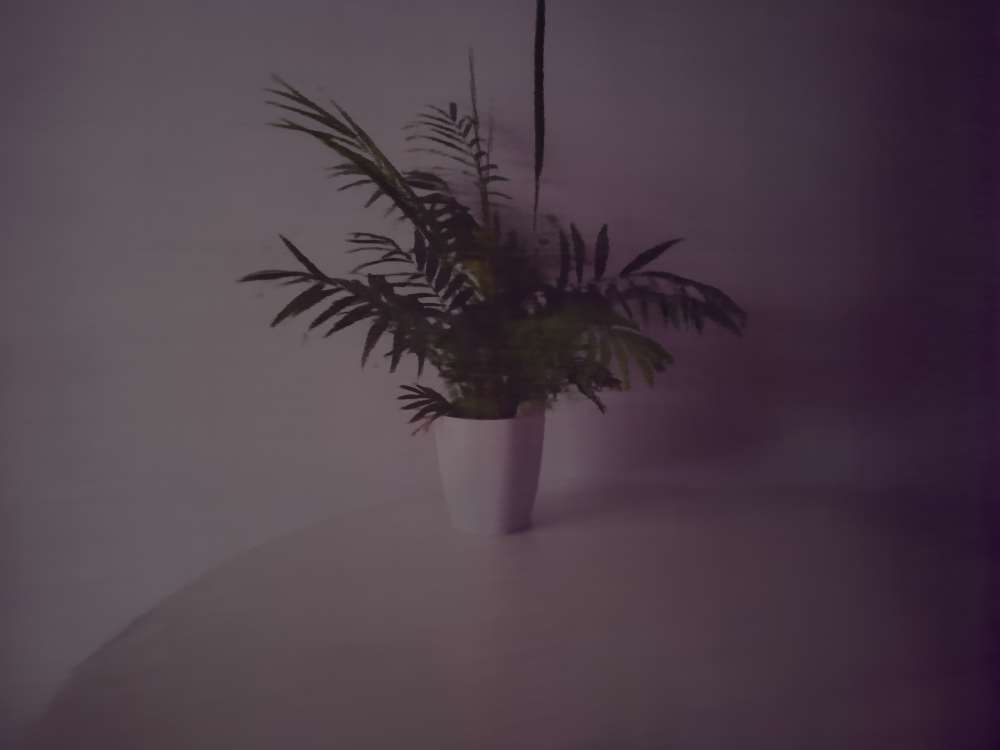}\hfill 
    \includegraphics[width=\dlen]{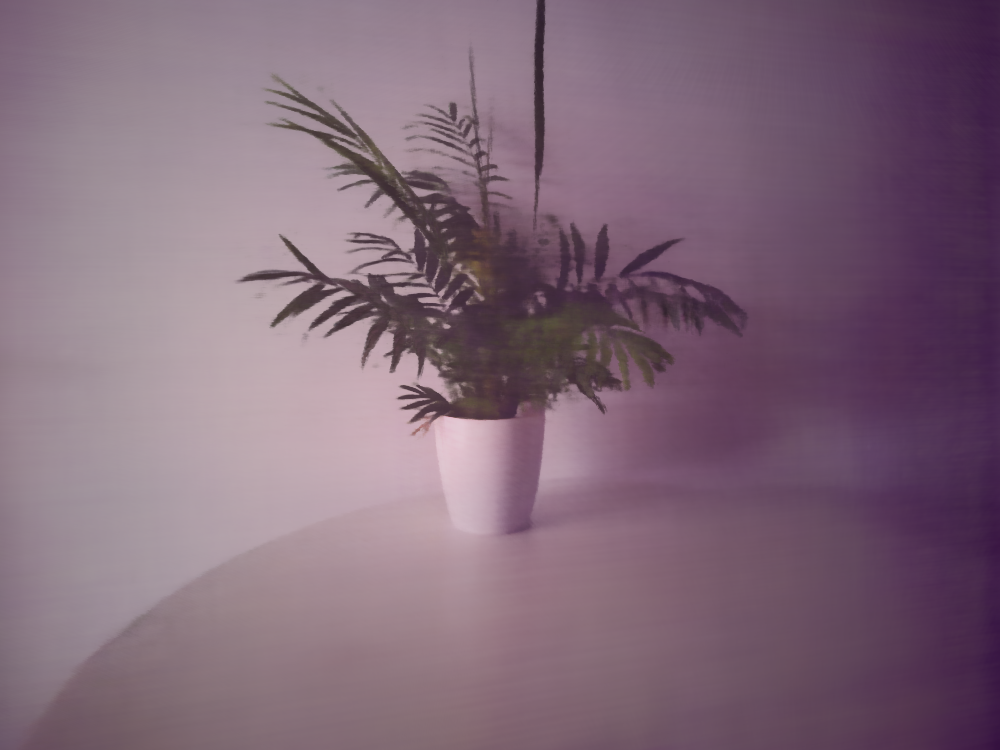}\hfill 
    \includegraphics[width=\dlen]{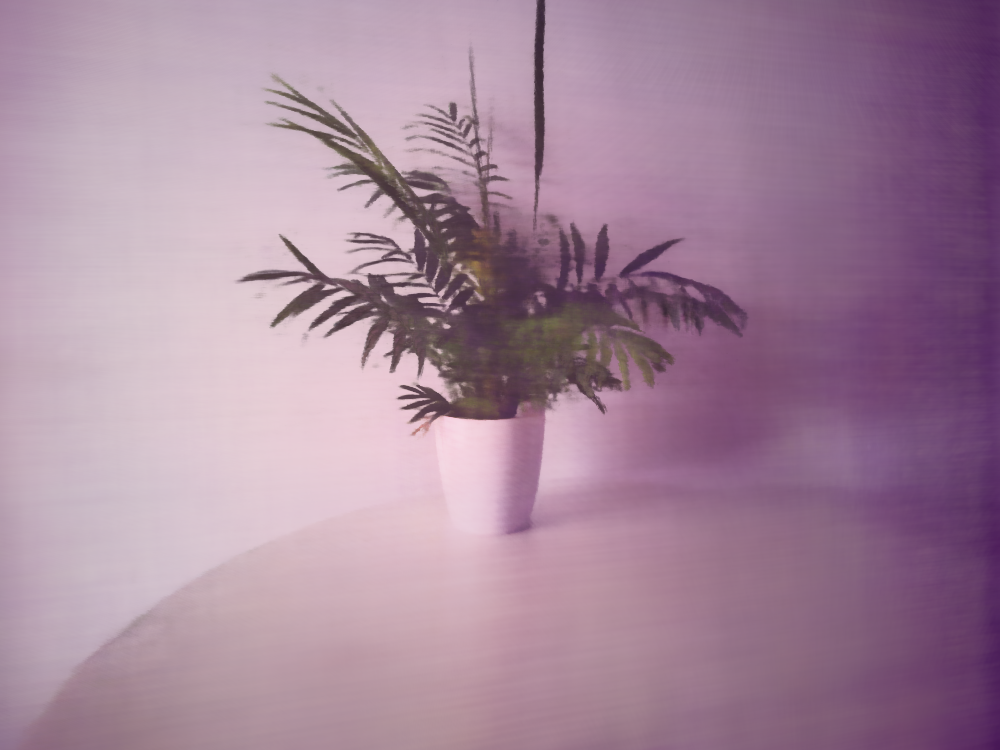}\\
    \hspace{-3.5mm}\rotatebox{90}{\makebox[1.5cm][c]{\small{{\scene{Plant (L)}}}}}\hfill
    \includegraphics[width=\dlen]{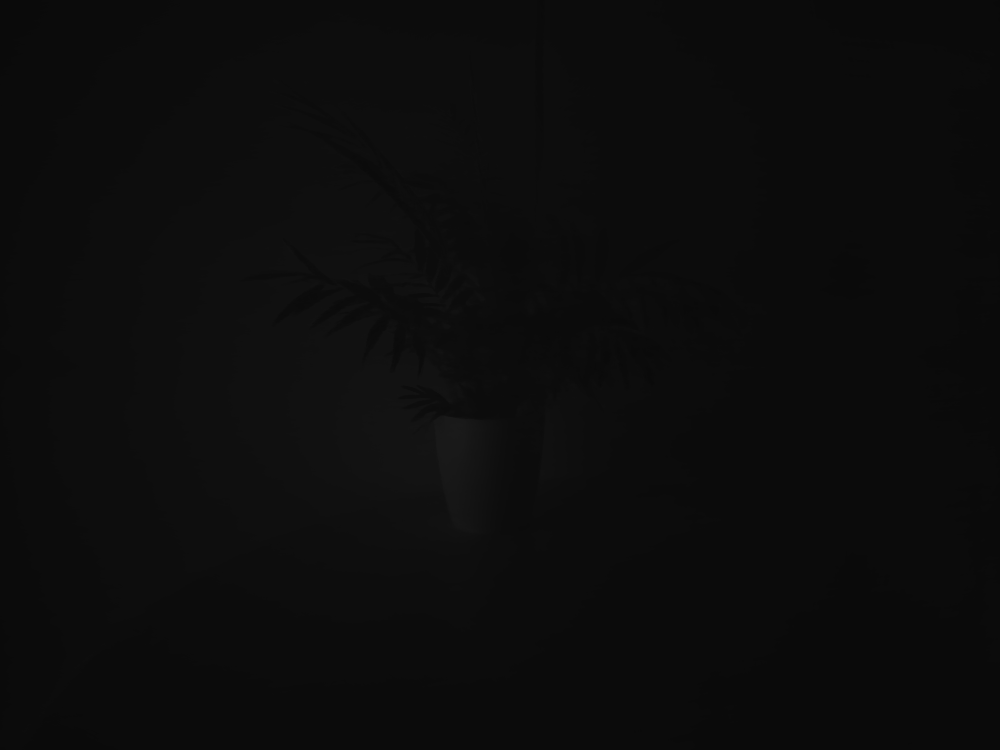}\hfill
    \includegraphics[width=\dlen]{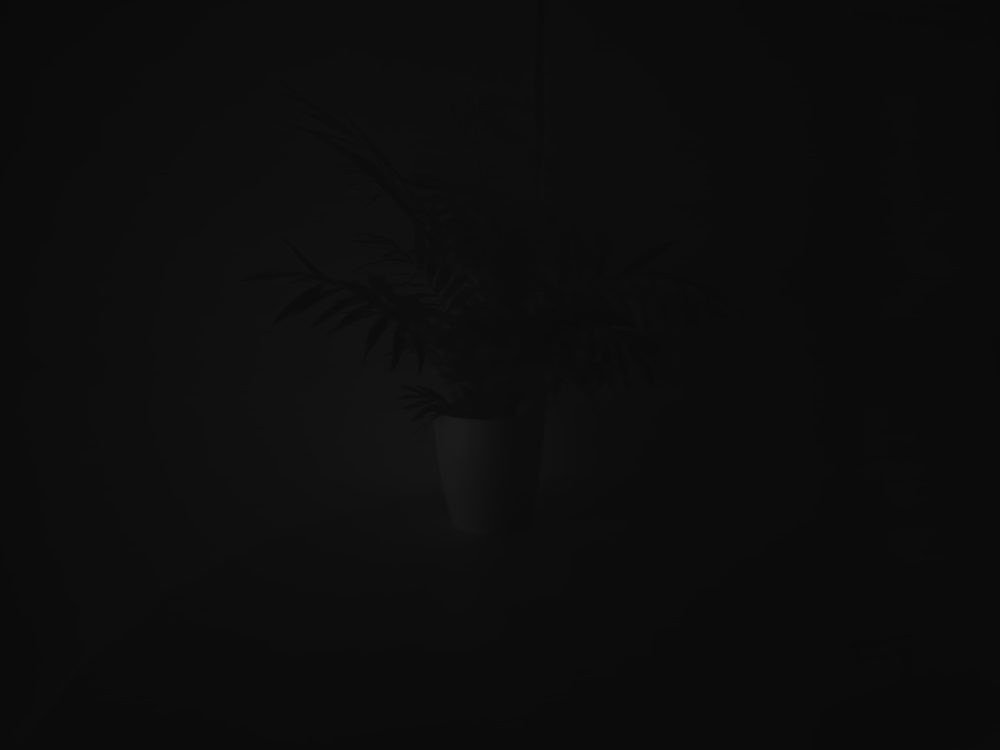}\hfill 
    \includegraphics[width=\dlen]{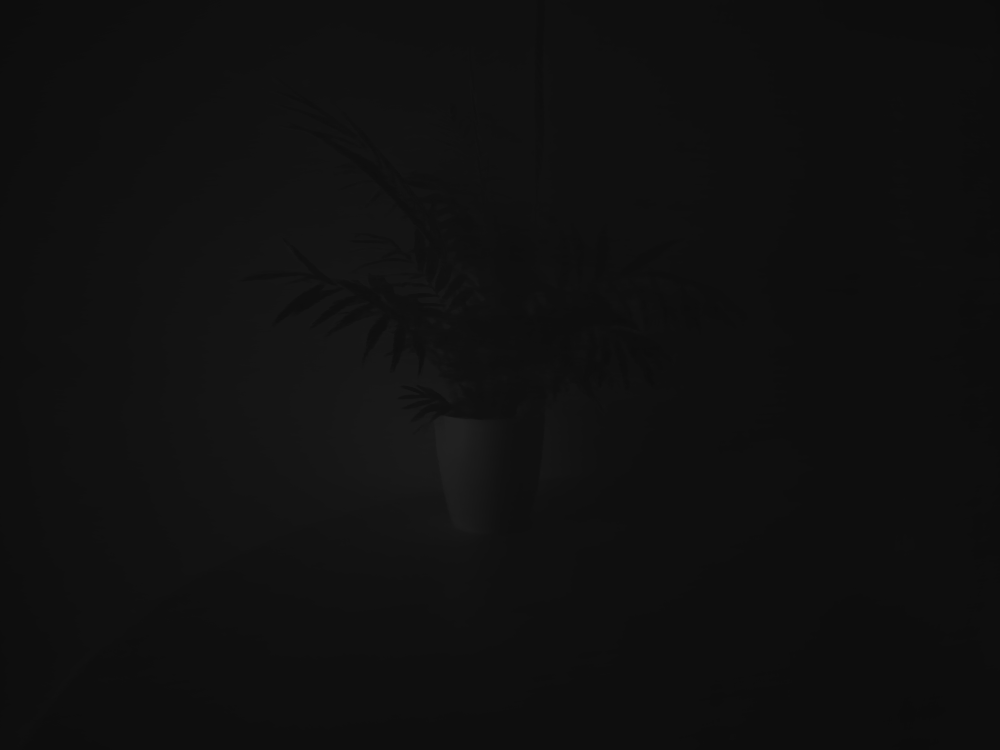}\hfill
    \includegraphics[width=\dlen]{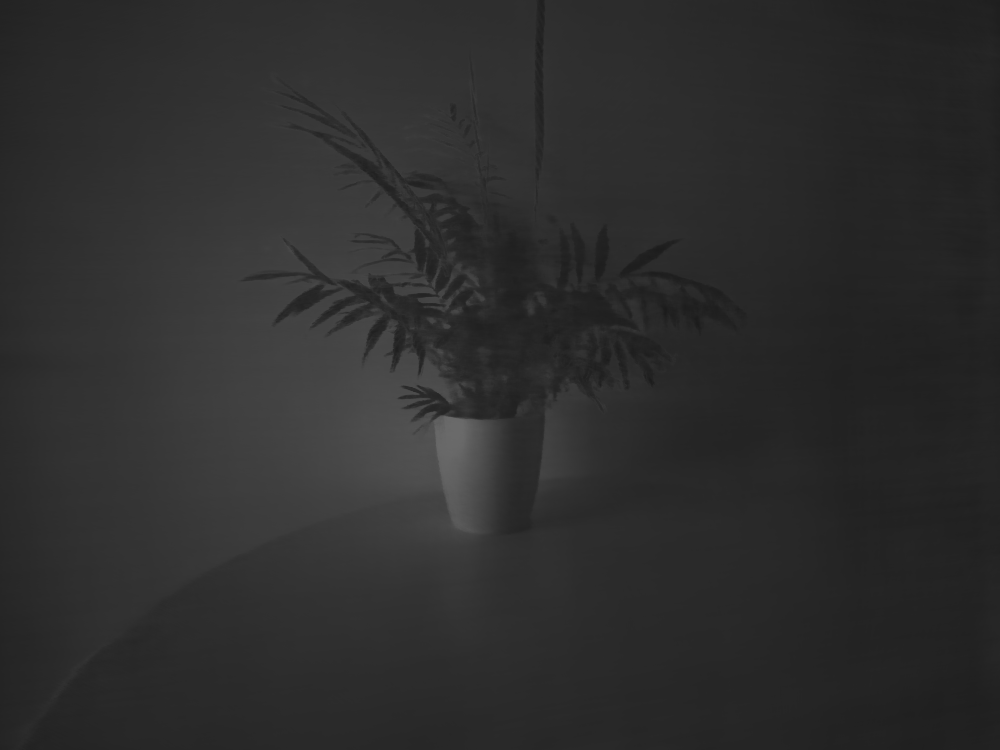}\hfill 
    \includegraphics[width=\dlen]{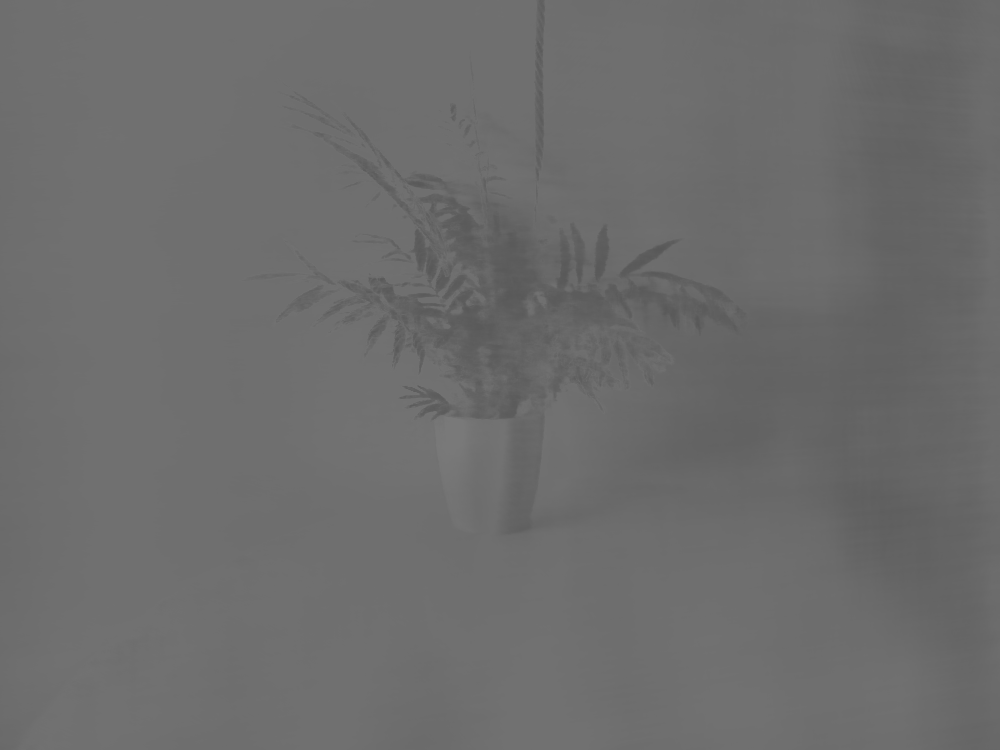}\hfill 
    \includegraphics[width=\dlen]{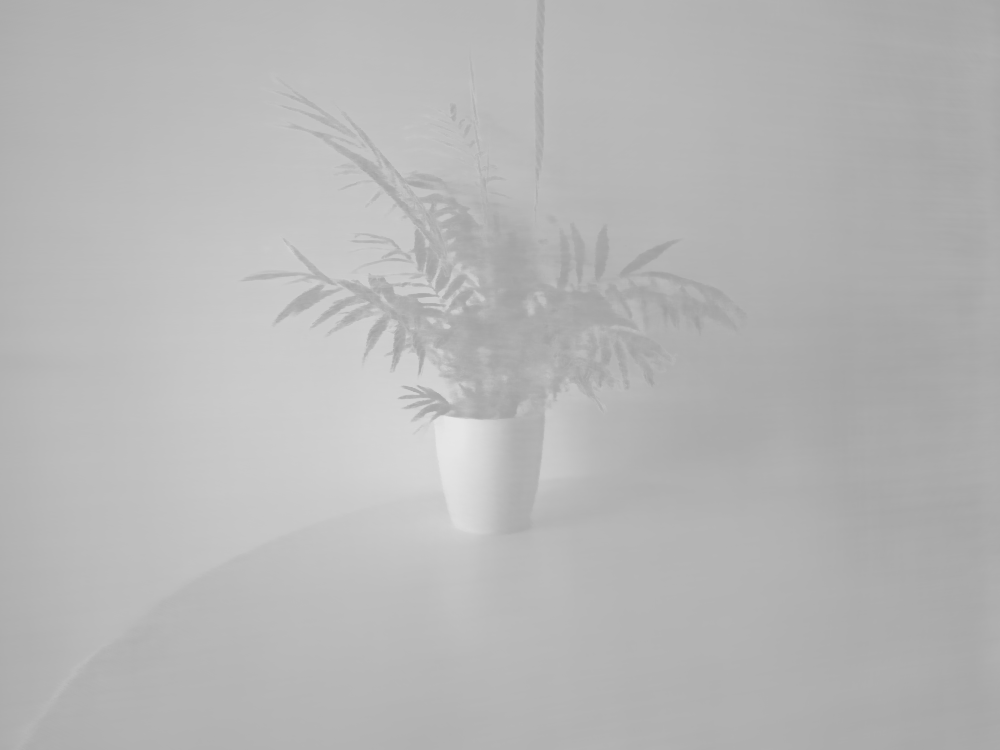}\hfill 
    \includegraphics[width=\dlen]{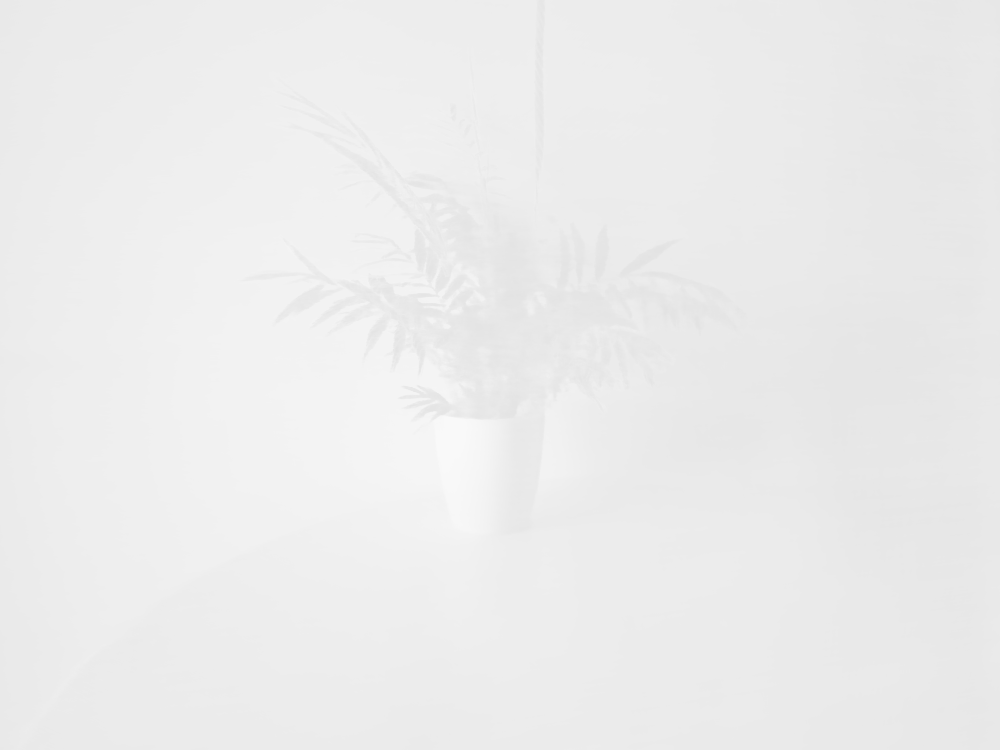}\\
    \makebox[\dlen][c]{\small{$\epsilon= 0.125$}}\hfill
    \makebox[\dlen][c]{\small{$\epsilon= 0.25$}}\hfill
    \makebox[\dlen][c]{\small{$\epsilon= 0.5$}}\hfill
    \makebox[\dlen][c]{\small{\bf{Stage 2 input}}}\hfill
    \makebox[\dlen][c]{\small{$\epsilon= 2$}}\hfill
    \makebox[\dlen][c]{\small{$\epsilon= 4$}}\hfill
    \makebox[\dlen][c]{\small{$\epsilon= 8$}}\hfill\\
    \caption{\label{fig:adjustment}Demonstrations on the generalization ability of our illumination adjustment module on the \scene{Link} and \scene{Plant} scenes. Note that the $\epsilon$ values $0.125$, $0.25$, $4$, and $8$ are not observed during the training. The second and fourth rows present the adjusted illumination components. The first and third rows are the product of the reflectance and the adjusted illumination.}
\end{figure*}

In \cref{fig:adjustment}, we demonstrate the generalization ability of our illumination adjustment module to achieve new brightness levels that are not observed from training images. Along with the adjusted images, we present their illumination components.
Our approach can robustly darken or enhance the input illumination component of the stage $2$ (4th column). If the ratio $\epsilon>1$ is set, the illumination is enhanced; If $\epsilon$ is within $\left(0, 1\right)$, the illumination is darkened. As the ratio increases, the image becomes brighter. It is worth noting that the ratio values $0.125$, $0.25$, $4$, and $8$ are not observed during training, but our approach still yields reasonable adjustment results without undesired noise and artifacts.

\subsection{Ablation Study}

\begin{figure}[t]
    \centering
    \hspace{-3.0mm}\rotatebox{90}{\makebox[1.5cm][c]{\small{{\scene{Homo.}}}}}\hfill
    \includegraphics[width=\elen]{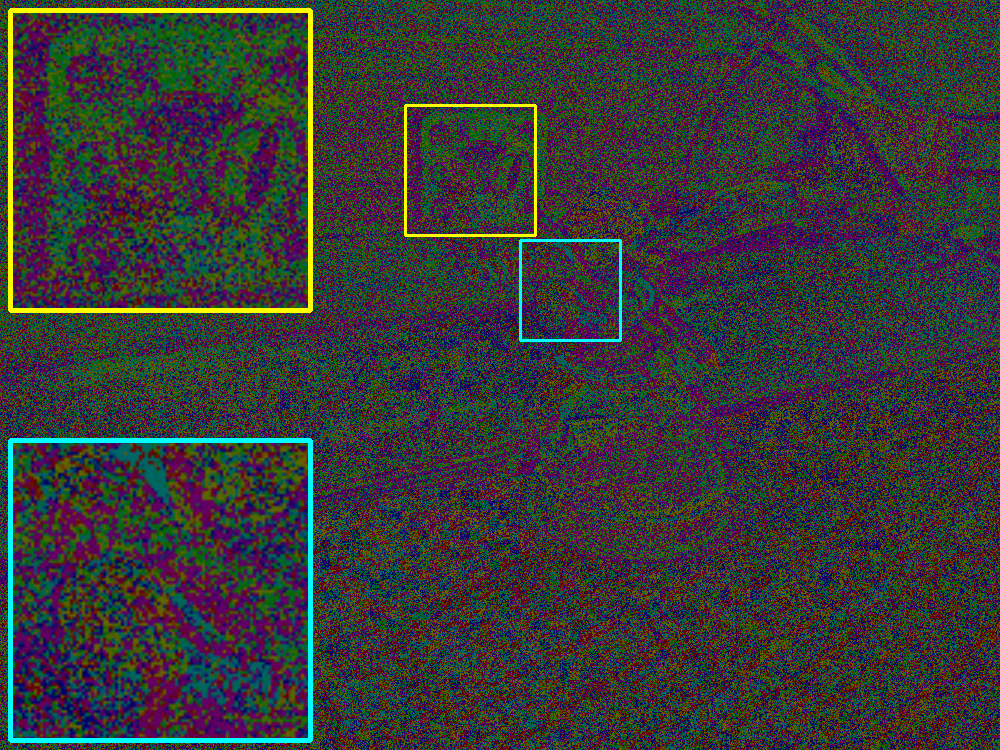}\hfill 
    \includegraphics[width=\elen]{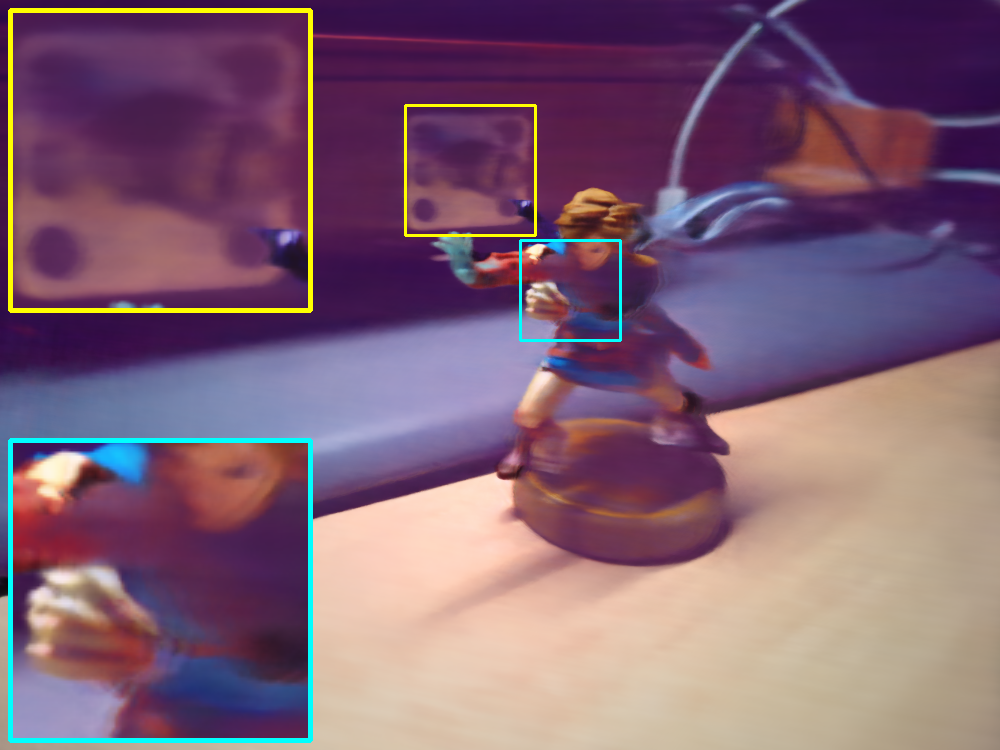}\hfill 
    \includegraphics[width=\elen]{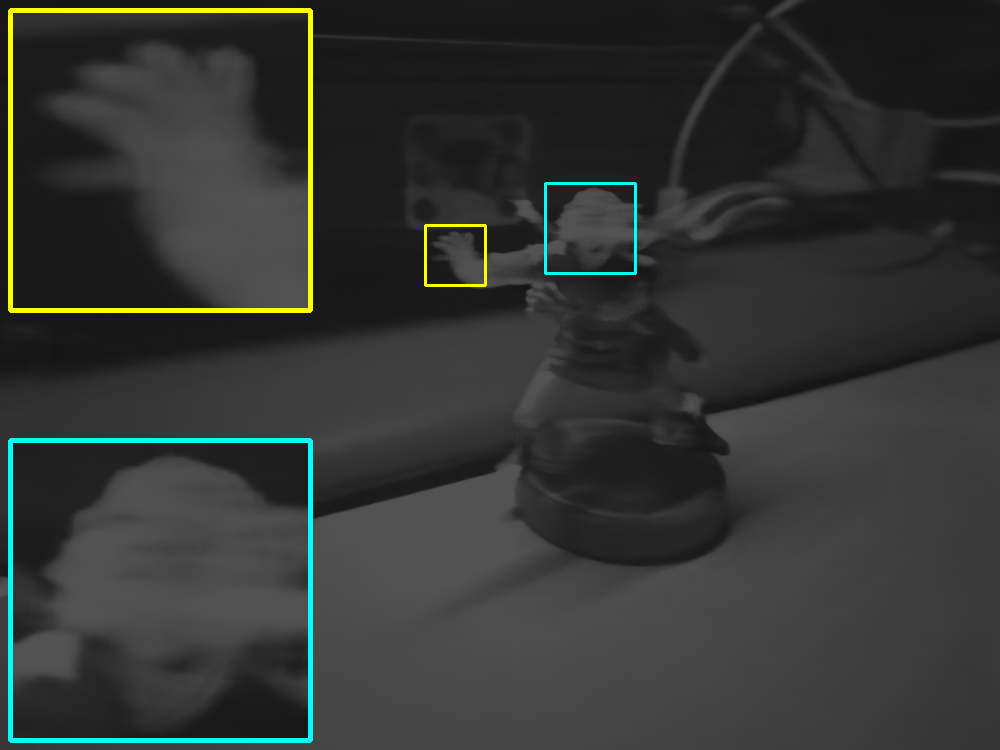}\\
    \hspace{-3.0mm}\rotatebox{90}{\makebox[1.5cm][c]{\small{{\scene{Ours}}}}}\hfill
    \includegraphics[width=\elen]{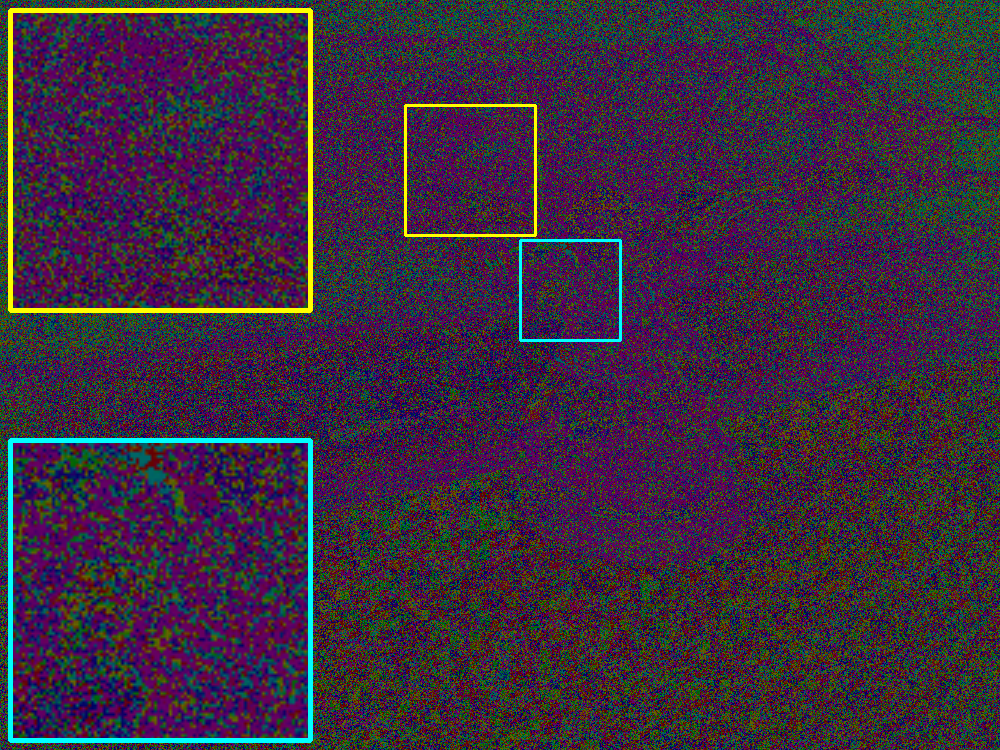}\hfill 
    \includegraphics[width=\elen]{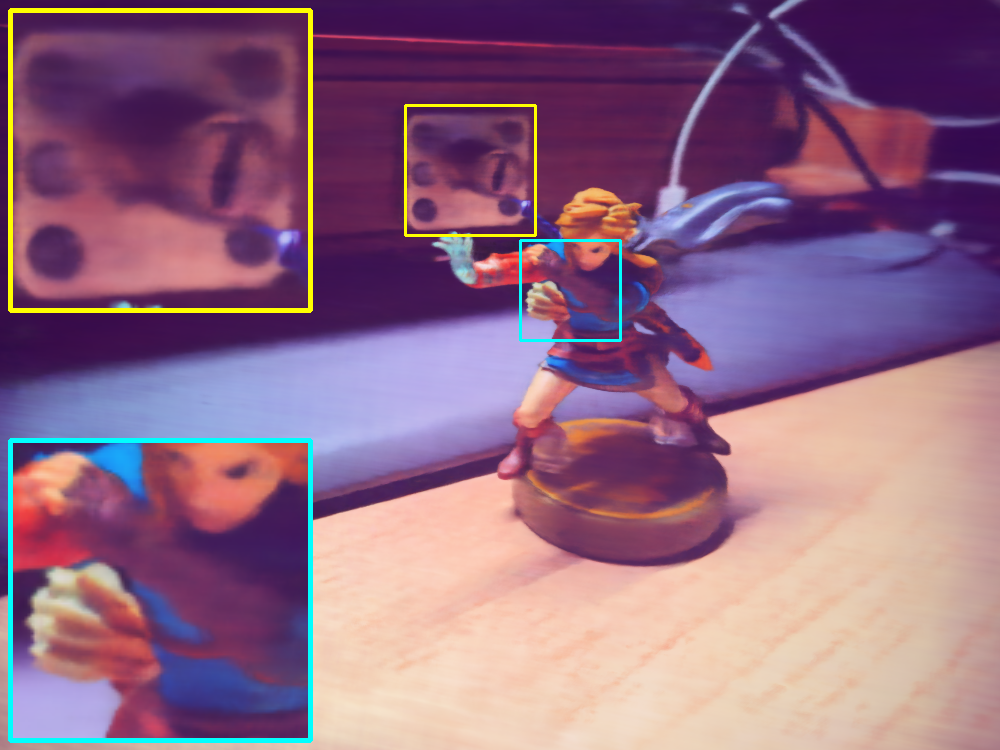}\hfill 
    \includegraphics[width=\elen]{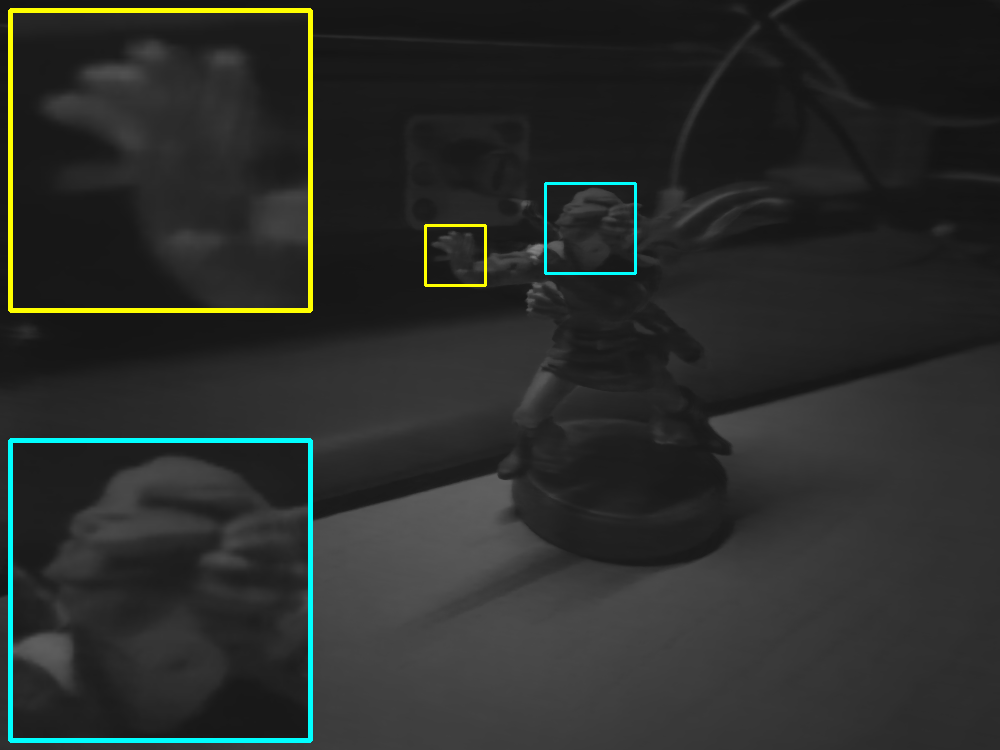}\\
    \makebox[\elen][c]{\small{Noise}}\hfill
    \makebox[\elen][c]{\small{Reflectance}}\hfill
    \makebox[\elen][c]{\small{Illumination}}\hfill\\
    \caption{Brightness homogeneity ablation by the comparisons on the decomposition results between training using images with homogeneous brightness and ours for the \scene{Link} scenes. The noise maps are normalized to $[0, 0.4]$ for visualization.}
    \label{fig:brightness_ablation}
\end{figure}

\textbf{Homogeneous and heterogeneous brightness.}~In \cref{fig:brightness_ablation}, we investigate training our approach with images of the same brightness level. The count of training views is the same as our approach. While the ablation method successfully obtains a decomposition, our approach using images of heterogeneous brightness provides a higher decomposition quality. As shown in the closeup views of the reflectance (\cref{fig:brightness_ablation}), the ablated method misses the textural details on the wall and the face of the toy. Also, its illumination presents inconsistency in local regions. Due to its decomposition error, visible geometric and textural details remain as residuals in the noise map.     

\begin{figure}[t]
    \centering
    \hspace{-2.8mm}\rotatebox{90}{\makebox[1.5cm][c]{\small{{\scene{W.o. noise}}}}}\hfill
    \includegraphics[width=0.48\linewidth]{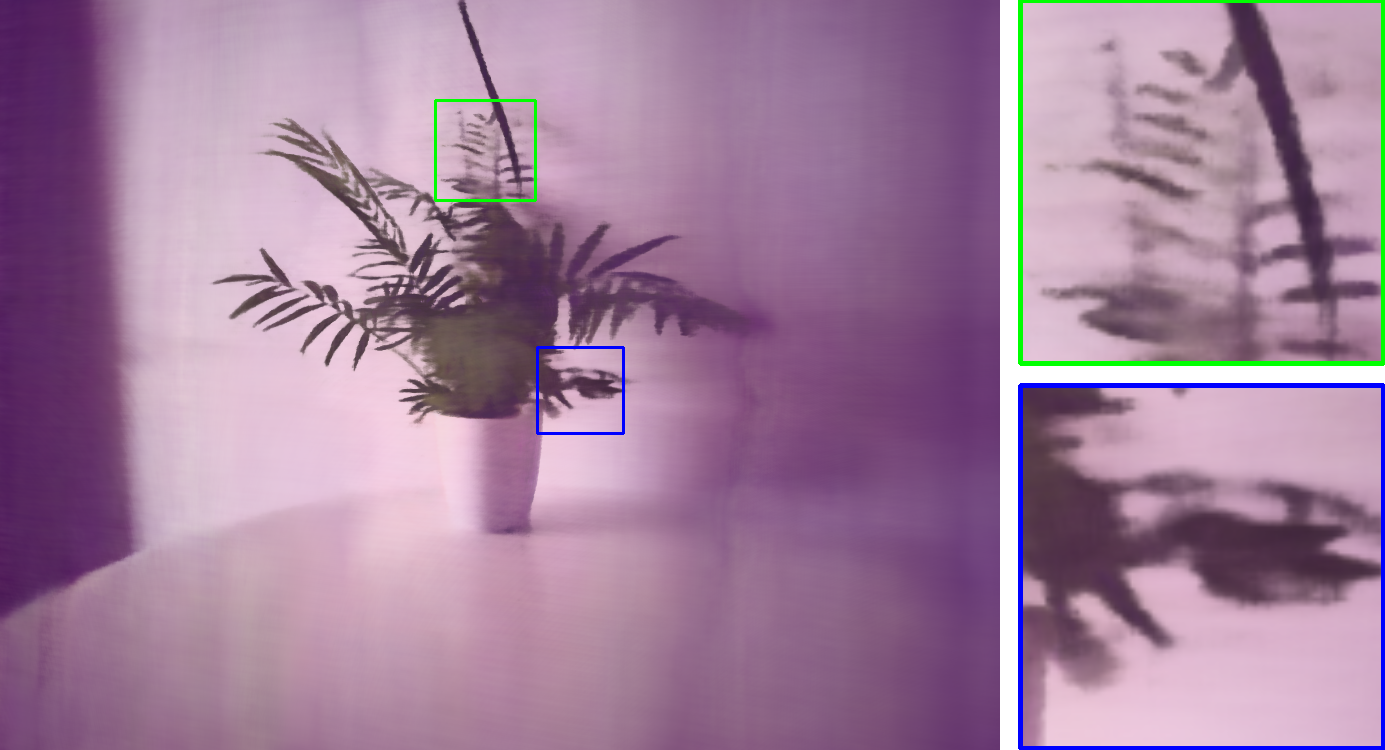}\hfill 
    \includegraphics[width=0.48\linewidth]{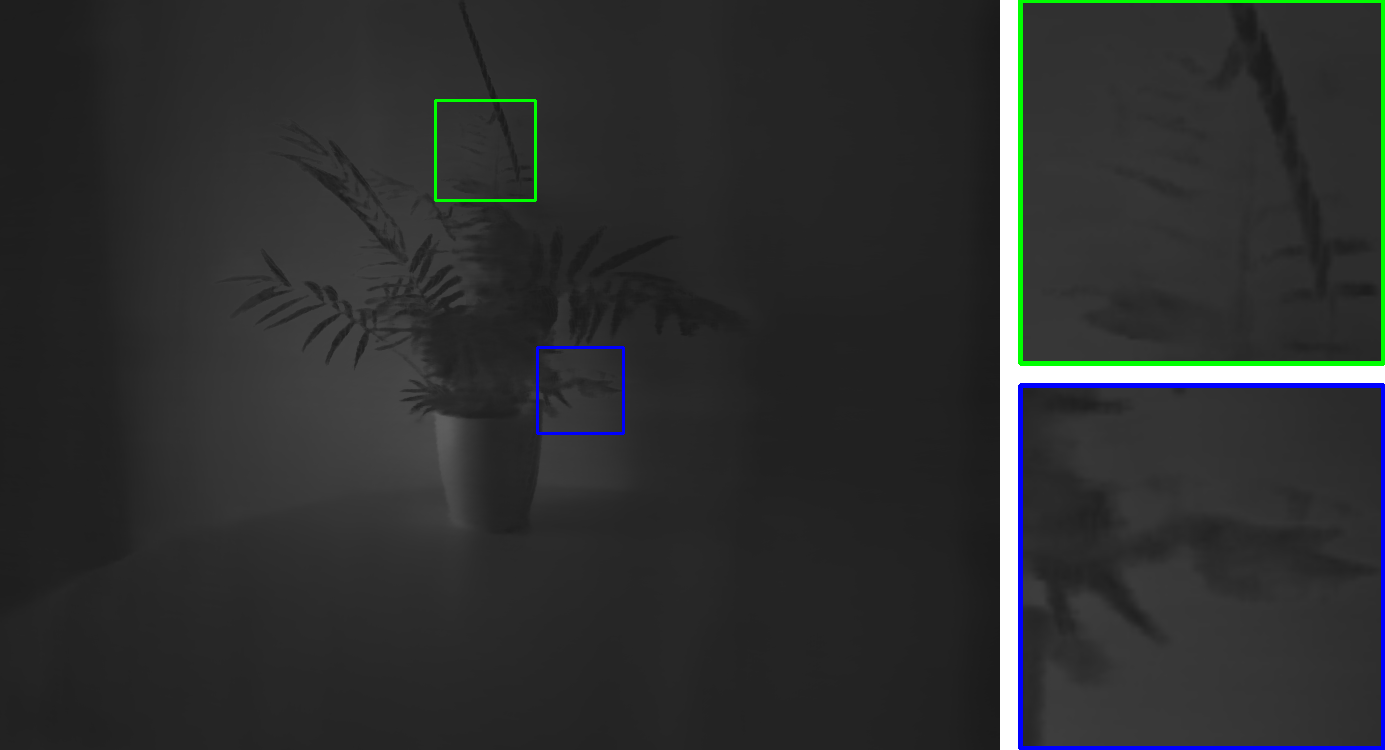}\\
    \hspace{-2.8mm}\rotatebox{90}{\makebox[1.5cm][c]{\small{{\scene{Ours}}}}}\hfill
    \includegraphics[width=0.48\linewidth]{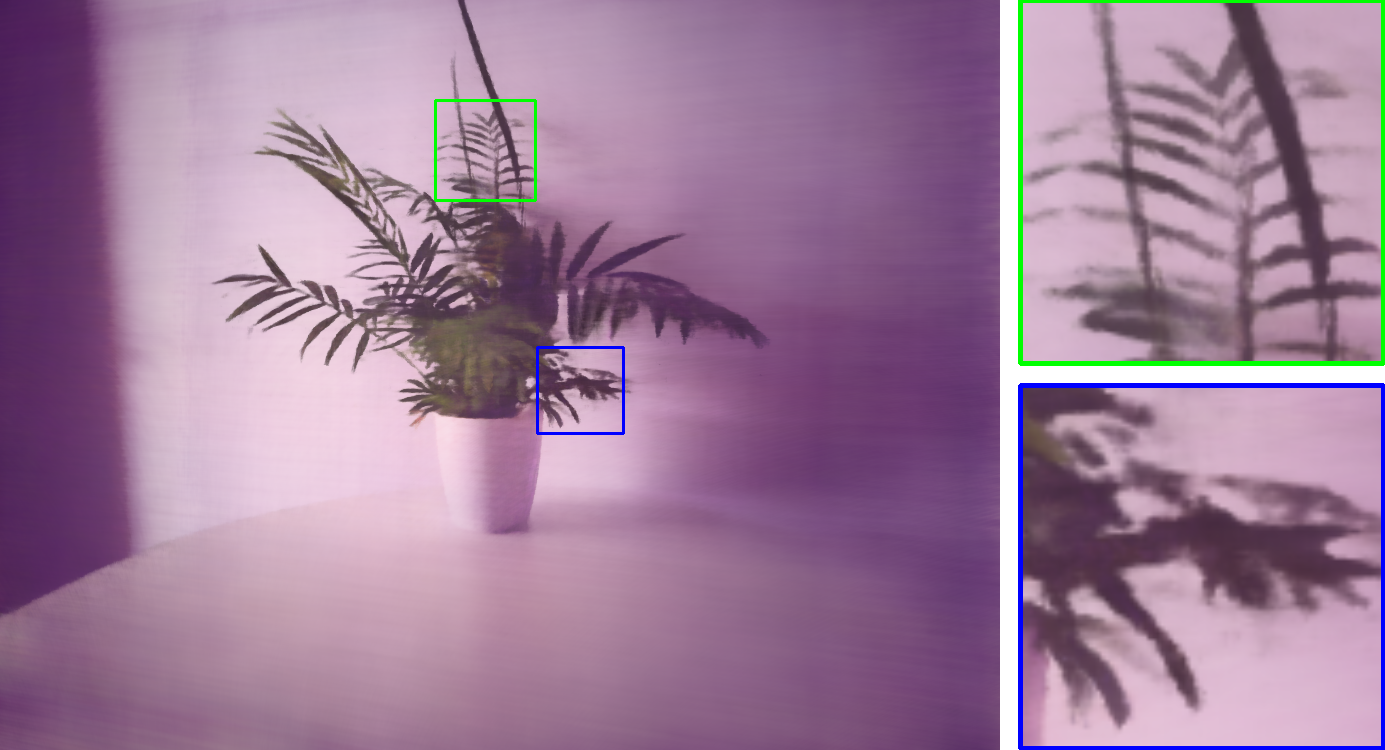}\hfill 
    \includegraphics[width=0.48\linewidth]{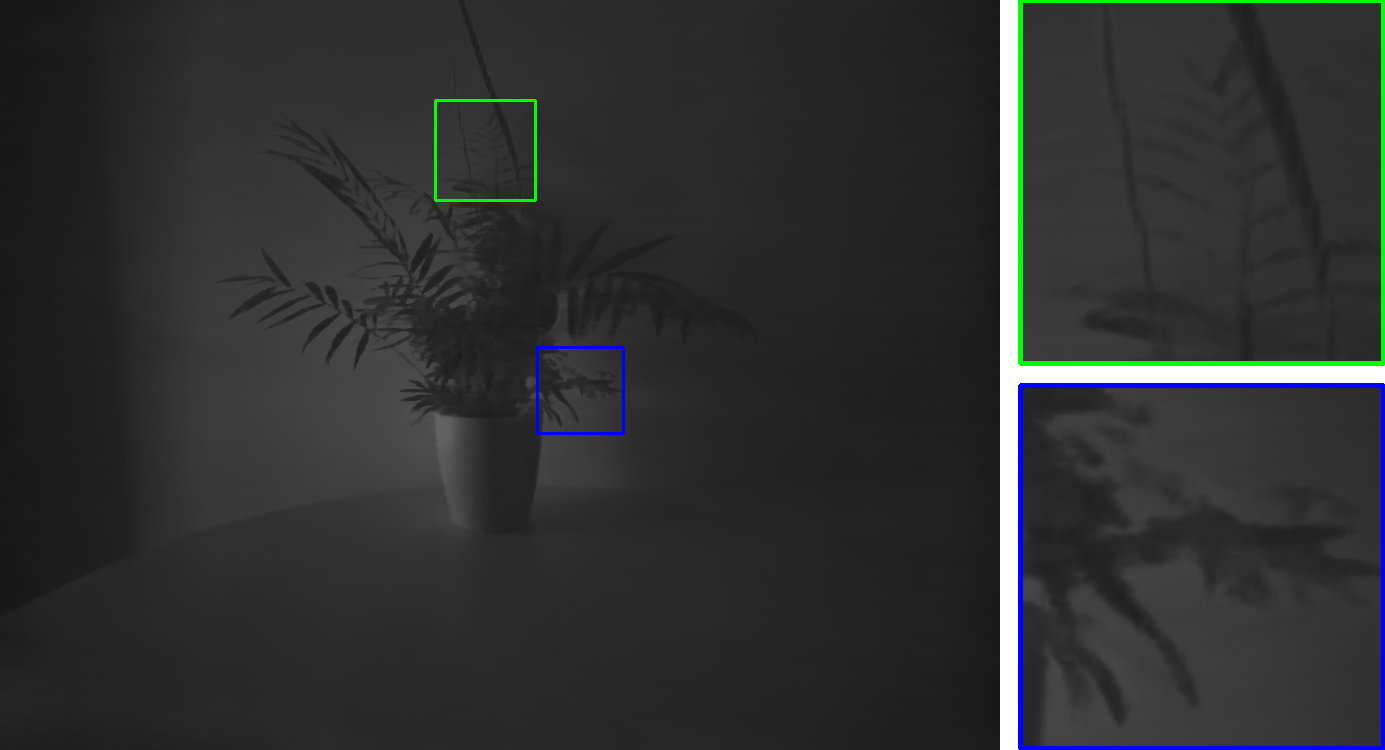}\\
    \makebox[0.48\linewidth][c]{\small{Reflectance}}\hfill
    \makebox[0.48\linewidth][c]{\small{Illumination}}\hfill\\
    \caption{Noise module ablation by the comparison on the decomposed reflectance and illumination for the \scene{Plant} scene. The ablated method cannot reconstruct the details of the foliage as shown in the closeup views.}
    \label{fig:noise_module}
\end{figure}

\noindent\textbf{Noise module.}~We conduct an ablation on the noise module of the Retinex decomposition. The noise map and its related losses are removed. We compare the decomposed reflectance and illumination components of a test view of the \scene{Plant} scene in \cref{fig:noise_module}. Without decomposing the noise, the noise causes distraction for learning the details of the scene. As shown in the closeup views on the top row, structural details of the plant are destroyed. By contrast, our approach manages to reconstruct these fine details.

\noindent\textbf{Loss functions.}~We further conduct the ablation of each loss function individually on the \scene{Potter} scene. The average quantitative metrics on the test images are tabulated in \cref{tab:loss_ablate}, which indicates that the noise regularization losses and the illumination losses play significant roles. The qualitative visual comparisons on the decomposed results and other ablation studies are presented in the supplementary.

\section{Limitations and Future Work}

\textbf{Generalization across scenes.}~Our novel view synthesis and enhancement approach is trained on a per scene basis. While it realizes plausible decomposition and enhancement results after the training on a real-world scene, it does not generalize directly to unseen scenes. A future avenue of this work is to enable the generalization across different scenes.

\noindent\textbf{Pose estimation for low-light views.}~We estimate camera poses using the images of the high scale. In the case of extremely low-light conditions, the weak signals in the captured images will cause difficulty for the camera pose estimation. We investigate the camera pose estimation using the images from the low scale and present the results in the supplementary. Tackling the pose estimation under extreme low illumination is an important future research avenue.

\section{Conclusion}

We have demonstrated a novel approach to learn neural representations from multi-view low-light sRGB images with heterogeneous brightness. The tough low-light conditions lead to low pixel values and significant camera sensor noise. Our core idea is to decompose the multi-view low-light images into the invariant reflectance, varying illumination, and individual noise map in an unsupervised manner, according to the robust Retinex theory. Based on the decomposition, we have introduced an effective and intuitive illumination adjustment module for editing the brightness of novel views, without altering the intrinsic reflectance. This work achieves a crucial step towards novel view synthesis from real-world heterogeneous low-light captures and improves the controllability of editing the brightness of novel views.

{
    \small
    \bibliographystyle{ieeenat_fullname}
    \bibliography{main}
}


\end{document}